\documentclass[conference]{IEEEtran}
\IEEEoverridecommandlockouts
\usepackage{cite}
\usepackage{amsmath,amssymb,amsfonts}
\usepackage{algorithmic}
\usepackage{graphicx}
\usepackage{textcomp}
\usepackage{xcolor}

\usepackage[capitalise,noabbrev]{cleveref} 
\usepackage{csvsimple-l3}
\usepackage{caption}
\usepackage{float}
\usepackage{subcaption}
\usepackage{array}
\newtheorem{definition}{Definition}
\DeclareMathOperator*{\argmax}{argmax}

\usepackage{url}
\def\BibTeX{{\rm B\kern-.05em{\sc i\kern-.025em b}\kern-.08em
    T\kern-.1667em\lower.7ex\hbox{E}\kern-.125emX}}
\begin{document}


\title{Who Is Lagging Behind: Profiling Student Behaviors with Graph-Level Encoding in Curriculum-Based Online Learning Systems}

\author{\IEEEauthorblockN{
Qian Xiao}
\IEEEauthorblockA{\textit{Maynooth International Engineering College} \\
\textit{Maynooth University}\\
Dublin, Ireland \\
qian.xiao@mu.ie}
\and
\IEEEauthorblockN{
Conn Breathnach}
\IEEEauthorblockA{\textit{School of Computer Science and Statistics} \\
\textit{Trinity College Dublin}\\
Dublin, Ireland \\
cobreath@tcd.ie}
\and
\IEEEauthorblockN{
Ioana Ghergulescu}
\IEEEauthorblockA{
\textit{Adaptemy}\\
Dublin, Ireland \\
ioana.ghergulescu@adaptemy.com}
\and
\IEEEauthorblockN{
Conor O'Sullivan}
\IEEEauthorblockA{
\textit{Adaptemy}\\
Dublin, Ireland \\
conor.osullivan@adaptemy.com}
\and
\IEEEauthorblockN{
Keith Johnston}
\IEEEauthorblockA{\textit{School of Education} \\
\textit{Trinity College Dublin}\\
Dublin, Ireland \\
keith.johnston@tcd.ie}
\and
\IEEEauthorblockN{
Vincent Wade}
\IEEEauthorblockA{\textit{School of Computer Science and Statistics} \\
\textit{Trinity College Dublin}\\
Dublin, Ireland \\
vincent.wade@tcd.ie}
}

\maketitle

\begin{abstract}
The surge in the adoption of Intelligent Tutoring Systems (ITSs) in education, while being integral to curriculum-based learning, can inadvertently exacerbate performance gaps.
To address this problem, student profiling becomes crucial for tracking progress, identifying struggling students, and alleviating disparities among students. Such profiling requires measuring student behaviors and performance across different 
aspects, such as content coverage, learning intensity, and proficiency in different concepts within a learning topic. 

In this study, we introduce \texttt{CTGraph}, a graph-level representation learning approach to profile learner behaviors and performance in a self-supervised manner. 
Our experiments demonstrate that \texttt{CTGraph} can provide a holistic view of student learning journeys, accounting for different aspects of student behaviors and performance, as well as variations in their learning paths as aligned to the curriculum structure. We also show that our approach can identify struggling students and provide comparative analysis of diverse groups to pinpoint when and where students are struggling. As such, our approach opens more opportunities to empower educators with rich insights into student learning journeys and paves the way for more targeted interventions.
\end{abstract}


\section{Introduction}
Intelligent Tutoring Systems (ITSs) are becoming deeply integrated at all levels of curriculum-based learning, ranging from K-12 education to higher education and professional upskilling. The development of ITSs is often motivated by their potential to 
provide personalized learning paths
that are more suitable for individual students' needs.
This motivation aligns with the United Nations’ sustainable development goal on quality education (SDG 4): ``Ensure inclusive and equitable quality education and promote lifelong learning opportunities for all'' \cite{lane2017sdg4}. However, without significantly scaling up the capacity to (a) monitor student progress and (b) provide a comprehensive data-driven framework to guide educational priorities and deepen educators' understanding of students' learning journeys, this surging adoption of digital platforms in education may risk widening 
achievement gaps
among students \cite{holstein2021equity, crouch2021priorities,suto2023holistic}.

Meanwhile, generative AI is increasingly recognised for its potential positive influence on education, such as enabling broader student access to customised learning materials. However, there is growing concern regarding the assessment of work produced by students using such generative AI tools. For instance, students may use these tools to generate assignment solutions without any genuine comprehension or deep engagement with the subject by themselves. Summative assessments, conducted entirely outside of an invigilated environment, are becoming increasingly unreliable. 
Current estimates indicate that over 70\% of students regularly use ChatGPT \cite{ibrahim2023chatgpt}. 
Therefore, in the era of generative AI, much more sophisticated and comprehensive performance management techniques are necessary to adequately reflect student effort and performance. These techniques need to identify where intervention may be required and encourage students to adopt good learning behaviours and practices. This highlights the pressing need for comprehensive monitoring and assessment of various student behaviors and performance measures. 

\begin{figure*}[th!]
\captionsetup{justification=raggedright}
     \begin{subfigure}[b]{0.33\textwidth}
         \centering
         \includegraphics[width=\textwidth,
         height=4cm,
         ,trim={4.5cm 1cm 4cm 4.5cm},clip
         ]
        {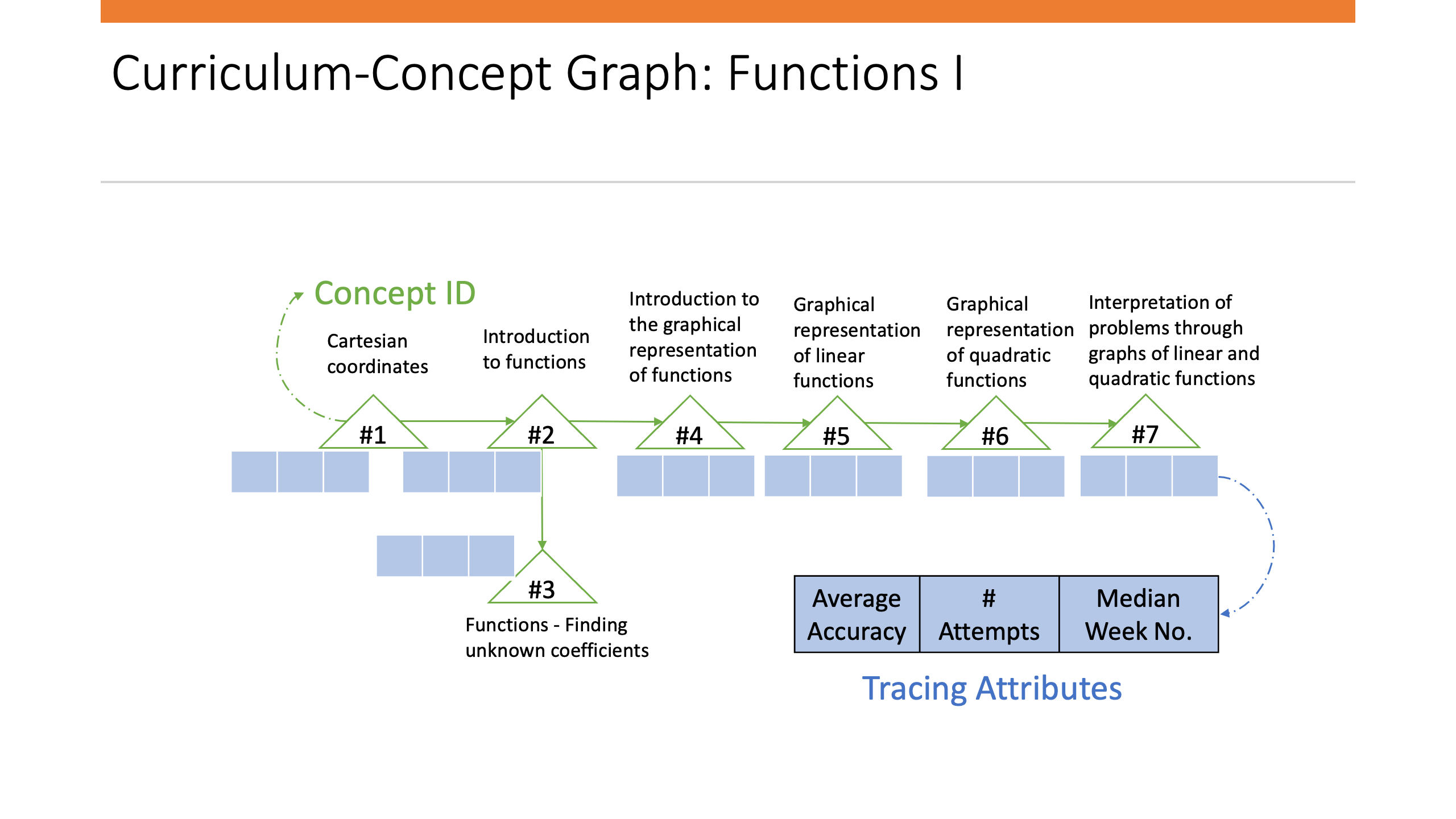}
         \caption{Curriculum Structure for the Topic \textit{Functions I}}
         \label{fig:running:curriculum:functionsI}
     \end{subfigure}
     \begin{subfigure}[b]{0.28\textwidth}
         \centering
         \includegraphics[width=\textwidth, 
         height=4cm
         ,trim={3cm 0cm 5.5cm 0cm},clip
         ]
        {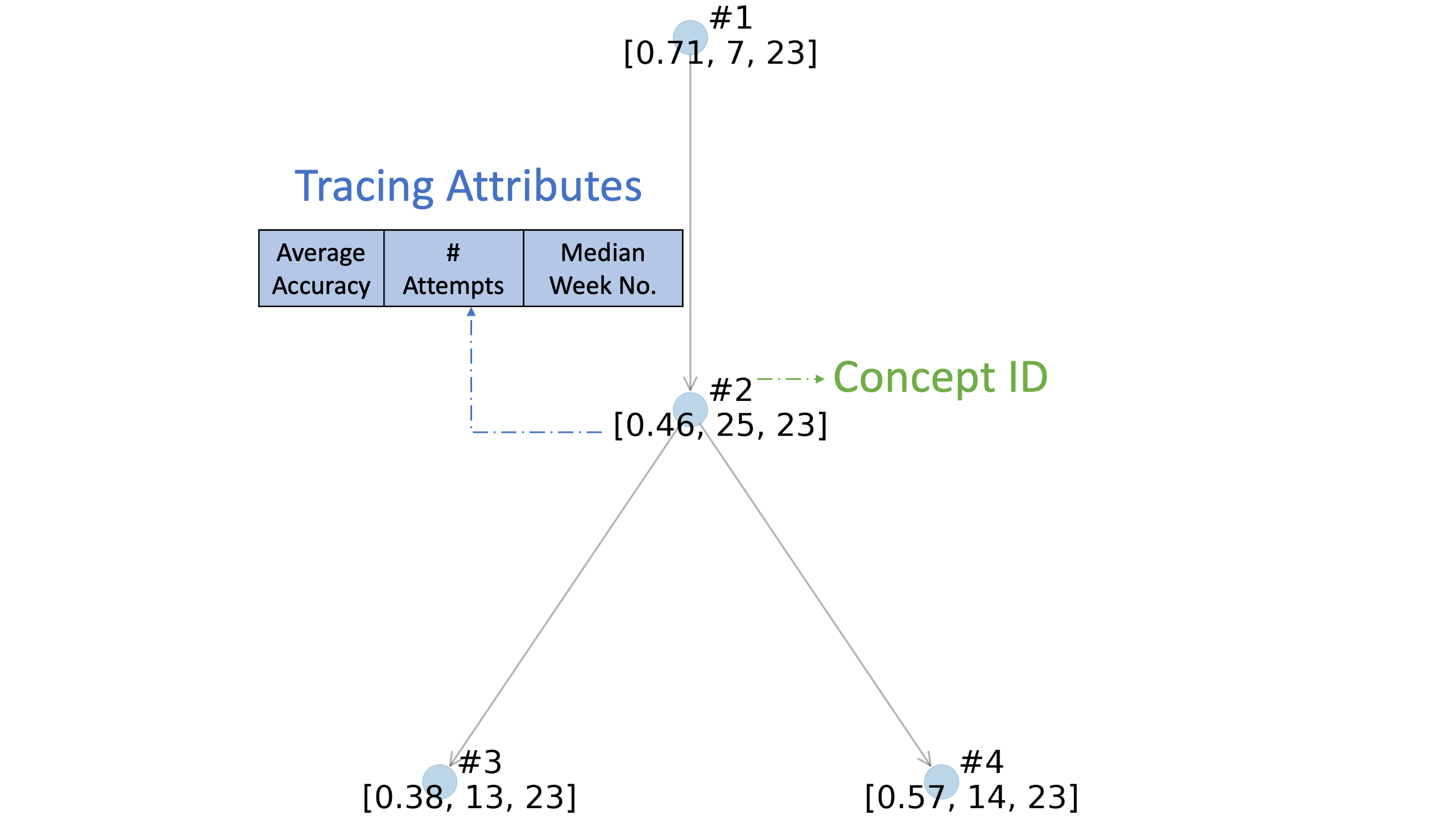}
         \caption{Student 1}
         \label{fig:running:stu1}
     \end{subfigure}
          \begin{subfigure}[b]{0.18\textwidth}
         \centering
         \includegraphics[width=\textwidth, 
         height=4cm
         ,trim={1cm 0cm 0cm 0cm},clip
         ]
        {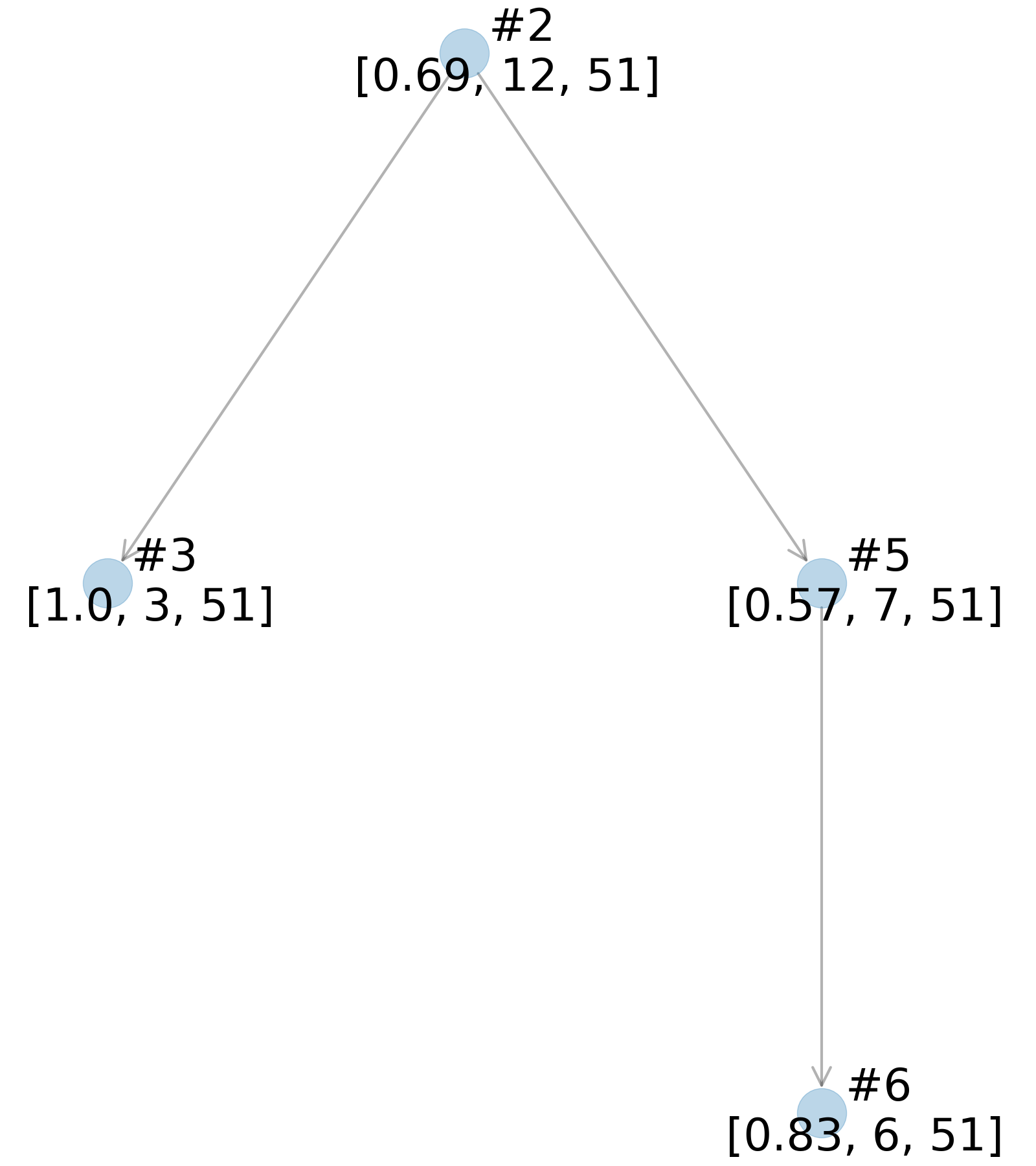}
         \caption{Student 2}
         \label{fig:running:stu2}
     \end{subfigure}
               \begin{subfigure}[b]{0.18\textwidth}
         \centering
         \includegraphics[width=\textwidth, 
         height=4cm
         ,trim={0cm 0cm 0cm 0cm},clip
         ]
        {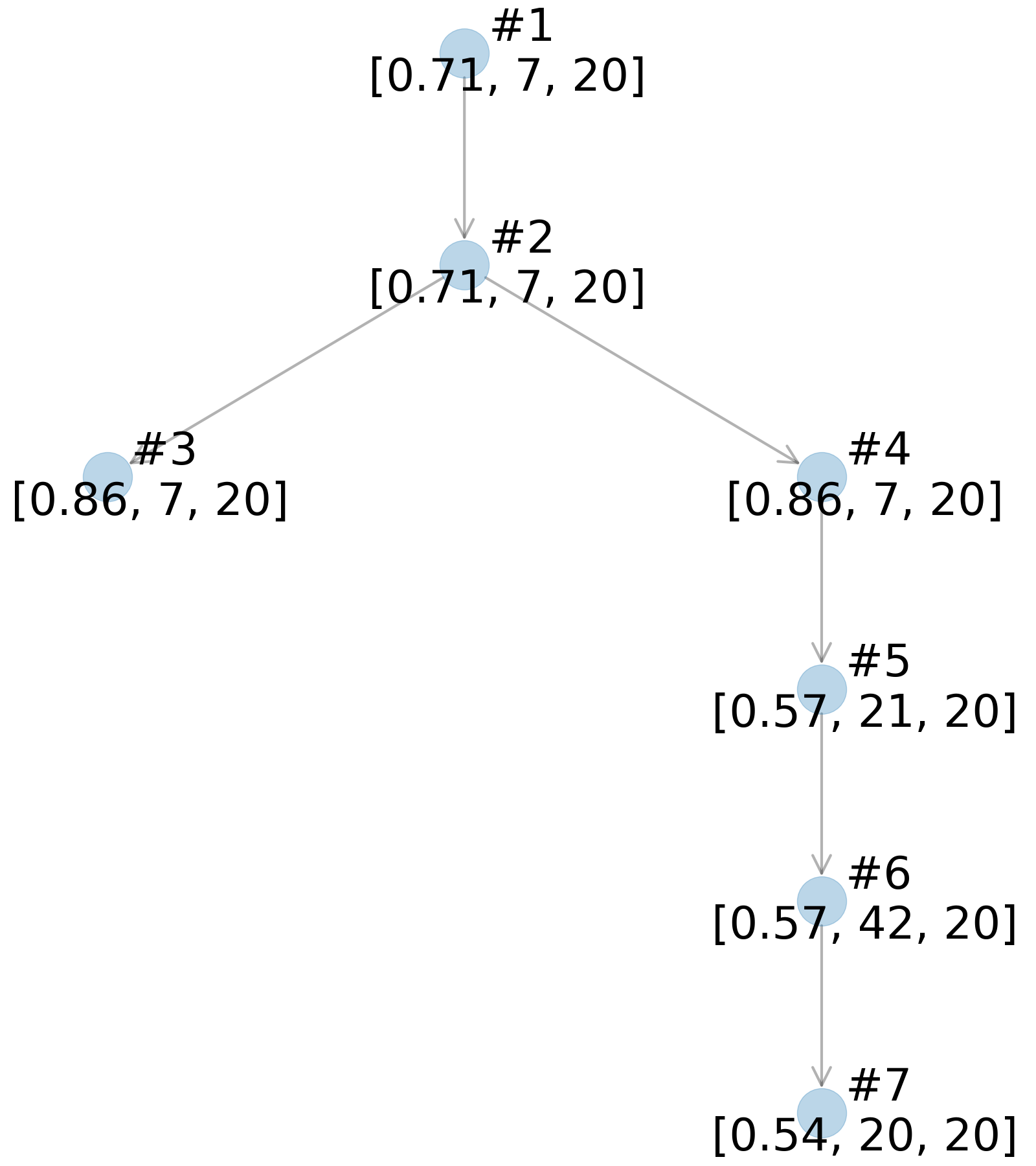}
         \caption{Student 3}
         \label{fig:running:stu3}
     \end{subfigure}
         \caption{Motivating Example: (a) presents a graph illustrating the curriculum structure for the learning topic \textit{Functions I}, where nodes represent concepts (i.e., knowledge items), directed edges represent prerequisite conditions between concepts, and node attributes include statistics of learning traces for each concept; (b) to (d) display the learning paths of three students as aligned with the curriculum structure, along with the corresponding statistics of learning traces as node attributes for each concept.}
        \label{fig:running}
\end{figure*}


Thus, it is crucial to profile students with a holistic and evidence-based framework to support more accurate and informative assessment \cite{suto2023holistic}. 
Specifically, given that learner performance is highly integrated and comprises many different interacting factors\cite{suto2023holistic}, this framework needs to include multifaceted elements of student behaviours that contribute to academic success. Moreover, to ensure leaving no one behind, we also need student profiling to identify any disparities or significant achievement gaps among individual students as well as between  student cohorts. Additionally, it is also essential to create student profiles capable of identifying when and where students are struggling.

Student profiling for ITSs requires a comprehensive assessment of student behaviors and performance across various 
aspects, particularly in terms of content coverage, learning intensity, and mastery of skills/concepts based on the curriculum structures. Unlike conventional classroom education, teachers do not have full control in ITSs over students' learning pacing and choices of studying materials. Therefore, even though the curriculum remains the same for each student, students’ navigation across the concepts in the curriculum and associated learning materials often differ significantly. 
To attain an objective assessment of student behaviors and performance in ITSs, we need 1) a holistic picture of students' varying learning status across different aspects (e.g. learning paths, intensity, coverage) 
to understand student progress; 2) the continuous monitoring and identifying when and where students are struggling; 3) the capability to enable comparative analysis among different groups of students for understanding the impact of different choices in learning paths and pacing on student performance. 


To better illustrate these challenges, we use one motivating example as depicted in Figure \ref{fig:running}. Figure \ref{fig:running:curriculum:functionsI} shows an example of concept structures for the mathematical topic \textit{Functions I} in a curriculum-based learning ITS platform. 
A student’s navigation and interactions with the learning materials associated with these concepts in this topic can be represented with attributes on each concept node. In this study, we extract average accuracy, the total number of attempts on the questions associated with a concept and the median week number in the academic calendar from tracing data to be the tracing attributes. In \cref{fig:running:stu1,fig:running:stu2,fig:running:stu3}, we list three graphs that demonstrate three different students’ navigation and interactions with the learning materials in Figure \ref{fig:running:curriculum:functionsI}. As shown in this example, only student 3 strictly adhered to the curriculum structure, while the other two students skipped learning on different concepts as indicated in the curriculum structure. Student 1 and 2 covered fewer concepts than Student 3. For each student, the performance (in terms of accuracy) varies on different concepts. For example, student 3 has much better performance in concepts 1-4 than in concepts 5-7. The learning intensity (in terms of the number of attempts on questions) and the timing of attempts (in terms of week number in the academic calendar year) are also very different. Due to such disparity of student behaviors and performance, we need to profile both student performance as well as their behaviors across different aspects to capture such diversity in students' learning journeys. Note that, single statistical indicators, while useful for understanding each aspect independently at a high level, cannot capture such cross-dimensional correlations or provide detailed comparative analysis at fine granularity for each student.

To address these challenges above, we present a novel student profiling approach that uses graph representation learning methods to model student behaviors, performance and learning paths simultaneously. Graphs are effective for representing diverse types of data in real-world applications. Representing student behaviors in graphs provides explicit structural information about the concept orders specified in the curriculum as well as students' choices in concept coverage and learning order. By assigning tracing attributes to the nodes, graphs can also incorporate rich information on student learning status for each specific concept, which is critical for identifying when and where some students are struggling. As such, graph representation learning can account for the differences across various aspects of students’ learning journeys.

Concretely, in this study, we explore the use of \textit{InfoGraph}, a graph-level representation learning technique \cite{sun2019infograph}, to profile learner behaviors and performance for curriculum-based learning in ITSs. Each student's behaviors and performance are encoded with a vector representation via graph modeling in a self-supervised training manner. This eliminates the need for specifying rules and manual labeling by educators. 


We summarize our contributions as follows:
\begin{enumerate}
    \item We propose \texttt{CTGraph}, a Curriculum learning status Tracing approach using Graph-level models in ITSs. Based on \textit{InfoGraph} \cite{sun2019infograph}, our approach can effectively capture student behavioral attributes, reflecting their variations in learning paths as aligned to the curriculum structure. 
    \item We empirically show that \texttt{CTGraph} can provide a holistic evidence-based view of student learning status in its latent space, which captures different interacting factors of student behaviors and performance. Our experiments demonstrate that such latent representations are effective in identifying different groups of students from multifaceted aspects and pinpointing students who are struggling.
    \item We discover that the latent representations obtained via \texttt{CTGraph} can be used to identify student cohort groups, in which student behaviors are mostly similar but exhibit subtle differences in certain behavioral attributes, contributory factors associated with performance, or learning paths. Our experiments show that this method enables comparative analysis among different cohort student groups with fine-grained details to understand when and where students are struggling. 
\end{enumerate}

To our knowledge, \texttt{CTGraph} is the first self-supervised learning approach for student profiling in curriculum-based ITSs. The goal of \texttt{CTGraph} is to provide representations of learning behaviors and performance that can reflect students' learning paths as aligned with the contextual structure of the curriculum. These representations can automatically identify students who are struggling and lagging behind in certain aspects for curriculum-based learning systems. We demonstrate the
superiority of \texttt{CTGraph} through extensive experiments on various learning topics using a real-world dataset. Moreover, our in-depth analysis also shows that the latent representations learned by \texttt{CTGraph} are
effective and interpretable for providing an objective measurement of learning status across diverse dimensions.







\section{Related Work}
\noindent\textbf{Student Profiling.} Profiling student behaviors to categorize students into different personas has received growing attention in educational AI research. Its applications include identifying at-risk learners \cite{aied2022profile,edm2021generalize,enhanced2021at-risk}, generating personalized feedback\cite{aied2022profile}, refining the design of learning experiences \cite{chi2017persona} and informing education priorities \cite{crouch2021priorities}. Nevertheless, most recent studies in this line use clustering approaches with metrics based on handcrafted formulation of contributory factors associated with learning. For example, Soussia et al. \cite{aied2022profile} and Mojarad et al. \cite{its2018profiling} employed k-means to cluster groups of students with similar performance and behavior characteristics based on specified rules that define consistency, pace and effort. Domenzai et al. \cite{aied2022flipped} used spectral clustering to identifying student profiles based on specified features regarding regularity, control, proactivity, etc. 

However, when such handcrafted formulations are applied to large datasets with graphs representing curriculum structures, they can lead to very high dimensional, sparse and non-smooth representations and thus yield poor generalization \cite{yanardag2015deep}. In addition, as noted in prior studies (\cite{psycho1998profilingdiff, suto2023holistic}), profiling individual differences and distinct behavior attributes is also important for educators to gain insights into students' learning journeys. However, current clustering approaches categorize students into one specified persona type. Therefore, they may overlook the impact of subtle behavioral differences among individual students. 

\noindent\textbf{Graph Neural Network Representation Learning.}
Graph Neural Networks (GNNs) learn informative representations for nodes or entire graphs through information propagation on graph-structured data. Given the versatility of graph modeling, GNN variants have found success in many diverse applications such as modeling users and personal interests for recommendation \cite{wu2021recommendation,gan2023matters,el2022twitter}, user targeting \cite{yang2023whointerested}, and graph-based service search \cite{wang2023longtail}.

Leveraging the effectiveness of GNNs, our approach jointly characterizes students' historical behaviors and performance while taking into account the contextual structure of the curriculum. To enhance the discriminative capabilities of graph models, we exploit Graph Isomorphism Networks (GINs) \cite{xu2018gin} in the architecture. GINs can effectively distinguish different structures of graphs at a level similar to the power of the Weisfeiler-Lehman graph isomorphism test.

In addition, as is common in online applications, many real-world datasets exhibit long-tail distributions. In our scenario, the distribution of student-concept interactions also exhibits such skewness, which can compromise the model's generalisation performance. Recent studies show that contrastive learning methods
 can augment models' generalization capability when learning representations of long-tail distributions 
\cite{wei2021contrastive}. Inspired by this insight, our approach 
incorporates contrastive learning and several data processing strategies to facilitate contrastive learning during the GIN training phase, which effectively enhance our model's generalization ability.

\section{Preliminaries}
In this section, we introduce the definitions of preliminary terms used in our framework.
\begin{definition}[Curriculum-Structure Graph]
A curriculum-structure graph $G_\phi=\{V_\phi, E_\phi\}$ is a directed acyclic graph for the topic $\phi$. The node set $V_\phi$ denotes the set of concepts $\Theta=\{\theta_1, \theta_2, ..., \theta_t\}$ that are included in the topic $\phi$. Each edge $e \in E_\phi$ represents the sequential order between two concept nodes, which specifies the prerequisite relationship as defined in the curriculum.
\end{definition}
\begin{definition}
[Multivariate Learning-Tracing Vector] A multivariate learning-tracing vector for the student $u$ is a vector concatenating a sequence of attribute vectors 
regarding $u$'s behaviors and performance on all attempted questions related to the concept $\theta$. It is represented as $\mathbf{x^{\theta}}(u)=\mathbf{x_1^\theta}(u) \oplus \mathbf{x_2^\theta}(u) \oplus ... \oplus \mathbf{x_m^\theta}(u)$, where $\mathbf{x_i^\theta}(u) \in \mathbb{R}^{\ell}$ is a vector with a variable length $\ell$, which describes one tracing attribute related to student behavior or performance on the concept $\theta$ in a numeric vector format. 
\end{definition}
\begin{definition}[Student Curriculum-Based Learning Graph]
    A student curriculum-base learning graph is a directed graph $G_{\phi}^u = \{V_{\phi}^u,E_{\phi}^u\}$ that represents the learning status and learning path for student $u$ on the topic $\phi$. The node set $V_{\phi}^u$ denotes the set of concepts covered by student $u$'s attempted questions in the ITS. Each node $v$ in the graph $G_{\phi}^u$ is associated with a multivariate learning-tracing vector as the node attributes, which represents student $u$'s learning behaviors and performance regarding the concept $\theta$. The structure of $G_{\phi}^u$ is derived from the curriculum-structure graph $G_{\phi}$. The directed edge $e=(v_i, v_j)$ in $G_{\phi}^u$ denotes that $v_j$ is the nearest successor of $v_i$ in $G_{\phi}$.
\end{definition}


\begin{figure*}[t!]
\captionsetup{justification=centering}
\begin{minipage}[b]{.3\textwidth}
         \centering
         \includegraphics[width=.9\textwidth, 
         height=3cm
         ,trim={1cm 3.5cm 23cm 11.5cm},clip
         ]
        {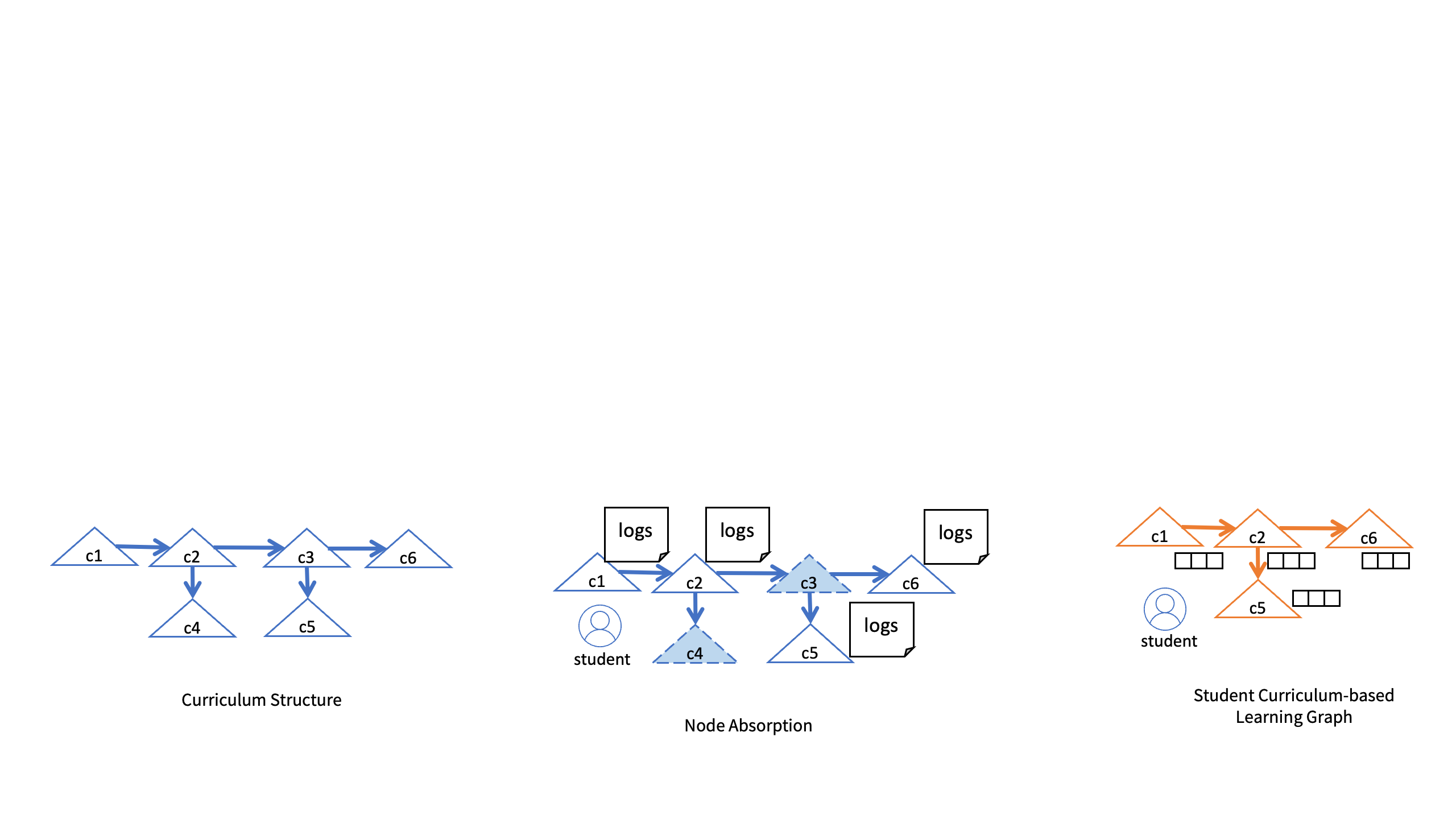}
         \caption{Curriculum-Structure Graph}
\label{fig:curriculum_graph}
\end{minipage}\hfill
\begin{minipage}[b]{.35\textwidth}
         \centering
         \includegraphics[width=\textwidth, 
         height=3cm
         ,trim={12cm 3cm 10.5cm 11.5cm},clip
         ]
        {sections/img/node-absorption.png}
         \caption{Node Absorption}
\label{fig:node_absorption}
\end{minipage}\hfill
\begin{minipage}[b]{.3\textwidth}
             \centering
         \includegraphics[width=.8\textwidth, 
         height=3cm
         ,trim={26cm 4cm 1cm 11cm},clip
         ]
        {sections/img/node-absorption.png}
        \caption{Student Curriculum-Based Learning Graph}
        \label{fig:learning_graph}
\end{minipage}
\end{figure*}

\begin{figure*}[th!]
     \centering
         \centering
         \includegraphics[width=\textwidth, 
         height=5cm
         ,trim={0cm 2cm 0cm 6.5cm},clip
         ]
        {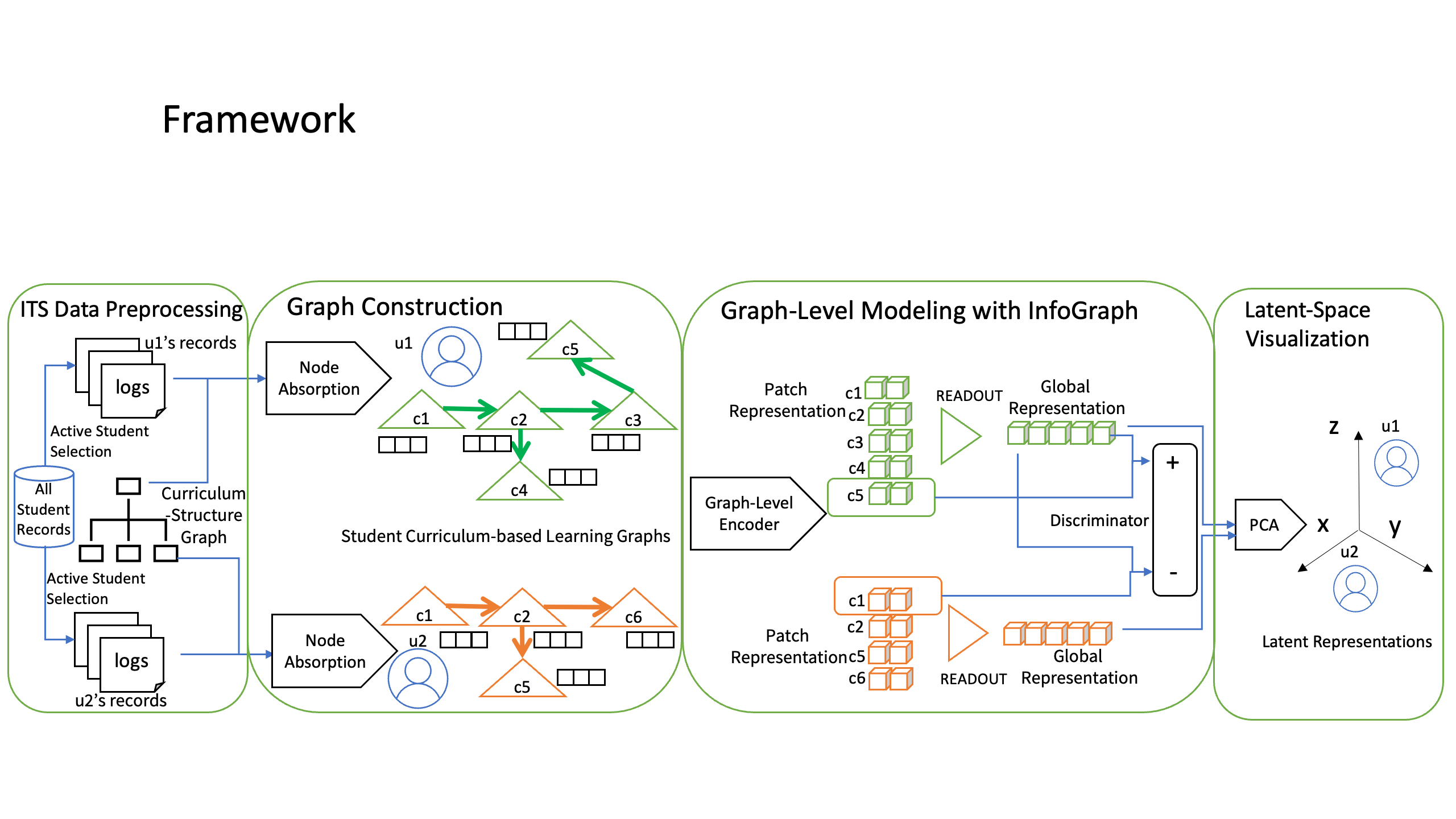}
         \caption{Overview of the \texttt{GTGraph} Framework}
         \label{fig:framework}

\end{figure*}

\section{Framework}

Our approach focuses on using graph neural networks (GNNs) to model students during curriculum-based learning.
Graphs models can explicitly represent the curriculum structure and the details of individual students' learning behaviors and performance on each concept. By employing graph models, we encode all these details into a graph-level representation for each student. 

In this section, we present the framework for our approach. We will begin with the overview of our framework and then discuss the framework design in details.

\subsection{Framework Overview}
\noindent\textbf{ITS 
Data Preprocessing.}
We collected data from the ITS, which includes 1) logs of each student's learning records that trace student behaviors and performance, and 2) the curriculum-structure graphs. In this study, for each concept and each student, we extract tracing attributes to compose the multivariate learning-tracing vector $\mathbf{x^\theta}(u)=\mathbf{x_1^\theta}(u) \oplus \mathbf{x_2^\theta}(u) \oplus \mathbf{x_3^\theta}(u)$. $\mathbf{x_1^\theta}(u)$ represents the average accuracy of all attempts made by student $u$ on questions related to the concept $\theta$, $\mathbf{x_2^\theta}(u)$ denotes the total number of such attempts, and $\mathbf{x_3^\theta}(u)$ corresponds to the median week number in the academic calendar when student $u$ attempted these questions in the ITS. It is worth noting that various other attribute types (e.g., temporal embedding via a transformer) can also be integrated into the tracing vector, capturing more nuanced details of student behaviors from different perspectives.

\noindent\textbf{Graph Construction.} For each topic $\phi$, we construct a curriculum-based learning graph $G_{\phi}^u=\{V_{\phi}^u, E_{\phi}^u\}$ for each student using \textit{Node absorption} (see details in the next subsection \ref{sec:node_absorption}). $G_{\phi}^u$ captures the details of student $u$'s learning behaviors and performance on the topic $\phi$. The graph structure reflects student $u$'s learning path in terms of  concepts. 
The node attributes, represented by the multivariate learning-tracing vectors, capture the details of $u$'s learning behaviors and performance. These attributes reflect the learning proficiency on each concept in the topic $\phi$, as well as the learning intensity, and the aggregated temporal information indicating the specific learning time in terms of the academic calendar.

\noindent\textbf{Graph-Level Encoding with InfoGraph.}
Our goal for graph-level encoding is to represent the rich information about student learning behaviors and performance 
in student curriculum-based learning graphs with a fixed-length vector. Such vectorized representation can be utilized for student profiling, which enables intuitive visualization and pattern discovery for enhancing educators' understanding of student learning journeys in the ITS. 

In this study, we leverage \textit{InfoGraph} for the graph-level encoding procedure. 
\textit{InfoGraph} learns graph-level representations by maximizing the mutual information between the global graph-level representation and the local patch representations of subgraph structures across different scales centered at each code in the graph. As such, the resultant encoding not only captures the global structure of learning paths but also the properties of tracing attributes associated with the concepts.

The flexibility of \textit{InfoGraph} makes our framework suitable for both relatively small classes with hundreds of students and large classes with thousands of students. It can also accommodate curriculum structures with topics' coverage ranging from a dozen concepts to hundreds of concepts. 


\noindent\textbf{Latent-Space Representation Extraction and Visualization.}
By extracting the vectorized representation from graph-level encoding models, we can explore student representations together in one view (visualization) inside the compressed 3D latent space. The visualization is derived via PCA compression.  With the multivariate tracing attributes as probing
indicators, the latent representations provide a comparative visualization of each student’s behaviours and performance, which
highlights the overall performance achieved by each student as well as the aggregated information that summarize their learning behaviors.

More critically, students positioned closely together in latent representation space typically selected similar learning paths on the concepts and have comparable performance. By
locating learners in the latent space, we can identify different cohorts of students who vary in the choice of learning paths, as well as in their learning intensity and timing.

\subsection{Diving Into The Framework Design Details}
In this subsection, we discuss the details of our design in the framework.

\subsubsection{Node Absorption}\label{sec:node_absorption}
Due to the abundance of question sets available, most student attempted only a subset of problems covering specific concepts within the curriculum. Consequently, for many of these students, there are no logs for certain concepts in the curriculum-structure graph. Graphs with a high number of null values in the node attributes pose challenges for contrastive learning. This is because they lead to a non-smooth distribution over node attributes, resulting in poor generalization performance \cite{narayanan2017graph2vec}.
To encourage rich and  discriminative representations, for each student, we apply a process named \textit{node absorption}.
Specifically, for each student, this process removes concept nodes that lack learning logs within the curriculum-structure graph. Nodes are subsequently reconnected if they are the nearest successors to other nodes within this structure. Through this method, we construct the student's curriculum-based learning graph for each student, which uses multivariate learning-tracing vectors as node attributes and eliminates any null values in the attribute vectors.

Figure \ref{fig:node_absorption} illustrates an example of node absorption, which is based on the curriculum-structure graph in Figure \ref{fig:curriculum_graph}. In this example, the student in Figure \ref{fig:node_absorption} did not attempt any questions related to concepts $c3$ and $c4$. Therefore, we first remove the concept nodes $c3$ and $c4$ from the graph. Subsequently, we reconnect the remaining nodes if they are the nearest successors in the corresponding curriculum-structure graph (as depicted in Figure \ref{fig:curriculum_graph}). The resulting student curriculum-based learning graph is depicted in Figure \ref{fig:learning_graph}. 

Such graph construction design is critical, which ensures graph-level representations remain discriminative towards other graph instances. 

\subsubsection{
Student Selection with Medium-High Concept Coverage Attainment Threshold}
For model training, we select students who attempted the questions covering at least a certain percentage of the total number of concepts for a given topic. This is because we encourage our model to learn diverse learning paths undertaken by students. By focusing on these students, we can ensure that each student had sufficient engagement with the ITS and studied questions covering a range of concepts that form diverse learning paths. 
More importantly, \textit{InfoGraph} is based on contrastive learning during training. Contrastive-learning methods often require a large number of negative samples to be effective \cite{hjelm2018learning}.
Thus, selecting students with diverse learning paths to generate different negative samples during training is crucial, since learning graph embeddings requires many different graph instances during training. 

\section{Methodology}
In this section, we formulate the graph representation learning problem formally and present the details of the graph-level encoding and representation learning method. 

\subsection{Problem Definition}
\noindent\textbf{Graph-Level Representation Learning.} Given a set of student curriculum-based learning graphs $\mathbb{G_\phi}=\cup_u^\mathcal{U}\{G_{\phi}^u\}$ on the topic $\phi$ for all students $\mathcal{U}$ and a positive integer $d$ (the expected embedding size of the representation), our goal is to learn a $d$-dimensional representation $\textbf{y}$ for each graph $G_{\phi}^u\in \mathbb{G}$, where
$\textbf{y}$ is a fixed-length vector $\left( y_1, y_2, ..., y_d\right) \in \mathbb{R}^{d}$. 

\subsection{Graph-Level Modeling with InfoGraph}
\begin{figure*}[h]
\centering
\begin{minipage}[b]{.4\textwidth}
         \centering
         \includegraphics[width=\textwidth, 
         height=3.5cm
         ,trim={0cm 3.5cm 9cm 5cm},clip
         ]
        {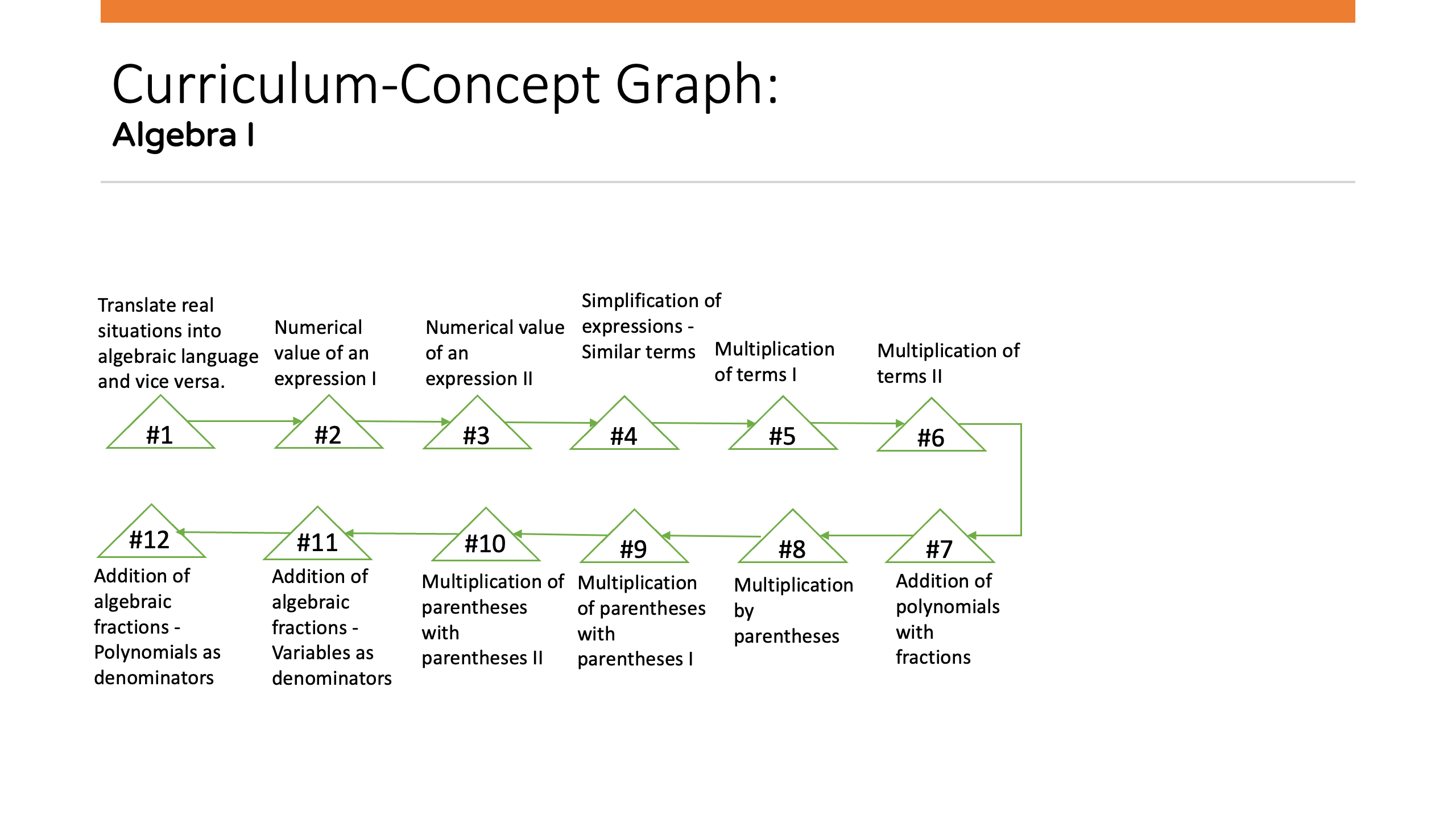}
         \caption{Curriculum-Structure Graph for the Topic \textit{Algebra I}}
\label{fig:curriculum_graph:algebraI}
\end{minipage}\hfill
\begin{minipage}[b]{.6\textwidth}
         \centering
         \includegraphics[width=\textwidth, 
         height=4cm
         ,trim={0cm 1cm 0cm 5cm},clip
         ]
        {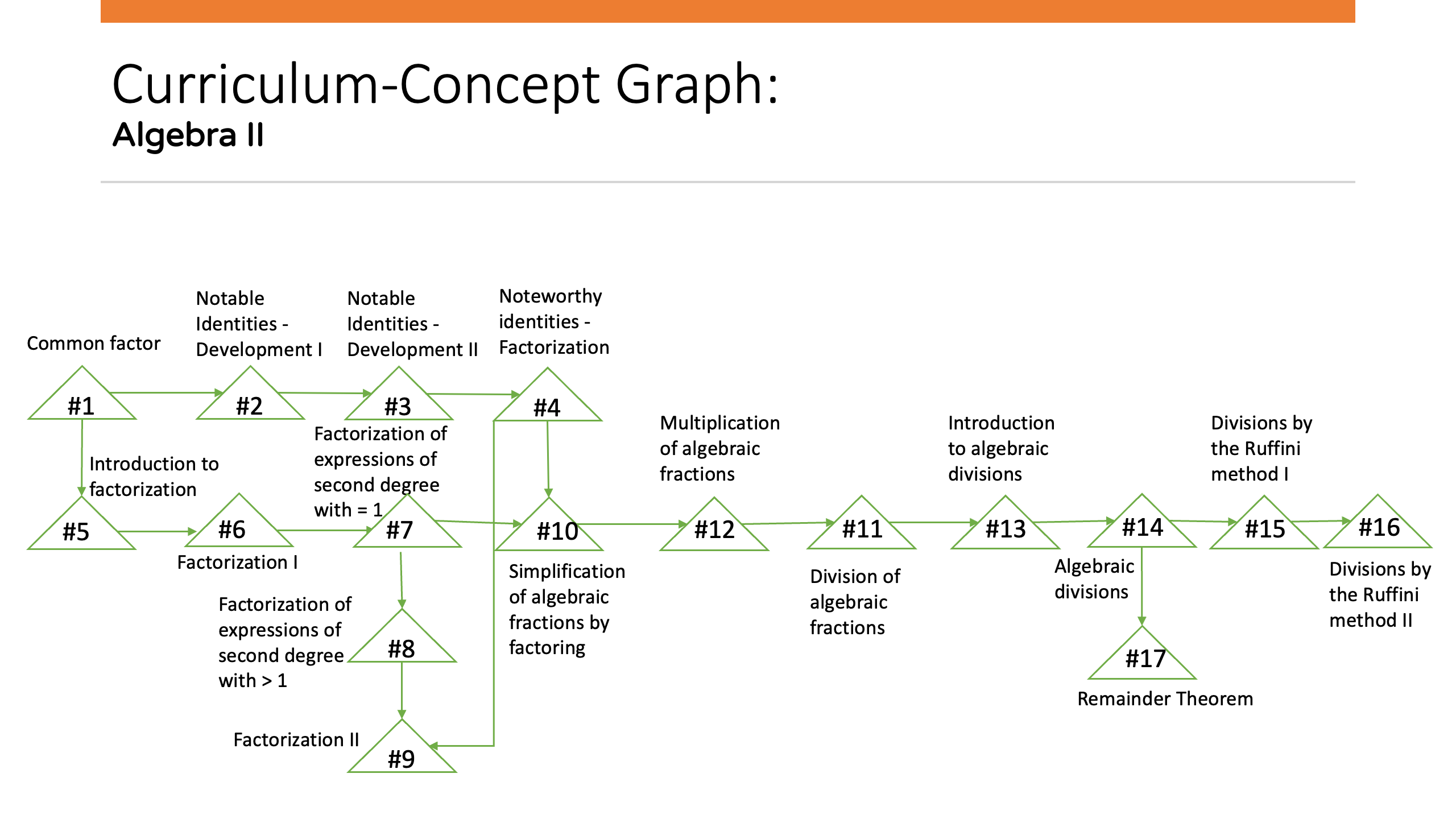}
         \caption{Curriculum-Structure Graph for the Topic \textit{Algebra II}}
\label{fig:curriculum_graph:algebraII}
\end{minipage}
\begin{minipage}[b]{.4\textwidth}
         \centering
         \includegraphics[width=.5\textwidth, 
         height=3cm
         ,trim={0cm 1cm 20cm 2cm},clip
         ]
        {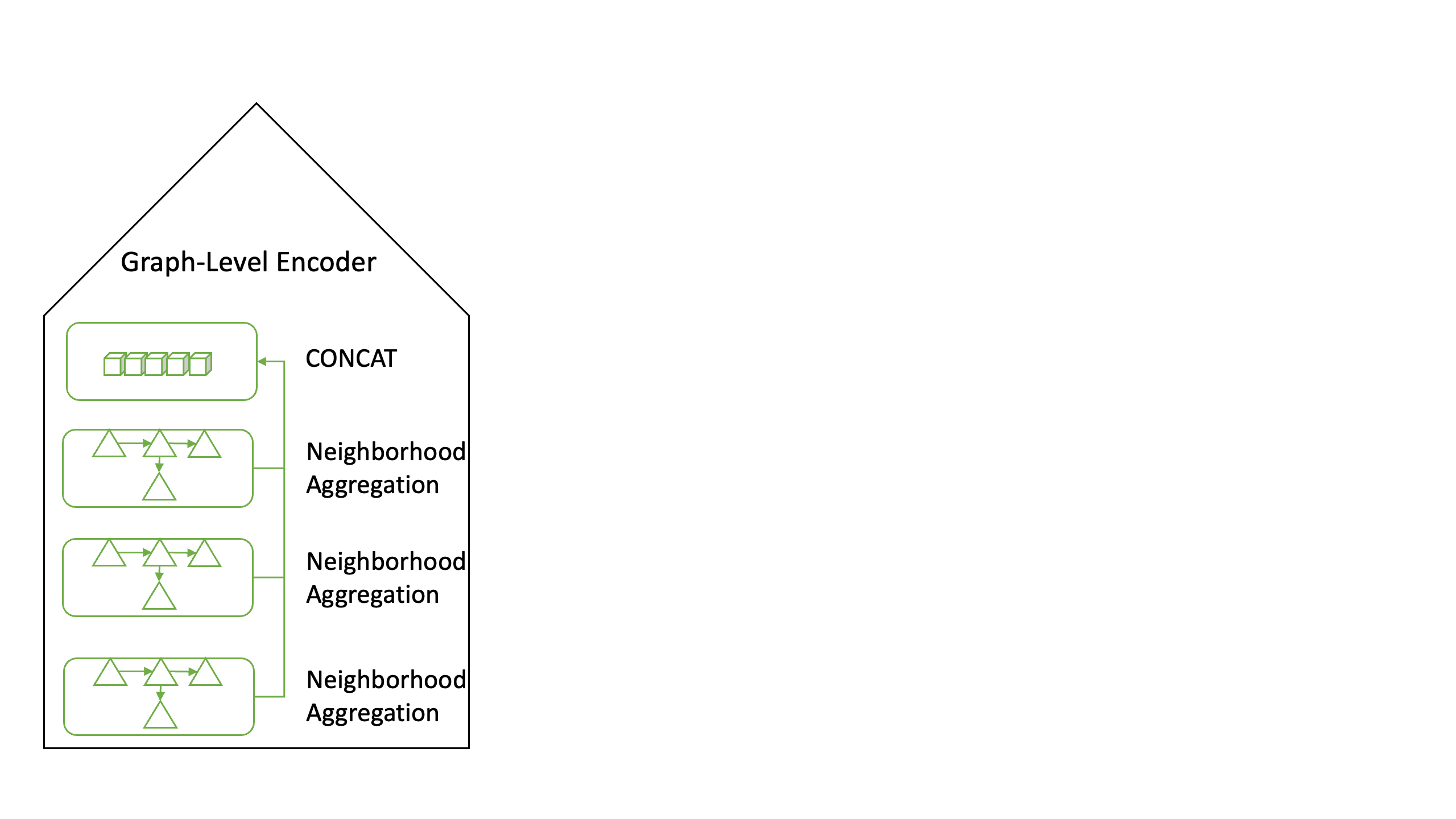}
         \caption{Graph-Level Encoder Architecture}
\label{fig:graph_level_encoder}
\end{minipage}\hfill
\begin{minipage}[b]{.6\textwidth}
         \centering
         \includegraphics[width=\textwidth, 
         height=3cm
         ,trim={0cm 3cm 0cm 5cm},clip
         ]
                {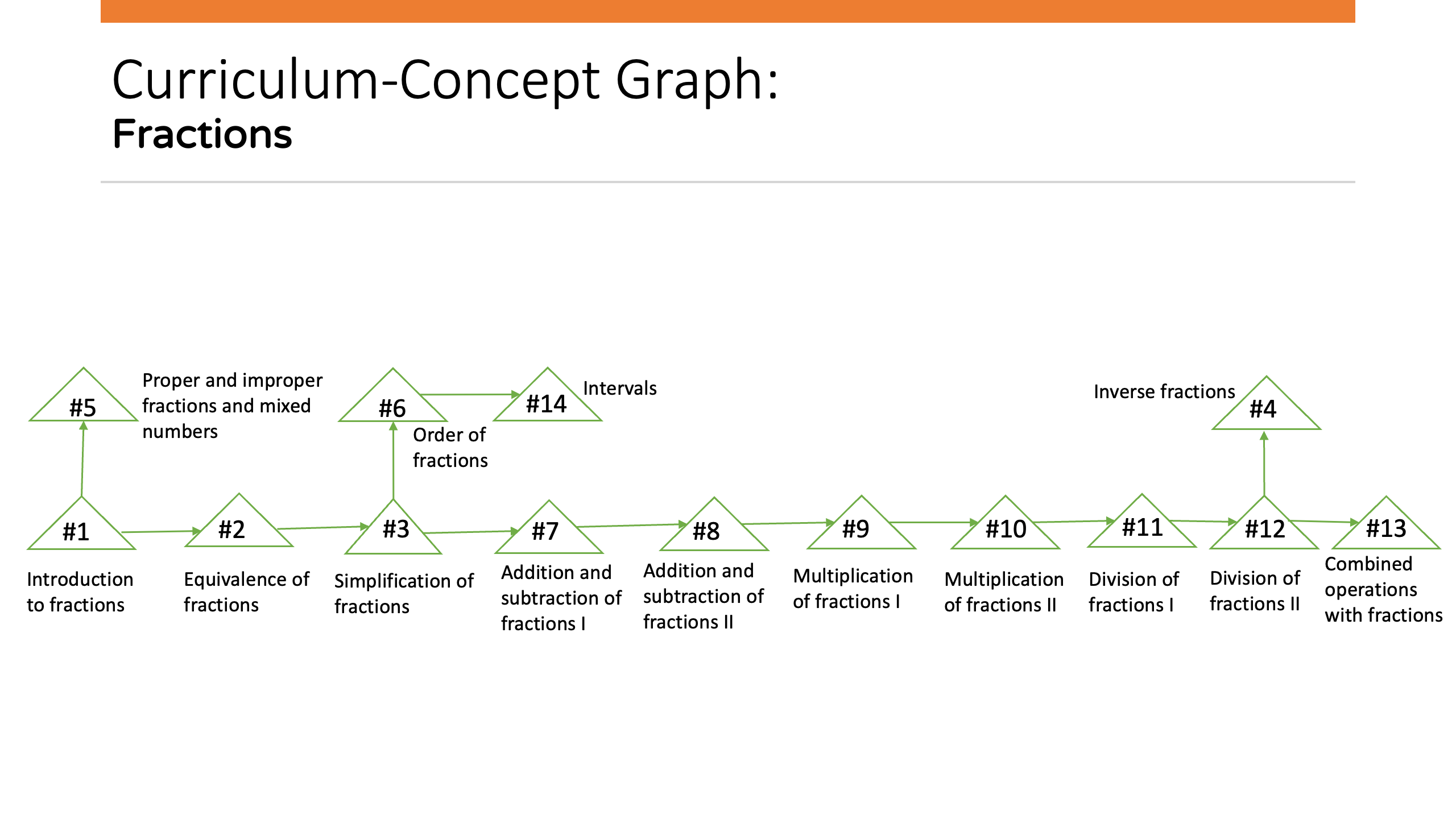}

         \caption{Curriculum-Structure Graph for the Topic \textit{Fractions}}
\label{fig:curriculum_graph:fractions}
\end{minipage}
\end{figure*}

In our approach, we utilize \textit{InfoGraph} to obtain the graph-level representation by maximizing the mutual information between the global representation and the local patch representation. Next, we will illustrate each component of the model separately.

\noindent\textbf{Graph-Level Encoder.}
We choose Graph Isomorphism Network (GIN) \cite{xu2018gin} as the graph convolution encoder in the model. The key distinguishing feature of GIN is that it can differentiate graphs that are not isomorphic to each other and hence achieve maximum discriminative power compared to other GNNs. The details of the graph-level encoder are illustrated in Figure \ref{fig:graph_level_encoder}.

\noindent\textbf{Patch Representation.}
To obtain the local patch representation, GIN passes message iteratively by updating each concept node's representation with aggregations of its neighbors' representations. After $k$ iterations, the patch representation captures the rich information within its $k$-hop neighborhood subgraph. Specifically, the update function and the aggregation scheme of the $k$-th layer in the framework of GIN are as follows:
\begin{align*} 
a_{v_i}^{(k)} &= \text{AGGREGATE}^{(k)}(h_{v_j}^{(k-1)}: {v_j} \in \mathcal{N}(v_i))),\\
h_{v_i}^{(k)} &= \text{COMBINE}((h_{v_i}^{(k-1)}, a_{v_i}^{(k)})
\end{align*} 
where $a_{v_i}^{(k)}$ is the aggregated representation of the concept node $v_i$'s neighbors $\mathcal{N}(v_i)$ and $h_{v_i}^{(k)}$ is the patch representation of node $v_i$ after the $k$-th layer in the graph encoder. $h_{v_i}^0$ is initialized with the multivariate learning-tracing feature vector.

\noindent\textbf{Global Representation.}
We utilize a READOUT function that summarizes all concept nodes' patch representations into a fixed-length vector to obtain the graph-level global representation. That is,
\begin{align*} 
h_\psi^{v_i} &= \text{CONCAT}(\{h_{v_i}^{(k)}\}_{k=1}^{K}),\\
H_\psi(G_\phi^u)&= \text{READOUT}(\{h_\psi^{v_i}\}_{i=1}^{N})
\end{align*} 
where $\psi$ denotes the set of parameters of a $K$-layer graph neural network, $h_\psi^{v_i}$ is the aggregated patch representation summarizing the neighborhood subgraphs centred at the concept node $v_i$, and $H_\psi(G_\phi^u)$ is the global graph-level representation for the  student curriculum-based learning graph $G_\phi^u$. In this study, we use sum over all the $N$ concept nodes' patch representations as the READOUT function.

\noindent\textbf{Contrastive Learning.}
\textit{InfoGraph} maximizes the mutual information (MI) between global and local representations, which is estimated with the discriminator $\mathcal{T_\beta}$ using contrastive learning. $\mathcal{T_\beta}$ is modeled by a feedforward neural network with parameters $\beta$. To estimate the maximized MI in practice, we use all possible combinations of global and local patch representations to generate negative samples across all graph instances for contrastive learning, as indicated in \textit{InfoGraph} \cite{sun2019infograph}. The MI estimator, $I_{\psi, \beta}$ maximizes the estimated MI on all the global/local pairs over the given dataset 
$\mathbb{G_\phi}$. Formally, for each topic $\phi$, the objective is to maximize $I_{\psi, \beta}$ as below:
\begin{align*} 
\hat{\psi}, \hat{\beta} &= \argmax_{\psi, \beta} 
\sum_{G_{\phi}^u \in \mathbb{G_\phi} }
\frac{1}{|G_{\phi}^u|} 
\sum_{v_i \in G_{\phi}^u}
I_{\psi, \beta} (h_\psi^{v_i}; H_\psi(G_\phi^u))
\end{align*}


\section{Experiments}
In this section, we discuss our analysis and findings of the graph-level representations learned by \texttt{CTGraph}.

\subsection{Datasets}
We conduct experiments on a real-world dataset provided by \texttt{Adaptemy}\footnote{https://www.adaptemy.com/}. This dataset was collected from \texttt{Adaptemy}'s ITS, which aids students in curriculum-based mathematical learning.  

The full curriculum comprises 26 topics in total. We present the results for four primary topics in this curriculum: \textit{Algebra I}, \textit{Algebra II}, \textit{Functions I} and \textit{Fractions}. The curriculum structure for each topic is illustrated in \cref{fig:curriculum_graph:algebraI,fig:curriculum_graph:algebraII,fig:running:curriculum:functionsI,fig:curriculum_graph:fractions}. 
As shown, each topic contains concepts (i.e., knowledge items) and the directed edges represent prerequisite conditions between concepts. A concept is a unit of learning. In the ITS, students can access each concept independently. Therefore, when engaging with the ITS, they are not required to follow a predetermined learning path as depicted in the curriculum. 

From the original data records of the learning traces, we extract the tracing attributes related to student behaviors and performance. These attributes include the student's average accuracy for each concept within the topic (i.e., the mean value of scores for all attempted questions covering each concept), the number of attempts (i.e., the total count of times a student attempted the questions on each concept), and the median week number in the academic calendar year when the student studied each concept. It is worthy noting that, while the attributes used in this study are simple aggregated values based on learning traces, more sophisticated attributes, such as temporal features and concept embedding, can also be incorporated  without altering the modeling framework.

Details of the datasets can be found in Table \ref{tab:data}.

\begin{table}[htb]
\centering
\begin{filecontents*}{sections/tables/dataset.csv}
Topic, Total Students, Active Students, # Concepts,# Interactions
Algebra I,9689,1415,12,65675
Algebra II,4106,290,17,32697
Functions I,4648,419,7,24153
Fractions,9243,867,14,63632
\end{filecontents*}
\csvreader[
    before reading=\footnotesize
        \caption{Dataset Statistics}\label{tab:algebraI}
          \setlength{\tabcolsep}{2.5pt},
        tabular={|>{\centering\arraybackslash}m{0.15\linewidth}
        |>{\centering\arraybackslash}m{0.15\linewidth} 
        |>{\centering\arraybackslash}m{0.25\linewidth} 
        |>{\centering\arraybackslash}m{0.15\linewidth} 
        |>{\centering\arraybackslash}m{0.18\linewidth} 
        |
        },
    table head =\hline Topic & \# Students 
    & \#  Students with Medium-High Concept Coverage
    & \# Concepts & \# Interactions with Questions\\\hline\hline,
    late after line= \\,
    late after last line=\\\hline
    ]{sections/tables/dataset.csv}{}
    {\csvcoli & \csvcolii & \csvcoliii & \csvcoliv & \csvcolv}
\label{tab:data}
\end{table}

\subsection{Experiments Configuration}
For each topic, we select students who covered at least 50\% of the concepts, indicating medium to high concept coverage attainment in the curriculum structure. The graph-level encoder used in the experiments consists of three hidden layers, each with a hidden dimension of 32. As a result, the global graph-level representation has a dimensionality of 96. We trained all models using the Adam optimizer with a learning rate of 0.01 and a batch size of 128.

\subsection{Analysis}
\begin{figure*}[th!]
\captionsetup{justification=centering}
     \begin{subfigure}[b]{0.3\textwidth}
         \centering
         \includegraphics[width=\textwidth, 
         height=4cm
         ,trim={0cm 0cm 0cm 0cm},clip
         ]
        {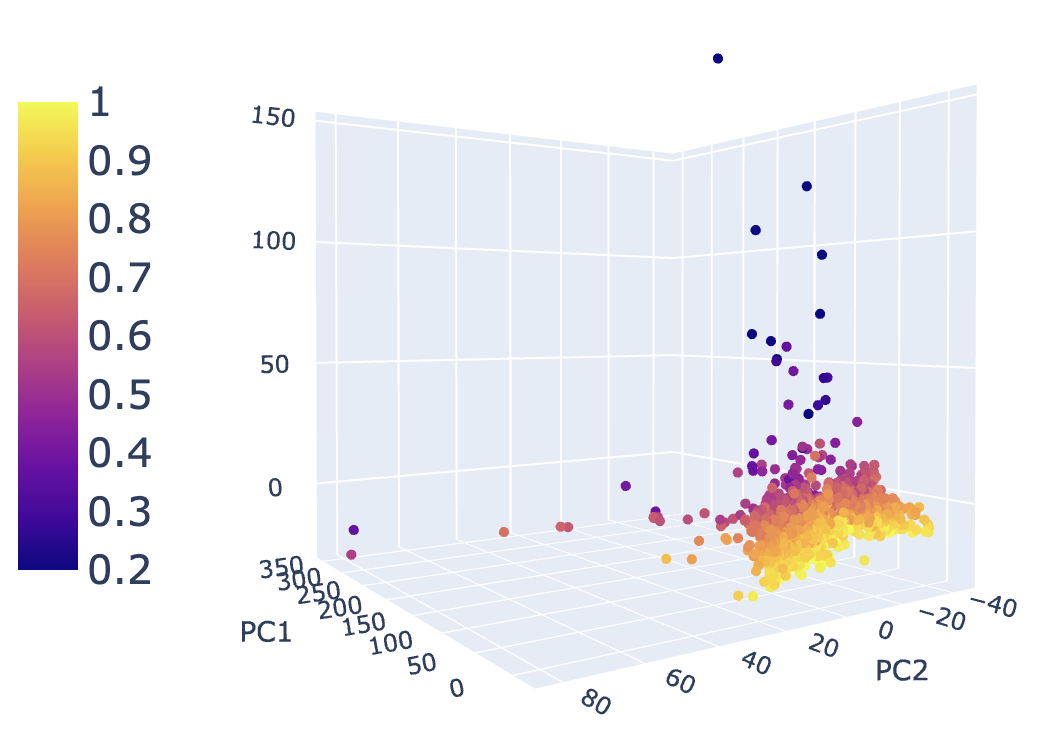}
         \caption{\textit{Algebra I} - Average Accuracy}
         \label{fig:latent:algebraI:acc}
     \end{subfigure}
     \hfill
     \begin{subfigure}[b]{0.3\textwidth}
         \centering
         \includegraphics[width=\textwidth, 
         height=4cm
         ,trim={0cm 0cm 0cm 0cm},clip
         ]
        {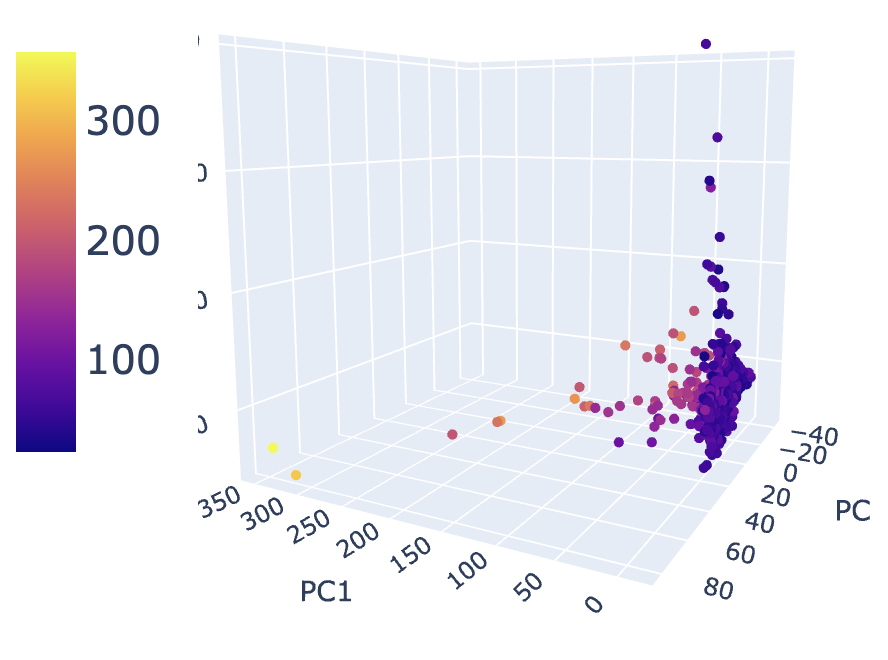}
         \caption{\textit{Algebra I} - Number of Attempts}
         \label{fig:latent:algebraI:sum_exercises}
     \end{subfigure}
    \hfill
     \begin{subfigure}[b]{0.3\textwidth}
         \centering
         \includegraphics[width=\textwidth, 
         height=4cm
         ,trim={0cm 0cm 0cm 0cm},clip
         ]
        {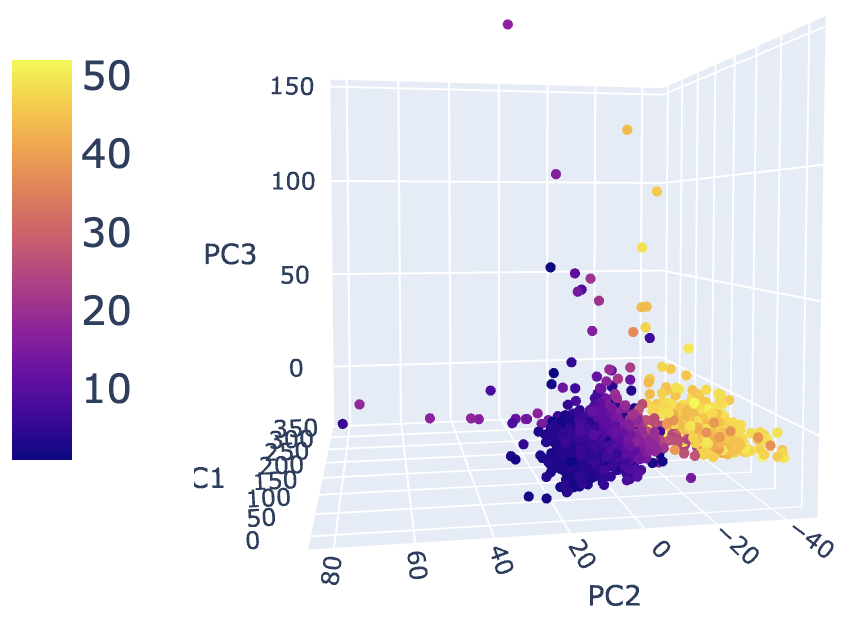}
         \caption{\textit{Algebra I} - Median Week Number}
         \label{fig:latent:algebraI:median_week}
     \end{subfigure}

         \caption{PCA Representations in the Latent Space for the Topic \textit{Algebra I}
        }
        \label{fig:latent:algebraI}
\end{figure*}

\begin{figure*}[h!]
\captionsetup{justification=centering}
     \begin{subfigure}[b]{0.3\textwidth}
         \centering
         \includegraphics[width=\textwidth, 
         height=4cm
         ,trim={0cm 0cm 0cm 0cm},clip
         ]
        {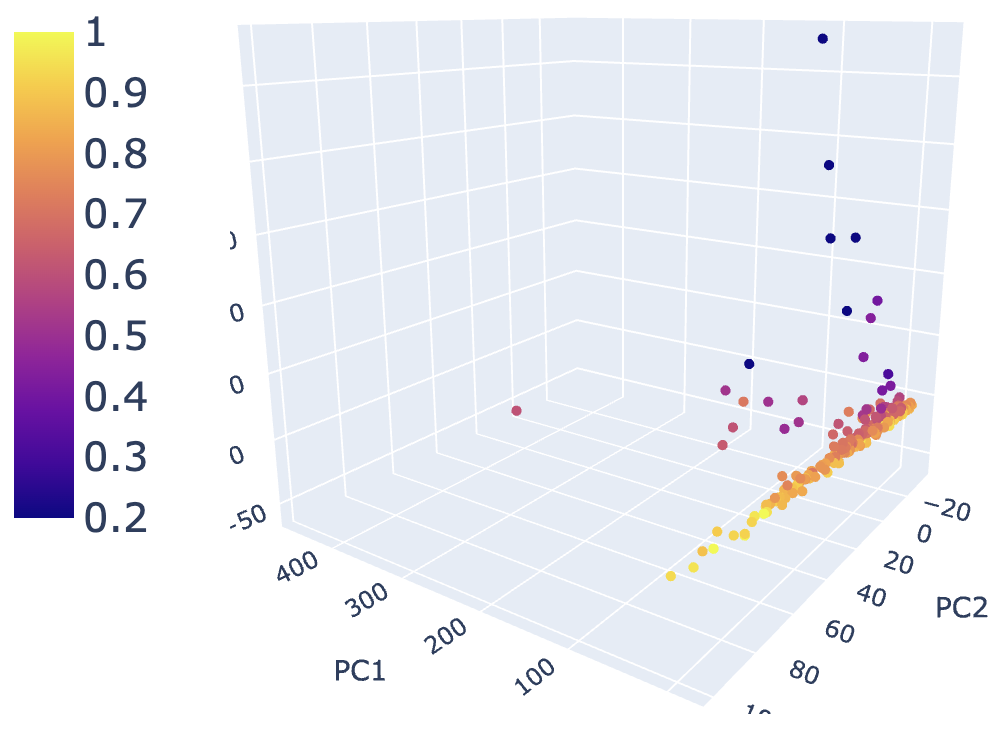}
         \caption{\textit{Algebra II} - Average Accuracy}
         \label{fig:latent:algebraII:acc}
     \end{subfigure}
     \hfill
     \begin{subfigure}[b]{0.3\textwidth}
         \centering
         \includegraphics[width=\textwidth, 
         height=4cm
         ,trim={0cm 0cm 0cm 0cm},clip
         ]
        {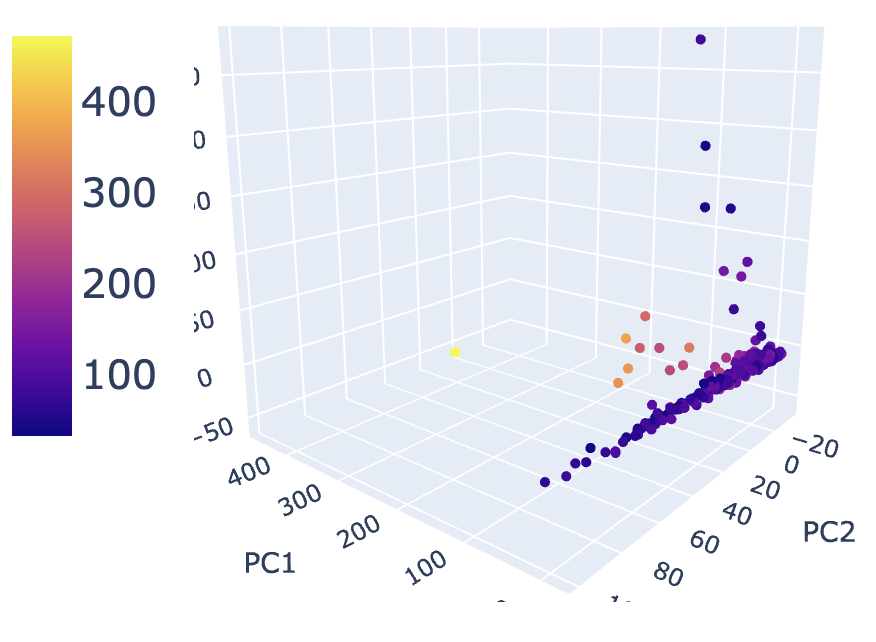}
         \caption{\textit{Algebra II} - Number of Attempts}
         \label{fig:latent:algebraII:sum_exercises}
     \end{subfigure}
    \hfill
     \begin{subfigure}[b]{0.3\textwidth}
         \centering
         \includegraphics[width=\textwidth, 
         height=4cm
         ,trim={0cm 0cm 0cm 0cm},clip
         ]
        {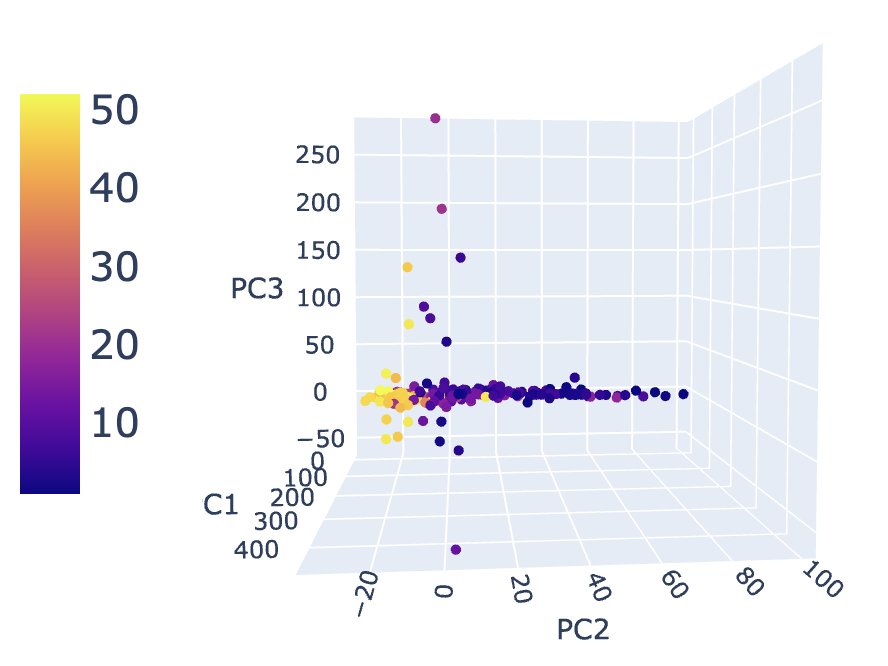}
         \caption{\textit{Algebra II} - Median Week Number}
         \label{fig:latent:algebraII:median_week}
     \end{subfigure}
         \caption{PCA Representations in the Latent Space for the Topic \textit{Algebra II}
        }
        \label{fig:latent:algebraII}
\end{figure*}

\begin{figure*}[h!]
\textbf{\captionsetup{justification=centering}
}
          \begin{subfigure}[b]{0.3\textwidth}
         \centering
         \includegraphics[width=\textwidth, 
         height=4cm
         ,trim={0cm 0cm 0cm 0cm},clip
         ]
        {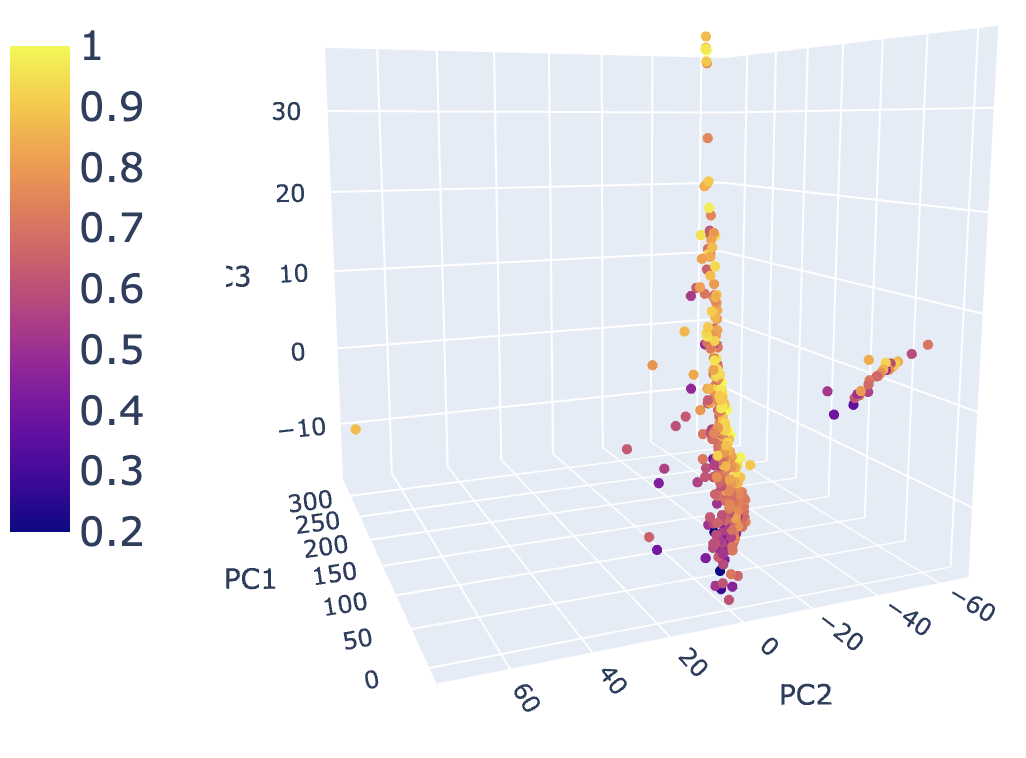}
         \caption{\textit{Functions I} - Average Accuracy}
         \label{fig:latent:function:acc}
     \end{subfigure}
     \hfill
          \begin{subfigure}[b]{0.3\textwidth}
         \centering
         \includegraphics[width=\textwidth, 
         height=4cm
         ,trim={0cm 0cm 0cm 0cm},clip
         ]
        {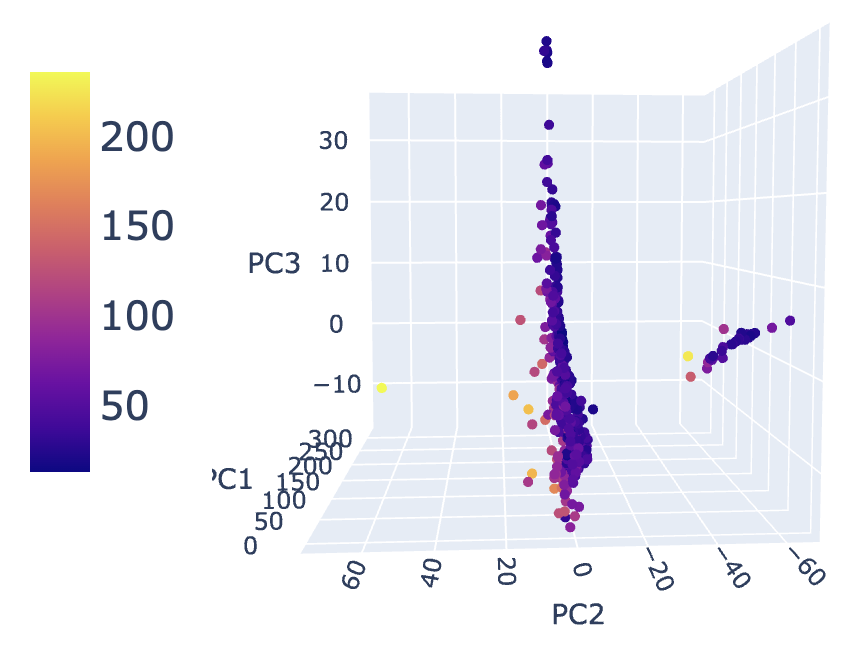}
         \caption{\textit{Functions I} - Number of Attempts}
         \label{fig:latent:function:sum_exercises}
     \end{subfigure}
    \hfill
          \begin{subfigure}[b]{0.3\textwidth}
         \centering
         \includegraphics[width=\textwidth, 
         height=4cm
         ,trim={0cm 0cm 0cm 0cm},clip
         ]
        {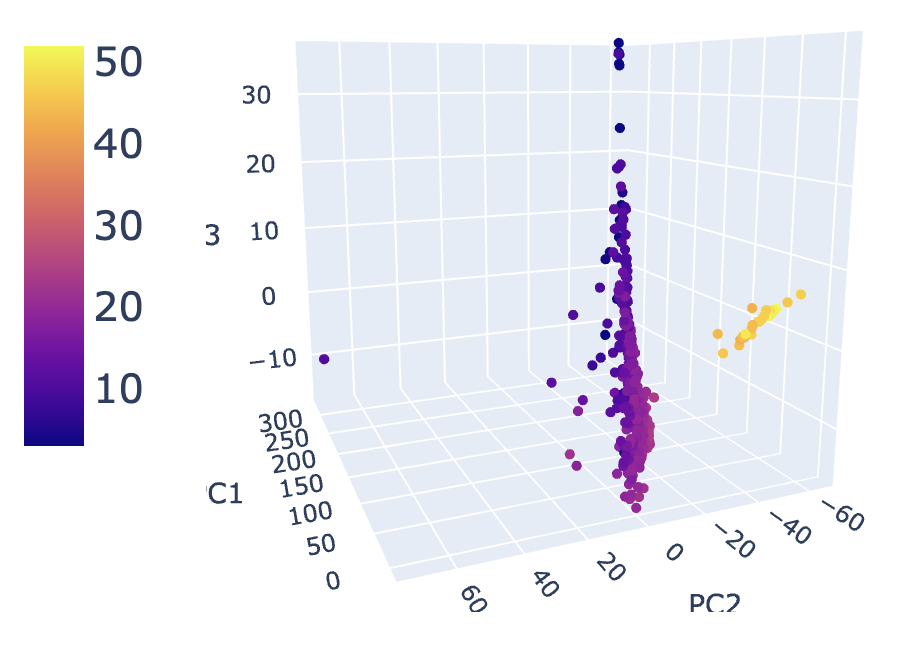}
         \caption{\textit{Functions I} - Median Week Number}
         \label{fig:latent:function:median_week}
     \end{subfigure}
         \caption{PCA Representations in the Latent Space for the Topic \textit{Functions I}
        }
        \label{fig:latent:functionsI}
\end{figure*}

\begin{figure*}[h!]
\captionsetup{justification=centering}
        \begin{subfigure}[b]{0.3\textwidth}
         \centering
         \includegraphics[width=\textwidth, 
         height=4cm
         ,trim={0cm 0cm 0cm 0cm},clip
         ]
        {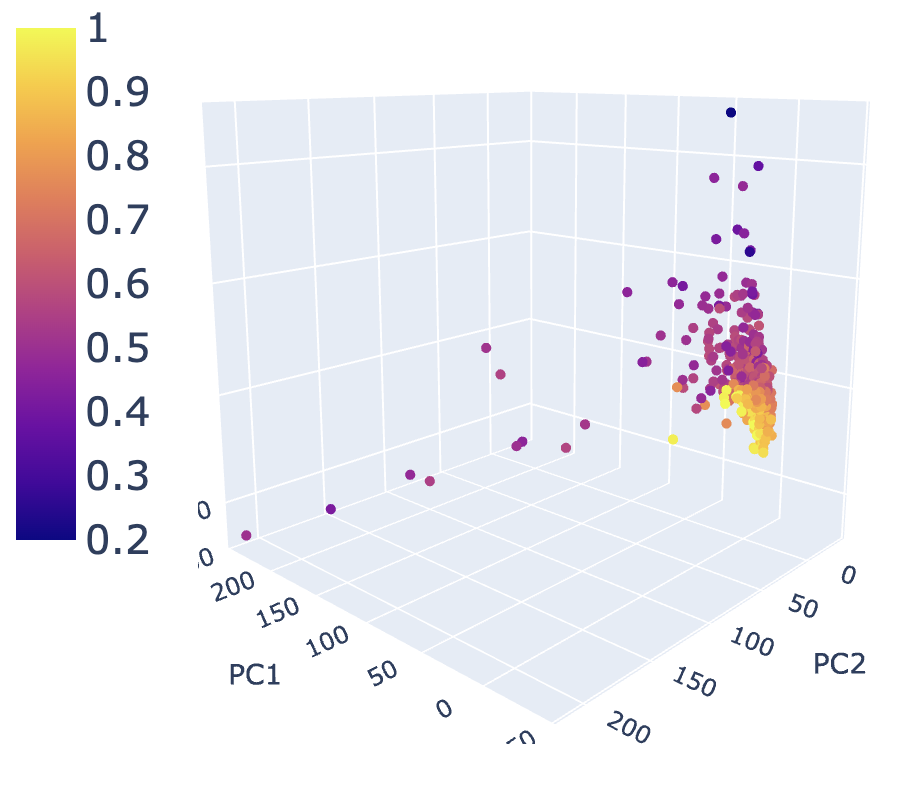}
         \caption{\textit{Fractions} - Average Accuracy}
         \label{fig:latent:fracciones:acc}
     \end{subfigure}
     \hfill
               \begin{subfigure}[b]{0.3\textwidth}
         \centering
         \includegraphics[width=\textwidth, 
         height=4cm
         ,trim={0cm 0cm 0cm 0cm},clip
         ]
        {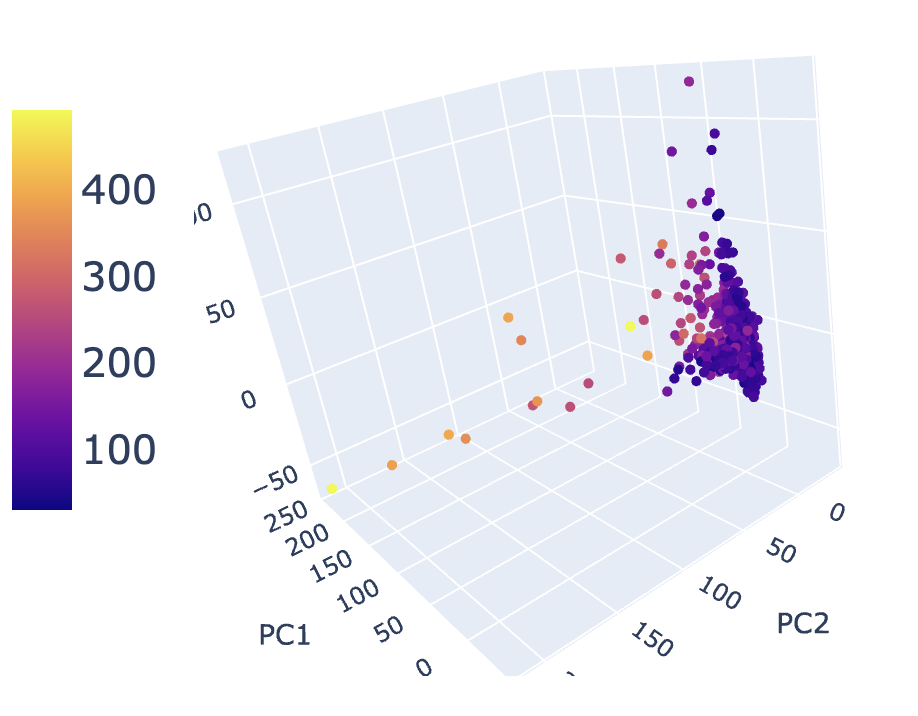}
         \caption{\textit{Fractions} - Number of Attempts}
         \label{fig:latent:fracciones:sum_exercises}
     \end{subfigure}
    \hfill
               \begin{subfigure}[b]{0.3\textwidth}
         \centering
         \includegraphics[width=\textwidth, 
         height=4cm
         ,trim={0cm 0cm 0cm 0cm},clip
         ]
        {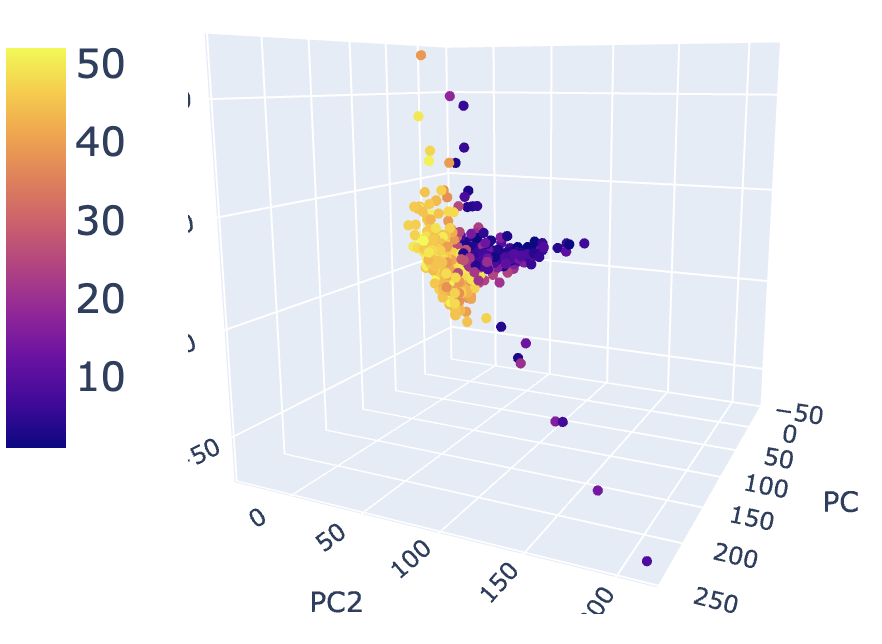}
         \caption{\textit{Fractions} - Median Week Number}
         \label{fig:latent:fracciones:median_week}
     \end{subfigure}
         \caption{PCA Representations in the Latent Space for the Topic \textit{Fractions}
        }
        \label{fig:latent:fractions}
\end{figure*}

In the following analysis,  we seek to answer the following questions:
\begin{enumerate}
\item[\textbullet]Q1: Can the representations in the latent space effectively encode a student's learning path, behaviors, and performance in the curriculum-based ITS? How can these representations be used to monitor student performance and behaviors, as well as identify those lagging behind?
\item[\textbullet]Q2: How can we use the representations to identify equivalent groups of similar learners, allowing educators to spot commonalities in the behaviors of students who are lagging behind?
\item[\textbullet]Q3: Can the representations capture the subtle differences in behaviors and learning paths among the students who are lagging behind? Will these differences reveal when and where the students began to lag behind in specific aspects? If so, how can we find students
with subtle differences in the latent space?
\end{enumerate}
\subsubsection{Representations in the Latent Space}
For each topic, we extract the latent representation in a 96-dimensional vector for each student. We then derive the 3D visualization by further compressing the latent representations to three dimensions using PCA compression. 

\cref{fig:latent:algebraI,fig:latent:algebraII,fig:latent:functionsI,fig:latent:fractions} display the 3D visualization of latent spaces for the four topics. As shown in the figures, most students cluster closely together in the latent space, while a small number of outliers are positioned further away from the central clusters. Notably,  in \cref{fig:latent:function:acc,fig:latent:function:sum_exercises,fig:latent:function:median_week} for the topic \textit{Functions I}, the distribution of latent representations clearly show two clusters.  This indicates the behaviors and performance of students in the smaller cluster differ significantly from the students in the larger cluster, as particularly evident in the median week number illustrated in Figure \ref{fig:latent:function:median_week}. 

Note that the extracted vector representation 
encompasses tracing attributes reflecting features of learning behaviors, such as proficiency, intensity, and the timing of when the student attempted questions covering certain concepts within the topic. To demonstrate that representations effectively encode behaviors and performance across different dimensions simultaneously, we use color in each figure to highlight the aggregated values of each tracing attribute for all the concepts in a student's curriculum-based learning graph. For example, in Figure \ref{fig:latent:algebraI:acc}, the color represents a student's average accuracy across all the attempts on all the concepts for the topic \textit{Algebra I}. In Figure \ref{fig:latent:algebraI:sum_exercises}, the color indicates the total number of attempts on all the concepts, while in Figure \ref{fig:latent:algebraI:median_week}, the color corresponds to the median of the median week numbers regarding all the concepts a student studied in this topic. 

It is important to note that, those aggregated values of tracing attributes are for illustration purposes only and are not used during model training. As clearly shown in the figures, color gradients in \cref{fig:latent:algebraI,fig:latent:algebraII,fig:latent:functionsI,fig:latent:fractions} represent the distribution of these three types of attributes in the latent space, providing evidence that the latent representations effectively encode these tracing attributes simultaneously.

The positioning of each student's representation in the latent space reveals a smooth distribution of different learning behaviors and performance (in terms of tracing attributes and learning paths). When we locate students in the latent space, it is evident that those clustered together achieved similar performance, exhibited comparable tracing attributes and followed similar learning paths. Specifically, in terms of performance, the students identified as outliers typically have an average accuracy in their attempts that is much lower than that of the majority. As such, we can identify students lagging behind by locating these outliers in the latent spaces.

\subsubsection{Similar Students}
\begin{figure*}[th!]
\captionsetup{justification=centering}
     \centering
     \begin{subfigure}[b]{0.3\textwidth}
         \centering
         \includegraphics[width=\textwidth, 
         height=4cm
         ,trim={0cm 0cm 1cm 0cm},clip
         ]
        {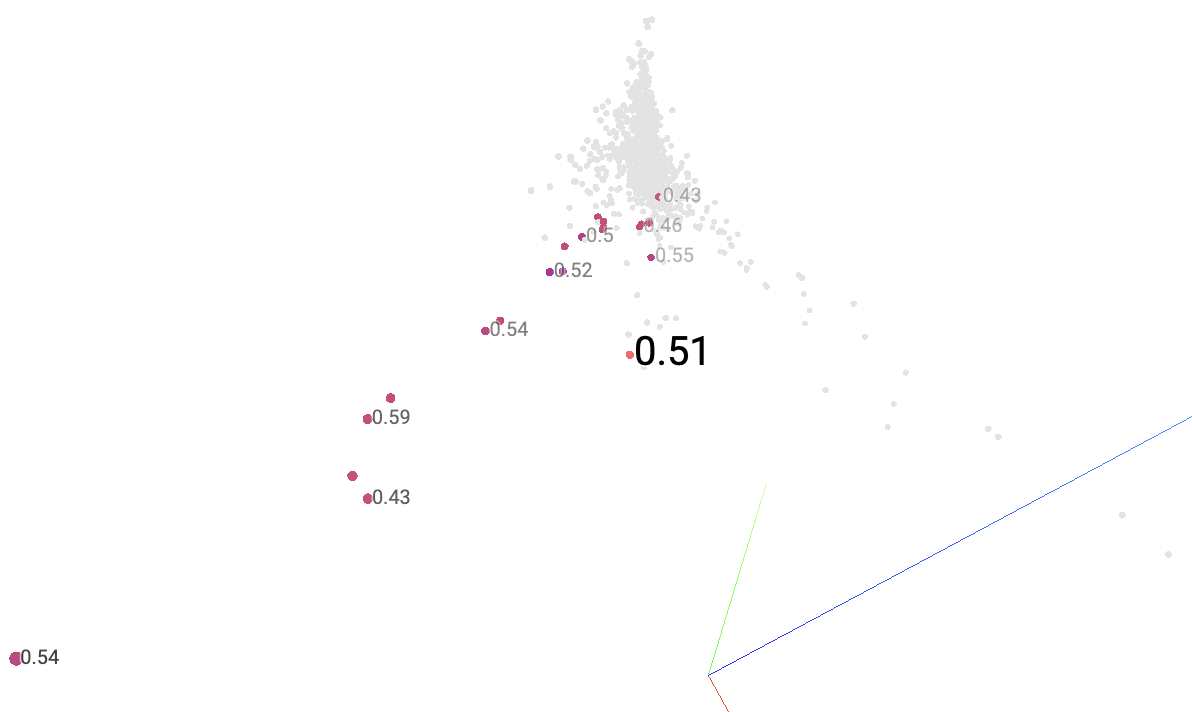}
         \caption{Average Accuracy,\\Top 20 Similar Students}
         \label{fig:latent:neigh:algebraI:accuracy:20}
     \end{subfigure}
     \hfill
     \begin{subfigure}[b]{0.3\textwidth}
         \centering
         \includegraphics[width=\textwidth, 
         height=4cm
         ,trim={0cm 0cm 1cm 0cm},clip
         ]
        {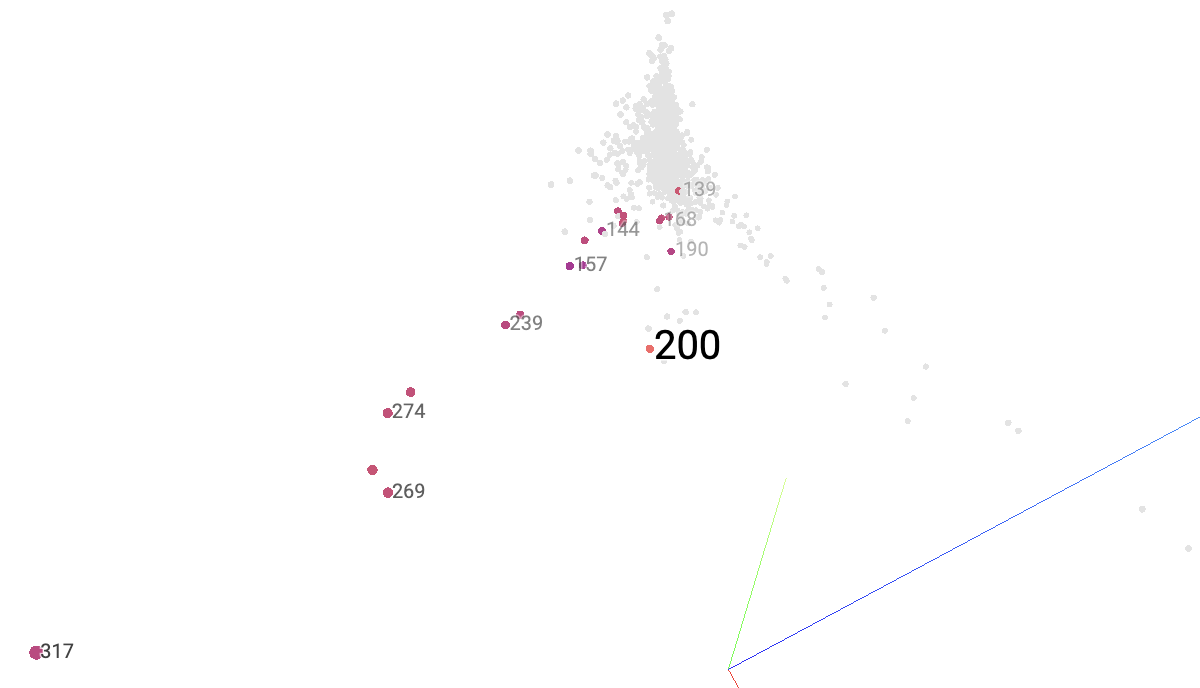}
         \caption{Number of Attempts,\\Top 20 Similar Students}
         \label{fig:latent:neigh:algebraI:attempts:20}
     \end{subfigure}
     \hfill
     \begin{subfigure}[b]{0.3\textwidth}
         \centering
         \includegraphics[width=\textwidth, 
         height=4cm
         ,trim={0cm 0cm 1cm 0cm},clip
         ]
        {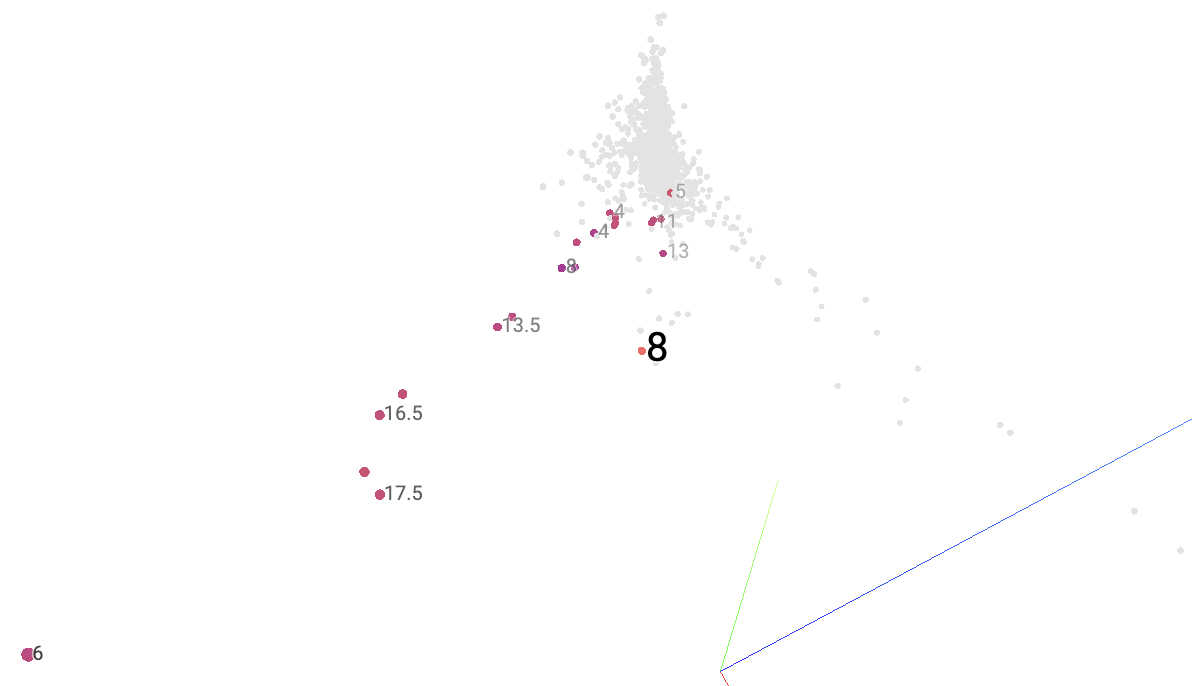}
         \caption{Median Week Number,\\Top 20 Similar Students}
         \label{fig:latent:neigh:algebraI:median:20}
     \end{subfigure}
     \hfill
     \begin{subfigure}[b]{0.3\textwidth}
         \centering
         \includegraphics[width=\textwidth, 
         height=4cm
         ,trim={0cm 0cm 1cm 0cm},clip
         ]
        {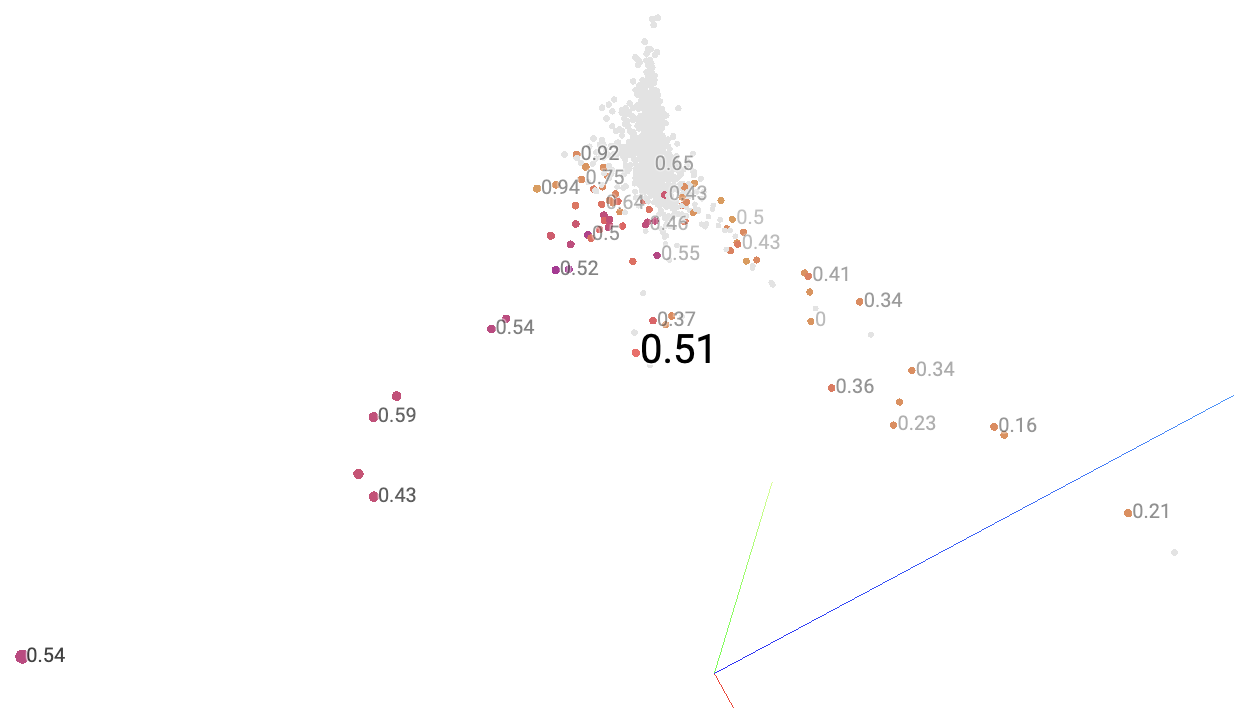}
         \caption{Average Accuracy,\\Top 80 Similar Students}
         \label{fig:latent:neigh:algebraI:accuracy:80}
     \end{subfigure}
     \hfill
     \begin{subfigure}[b]{0.32\textwidth}
         \centering
         \includegraphics[width=\textwidth, 
         height=4cm
         ,trim={0cm 0cm 1cm 0cm},clip
         ]
        {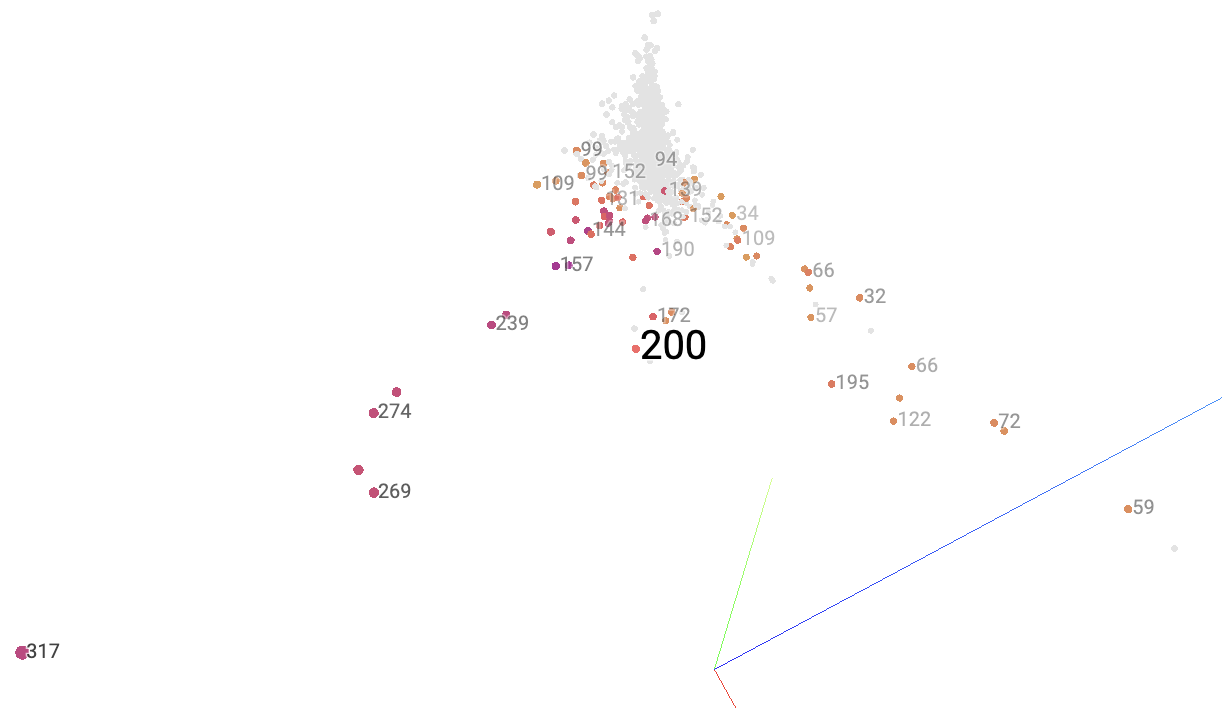}
         \caption{Number of Attempts,\\Top 80 Similar Students}
         \label{fig:latent:neigh:algebraI:attempts:80}
     \end{subfigure}
    \hfill
     \begin{subfigure}[b]{0.3\textwidth}
         \centering
         \includegraphics[width=\textwidth, 
         height=4cm
         ,trim={0cm 0cm 1cm 0cm},clip
         ]
        {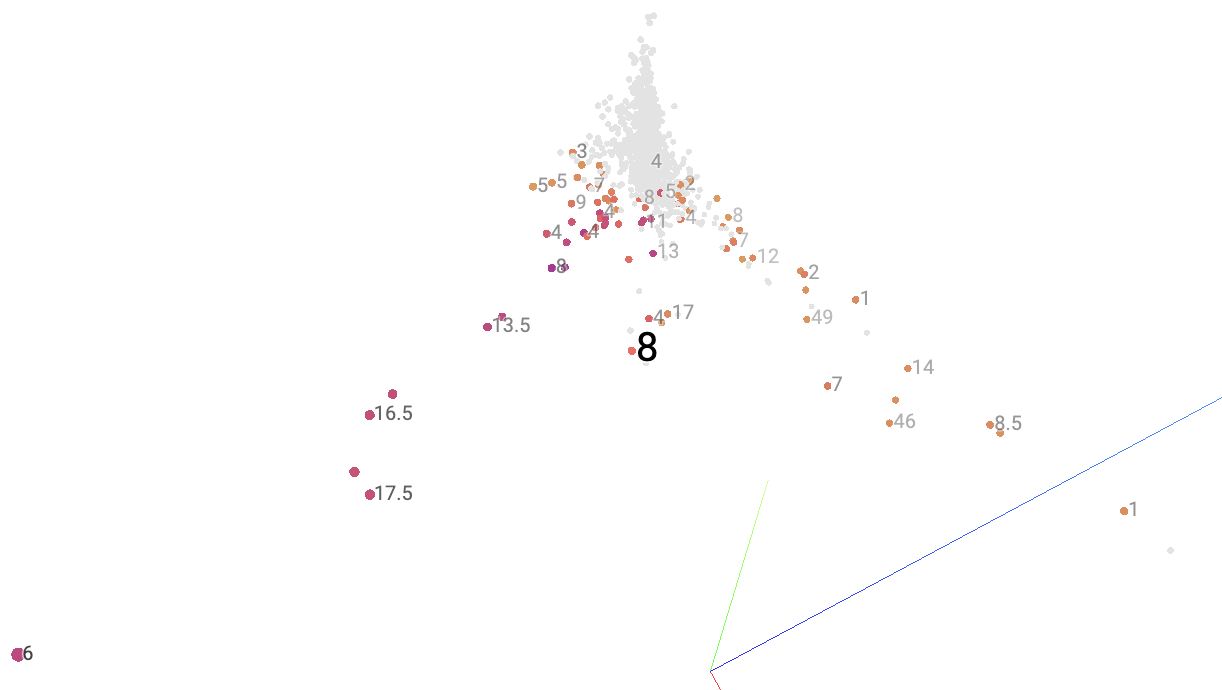}
         \caption{Median Week Number,\\Top 80 Similar Students}
         \label{fig:latent:neigh:algebraI:median:80}
     \end{subfigure}

         \caption{ Similar Students in the Latent Space
        for the Topic \textit{Algebra I}}
        \label{fig:latent:neigh}
\end{figure*}

\begin{figure*}[th!]
\captionsetup{justification=centering}
     \begin{subfigure}[b]{0.16\textwidth}
         \centering
         \includegraphics[width=\textwidth, 
         height=4cm
         ,trim={0cm 0cm 0cm 0cm},clip
         ]
        {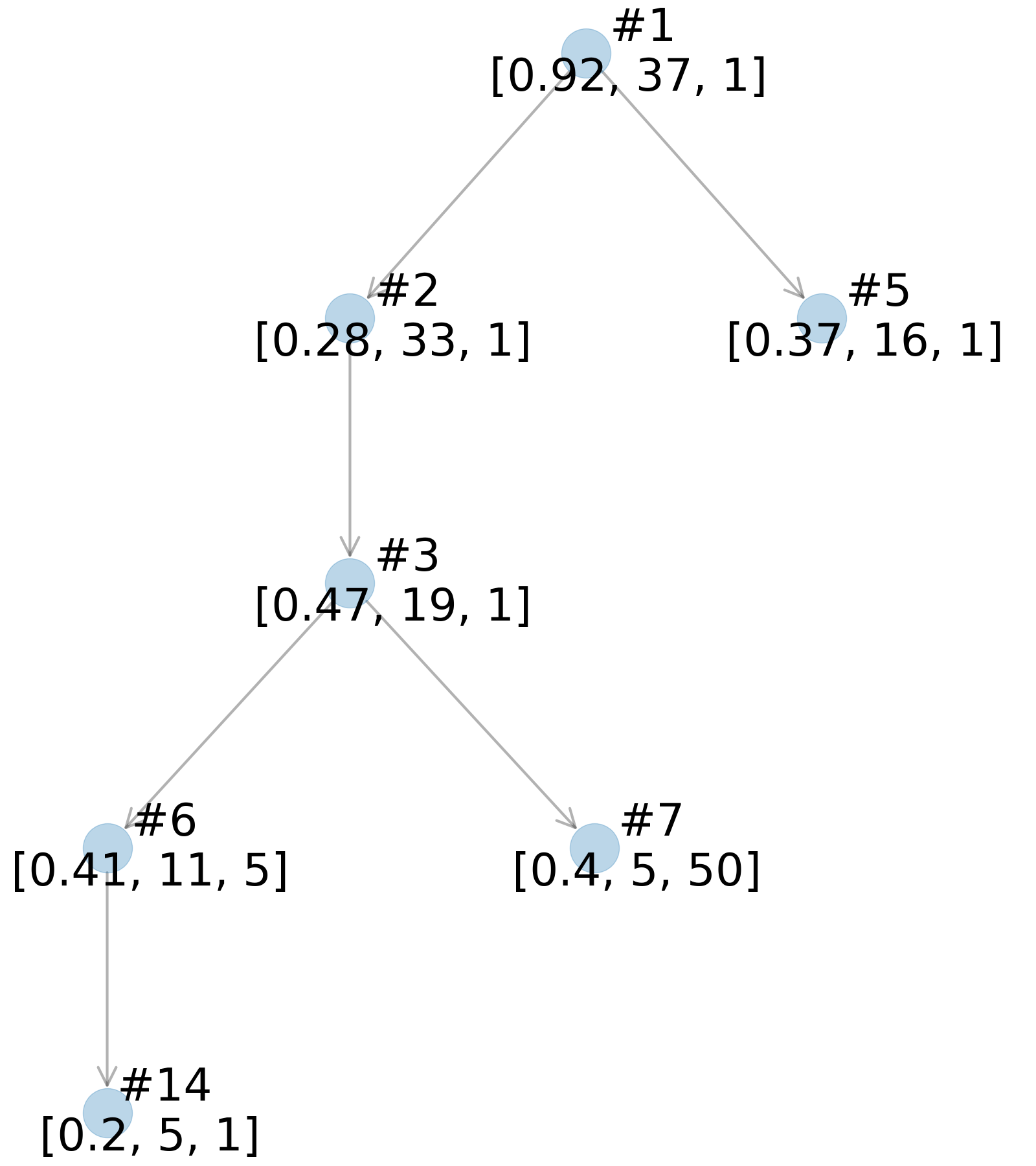}
         \caption{Student 1}
         \label{fig:neigh:fracciones:1}
     \end{subfigure}
     \begin{subfigure}[b]{0.16\textwidth}
         \centering
         \includegraphics[width=\textwidth, 
         height=4cm
         ,trim={0cm 0cm 0cm 0cm},clip
         ]
        {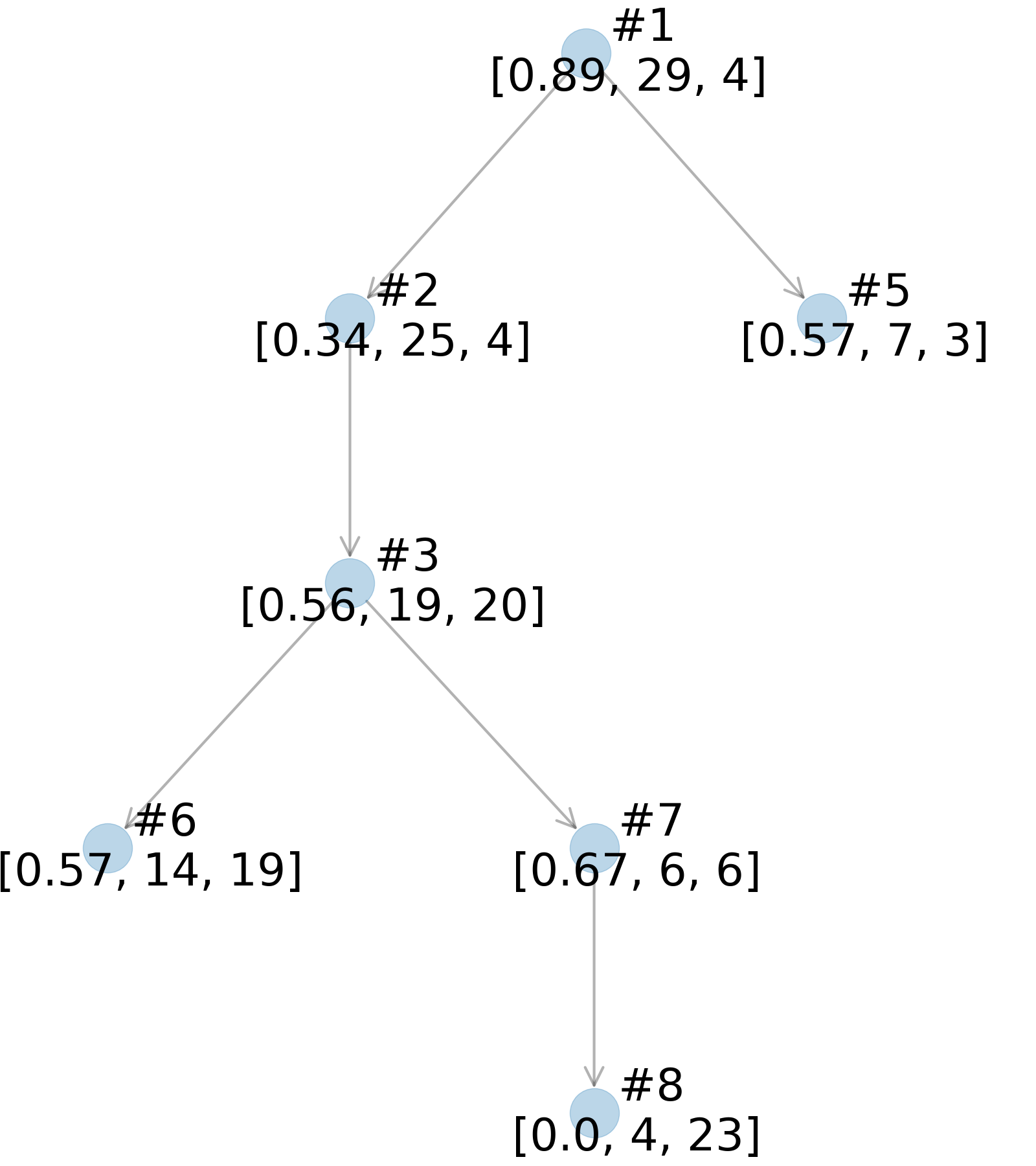}
         \caption{Student 2}
         \label{fig:neigh:fracciones:2}
     \end{subfigure}
          \begin{subfigure}[b]{0.16\textwidth}
         \centering
         
\includegraphics[width=\textwidth, 
         height=4cm
         ,trim={0cm 0cm 0cm 0cm},clip
         ]
        {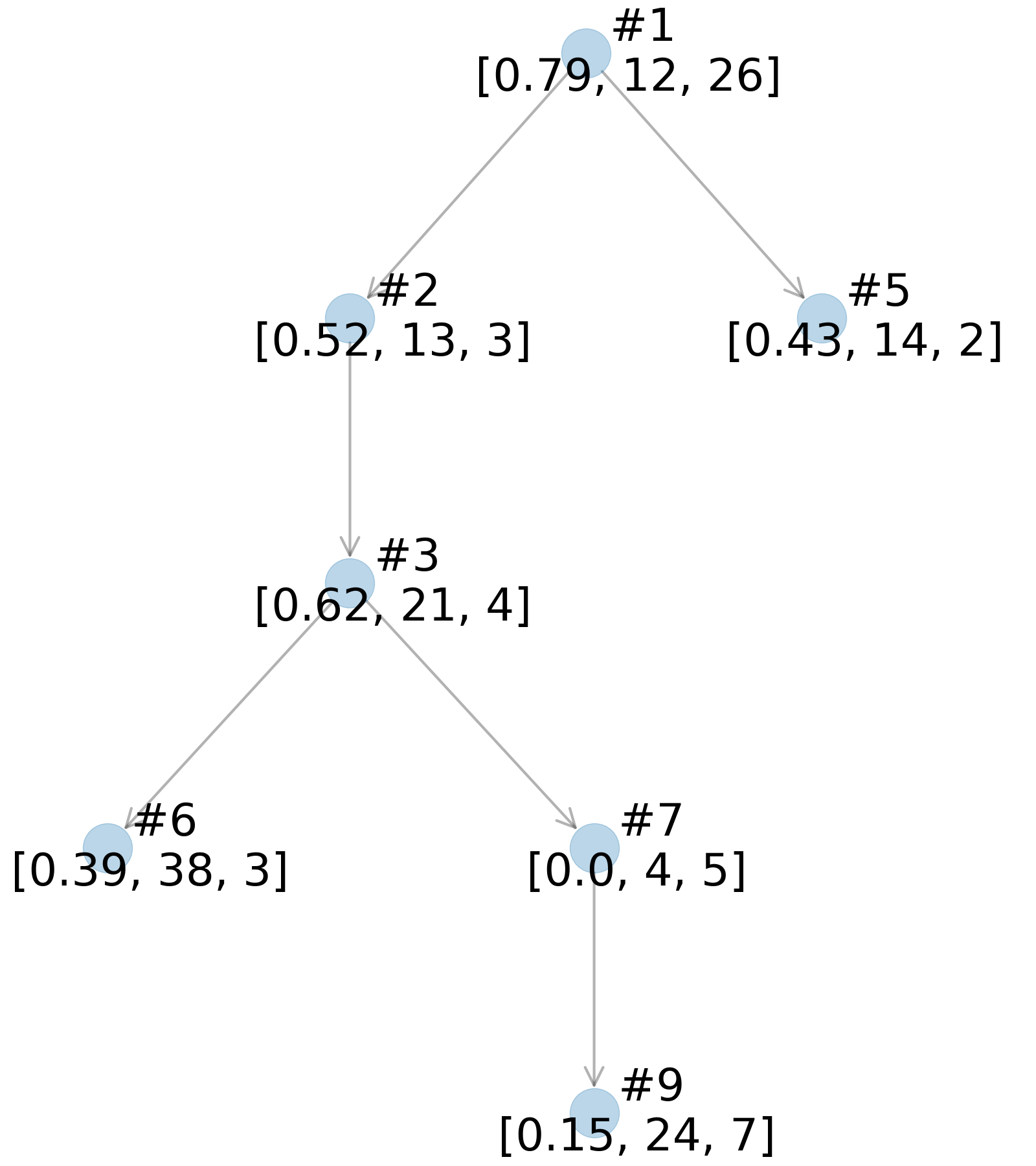}
         \caption{Student 3}
         \label{fig:neigh:fracciones:3}
     \end{subfigure}
          \begin{subfigure}[b]{0.16\textwidth}
         \centering
         
     \includegraphics[width=\textwidth, 
         height=4cm
         ,trim={0cm 0cm 0cm 0cm},clip
         ]
        {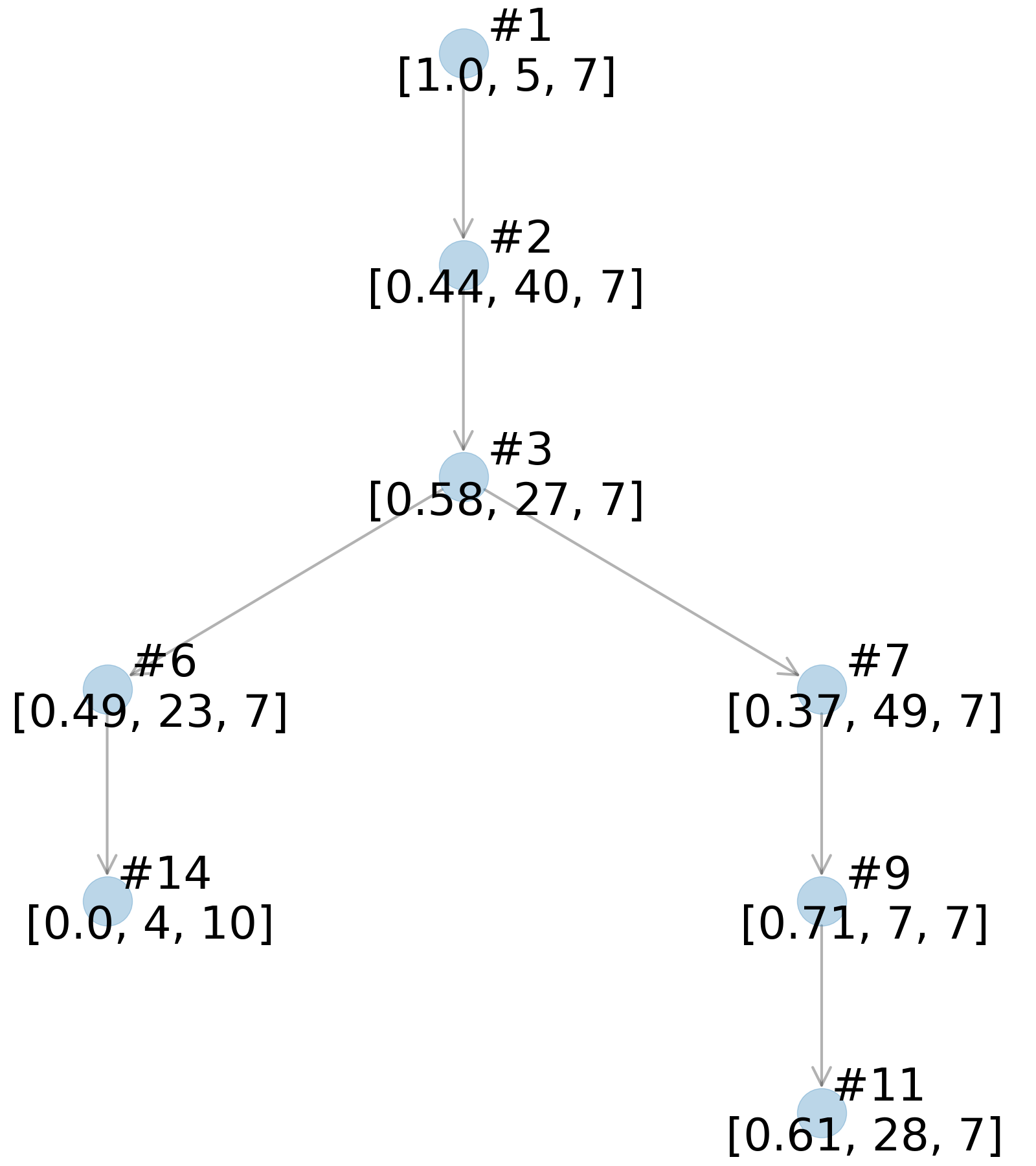}
         \caption{Student 4}
         \label{fig:neigh:fracciones:4}
     \end{subfigure}
          \begin{subfigure}[b]{0.16\textwidth}
         \centering
         
     \includegraphics[width=\textwidth, 
         height=4cm
         ,trim={0cm 0cm 0cm 0cm},clip
         ]
        {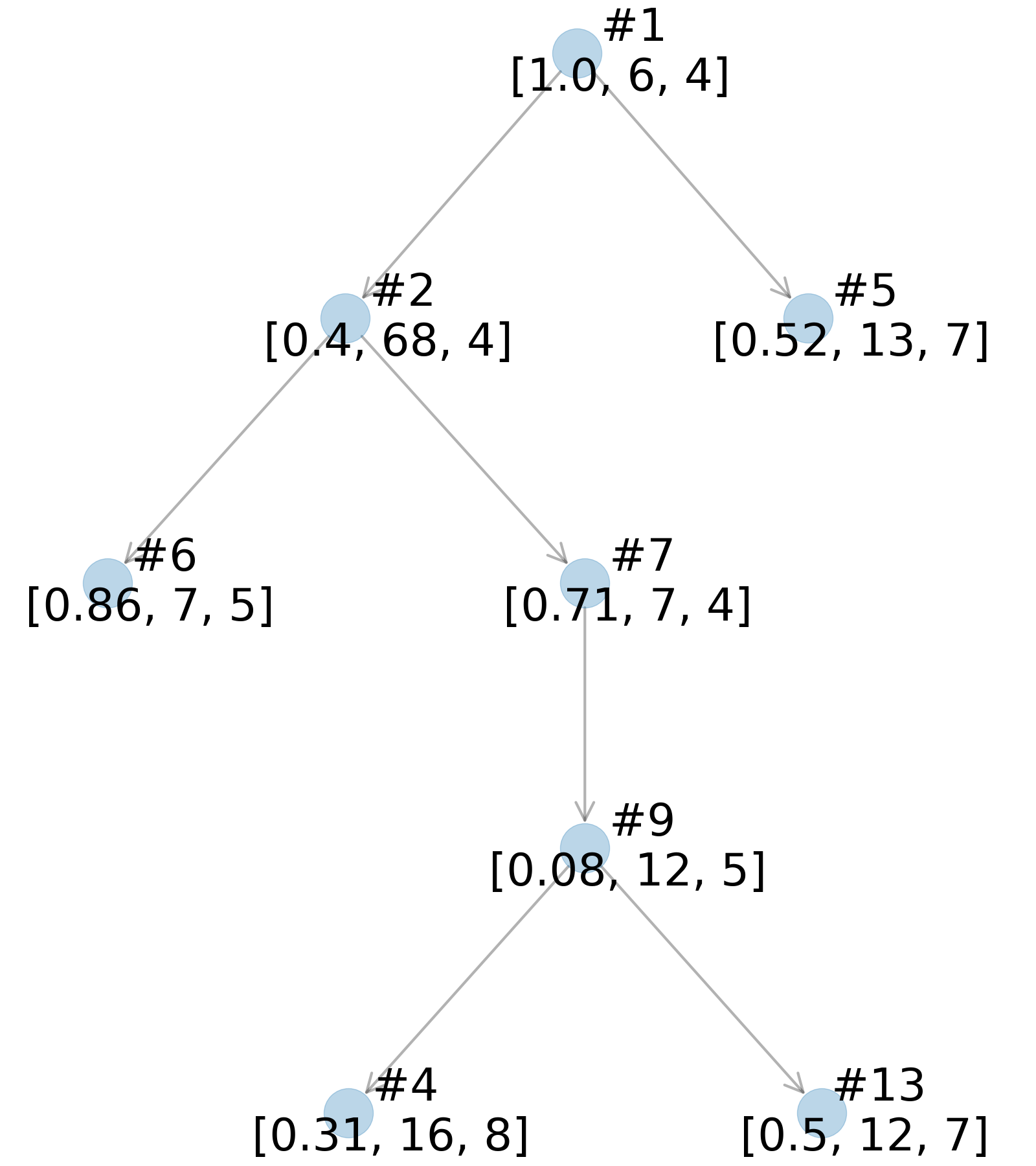}
         \caption{Student 5}
         \label{fig:neigh:fracciones:5}
     \end{subfigure}
               \begin{subfigure}[b]{0.16\textwidth}
         \centering
          \includegraphics[width=\textwidth, 
         height=4cm
         ,trim={0cm 0cm 0cm 0cm},clip
         ]
        {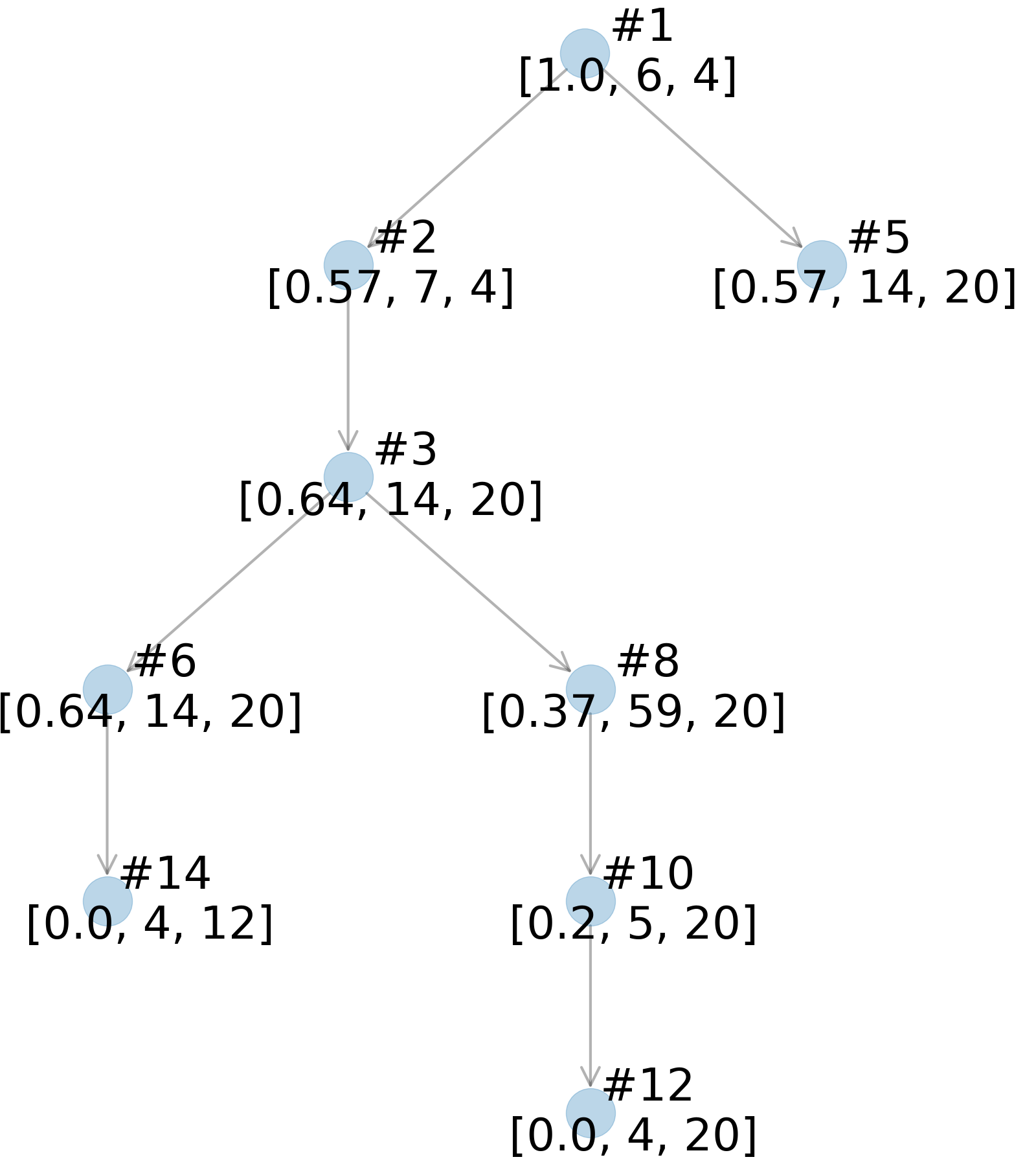}
         \caption{Student 6}
         \label{fig:neigh:fracciones:6}
     \end{subfigure}
         \caption{ Top 5 Neighboring Students of Student 1 in the Topic \textit{Fractions}
        }
        \label{fig:neigh}
\end{figure*}

Now we illustrate how to utilize the latent spaces from \texttt{CTGraph} for identifying similar students who are lagging behind at certain aspects. 

We use the topic \textit{Algebra I} as an example. First, we select one student $u$, whose representation is located far away from the majority in the 3D compressed latent space associated with the topic \textit{Algebra I}. Next, we identify the neighboring students around $u$ by locating those with the closest cosine distance to $u$ in the original 96-dimensional latent space. \cref{fig:latent:neigh:algebraI:accuracy:20,fig:latent:neigh:algebraI:attempts:20,fig:latent:neigh:algebraI:median:20} show the aggregated statistics for student $u$ (located in the center of the figure with attribute values highlighted in bold and a larger font) and the top 20 neighboring students surrounding $u$ (color-coded with a gradient from purple to yellow, indicating their cosine distance in the original latent space from smallest to largest). Across all three aspects, it's evident that the nearest neighboring students (highlighted in purple) closely resemble student 
$u$. The remaining students (colored in orange and yellow) also bear similarities to 
$u$ in certain aspects. When selecting students further from 
$u$ in the latent space, it is apparent that these individuals significantly differ from 
$u$ in tracing attributes compared to those who are closer. 

Furthermore, in \cref{fig:latent:neigh:algebraI:accuracy:80,fig:latent:neigh:algebraI:attempts:80,fig:latent:neigh:algebraI:median:80}, we show the top 80 neighboring students of student $u$. Notably, the students to the right of $u$, who are also distant from the majority of students, are now included. The aggregated values of these neighboring students' tracing attributes slightly differ from $u$. As shown in the figures, the representations reveal that these students on the right often have much lower average accuracy in their attempts compared to $u$, and they generally made fewer attempts than $u$ or tackled these questions at different weeks in the academic calendar. 

The aggregated statistics of the tracing attributes for each topic can provide an overview of a student's learning status. However, for a detailed understanding of student behaviors, performance and learning paths, it is essential to examine the student curriculum-based learning graphs more thoroughly. To further demonstrate that \texttt{CTGraph} captures both the statistical properties of students' learning status and the details of their learning paths relative to the curriculum, we provide an example from the topic \textit{Fractions}. Figure \ref{fig:neigh} shows the top 5 neighboring students for student 1 in the original latent space (arranged from the nearest to furthest in \cref{fig:neigh:fracciones:2,fig:neigh:fracciones:3,fig:neigh:fracciones:4,fig:neigh:fracciones:5,fig:neigh:fracciones:6}).  \cref{fig:neigh:fracciones:1,fig:neigh:fracciones:2,fig:neigh:fracciones:3,fig:neigh:fracciones:4,fig:neigh:fracciones:5,fig:neigh:fracciones:6} present the curriculum-based learning graph for each student. It's evident that all five students generally have low accuracy (around 0.5) in their attempts. 

Next, we demonstrate that how to examine the learning graphs of similar students to identify the commonality among students lagging behind. The learning graphs explicitly show the specifics of each student's choices in terms of concept order, intensity, and timing of the attempts. As shown in Figure \ref{fig:neigh}, each student and their nearest left/right neighbors have similar attributes on each concept and their learning paths also demonstrate similarity.
Furthermore, it is evident that these students generally achieved high accuracy on Concept 1 but significantly lower accuracy on immediate subsequent concepts (i.e., Concept 2 and Concept 5). After finishing Concept 6, they either skipped all other concepts after on that branch or achieved much lower accuracy (e.g., on Concept 14). In this way, educators can easily find out when and where these students started to fall behind based on the curriculum structure.

\subsubsection{Comparative Analysis}
In this part, we first illustrate the use of latent representation to identify cohort student groups, which are students that exhibit predominantly similar behaviors but also have subtle differences in certain aspects. We then use two different groups as examples and conduct comparative analysis within each group separately.

\begin{table}[h!b]
\centering
\captionsetup{justification=centering}
\begin{filecontents*}{sections/tables/funciones-I.csv}
row_id,stu_id,avg_acc,num_of_concepts,sum_exercises,median_week
1,7f631323-581f-4367-8100-fe1ba6db3c89,0.48,4,103,3
2,a62f9707-bc4c-4505-b837-63bbe01d616f,0.63,4,144,8
3,351f9110-80cf-4148-8f3a-04b94553e88a,0.49,4,83,3
4,5e1d56cc-b992-44cf-b899-1757e5cf9375,0.60,5,118,8
5,9610be74-6185-48ba-b407-6ace301b230e,0.56,6,203,13
\end{filecontents*}
\csvreader[
    before reading=\footnotesize
        \caption{Comparative Analysis for \textit{Functions I}: Aggregated Values of Tracing Attributes}\label{tab:funcionesI}
          \setlength{\tabcolsep}{2.5pt},
        tabular={|>{\centering\arraybackslash}m{0.15\linewidth} |>{\centering\arraybackslash}m{0.15\linewidth} |>{\centering\arraybackslash}m{0.15\linewidth}|>{\centering\arraybackslash}m{0.15\linewidth} |>{\centering\arraybackslash}m{0.15\linewidth}|},
    table head =\hline Student ID & Average Accuracy & \# Concepts & {\# Attempts} & Median Week No.\\\hline\hline,
    late after line= \\,
    late after last line=\\\hline
    ]{sections/tables/funciones-I.csv}{}
{\csvcoli & \csvcoliii & \csvcoliv & \csvcolv & \csvcolvi}
\end{table}

\begin{figure*}[t!]
     \centering
          \begin{subfigure}[b]{0.19\textwidth}
         \centering
         \includegraphics[width=\textwidth, 
         height=4cm
         ,trim={0cm 0cm 0cm 0cm},clip
         ]
        {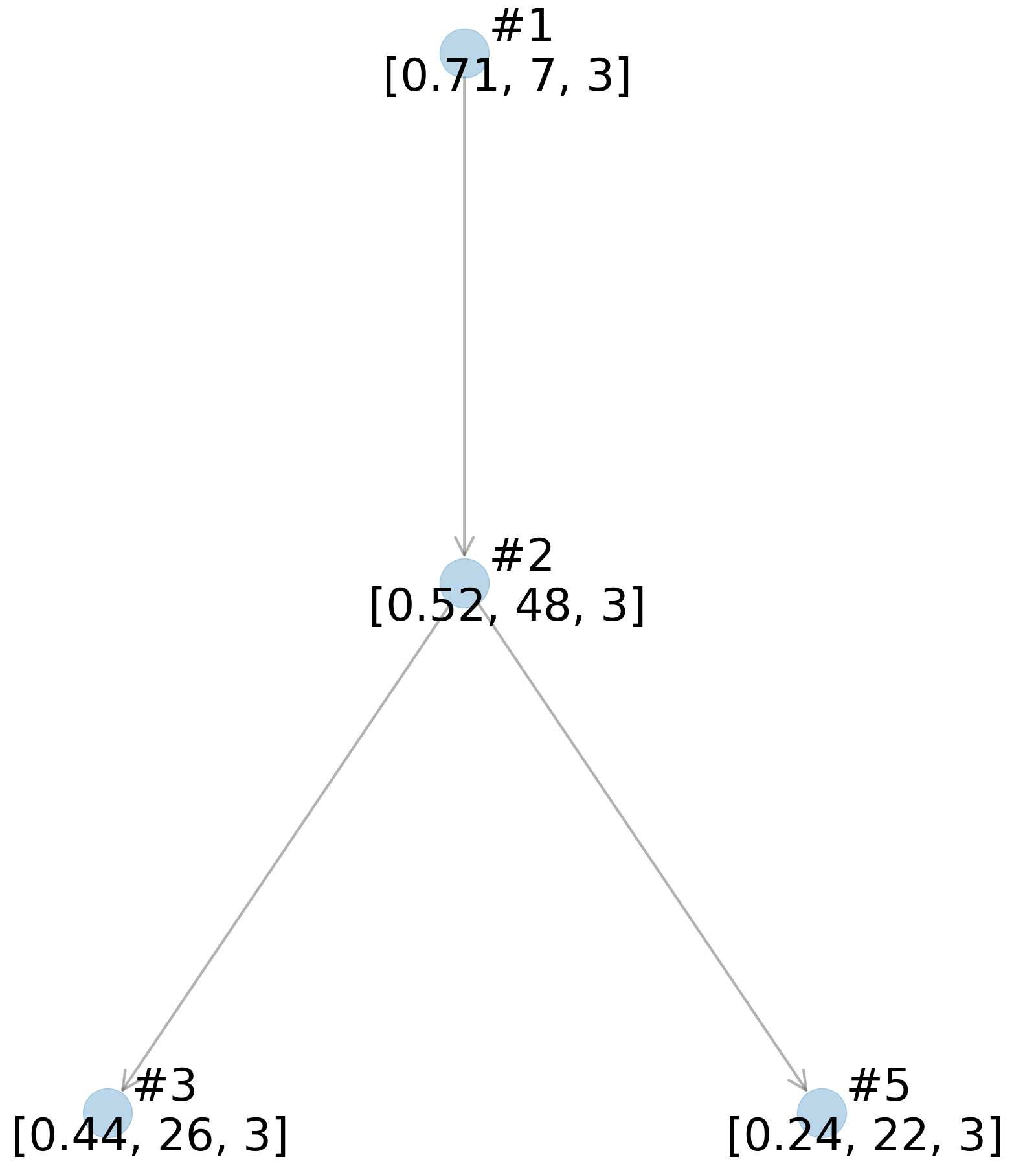}
         \caption{Student 1}
         \label{fig:comp:funcionesI:stu1}
     \end{subfigure}
          \begin{subfigure}[b]{0.19\textwidth}
         \centering
         \includegraphics[width=\textwidth, 
         height=4cm
         ,trim={0cm 0cm 0cm 0cm},clip
         ]
        {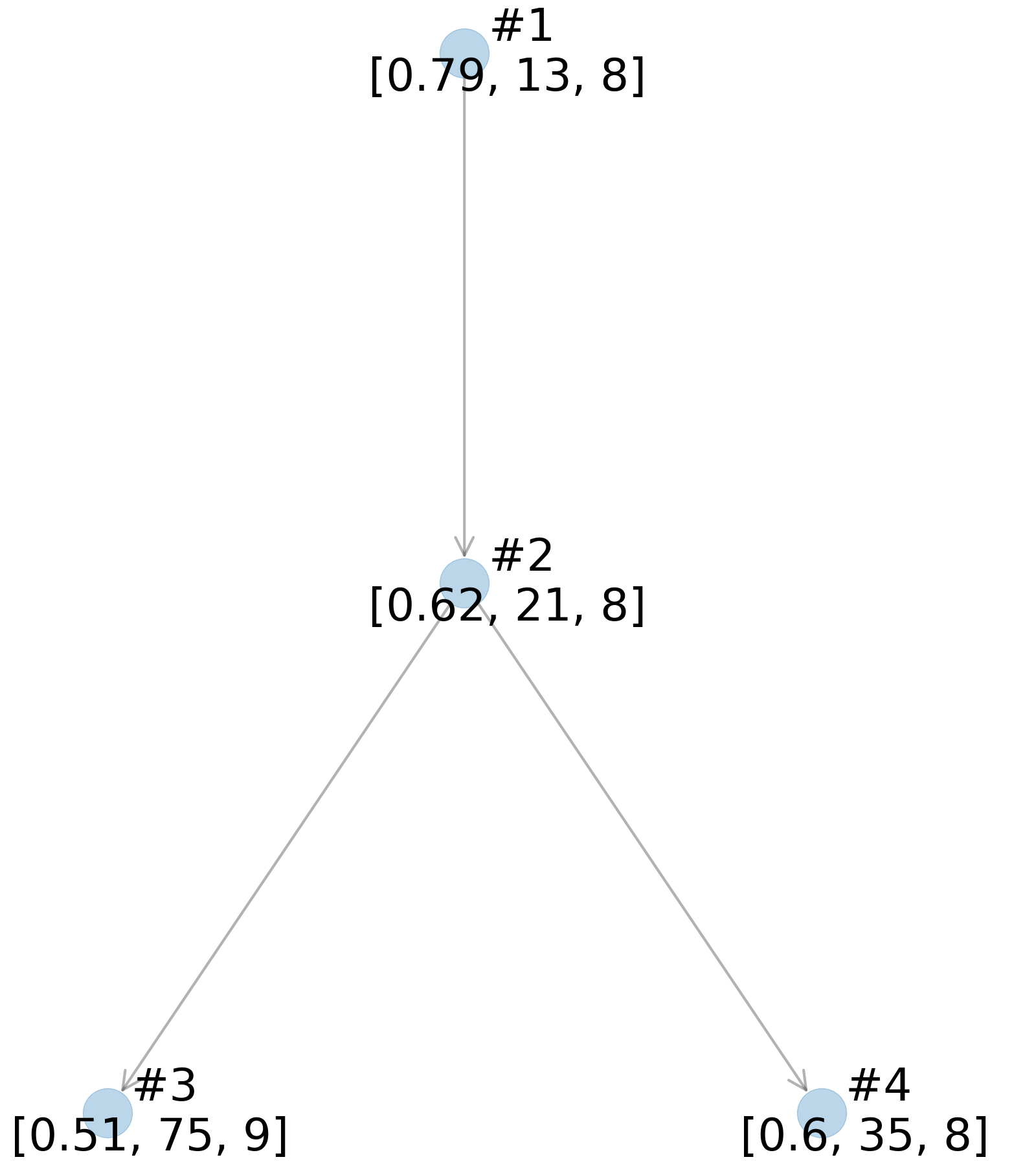}
         \caption{Student 2}
         \label{fig:comp:funcionesI:stu2}
     \end{subfigure}
               \begin{subfigure}[b]{0.19\textwidth}
         \centering
         \includegraphics[width=\textwidth, 
         height=4cm
         ,trim={0cm 0cm 0cm 0cm},clip
         ]
        {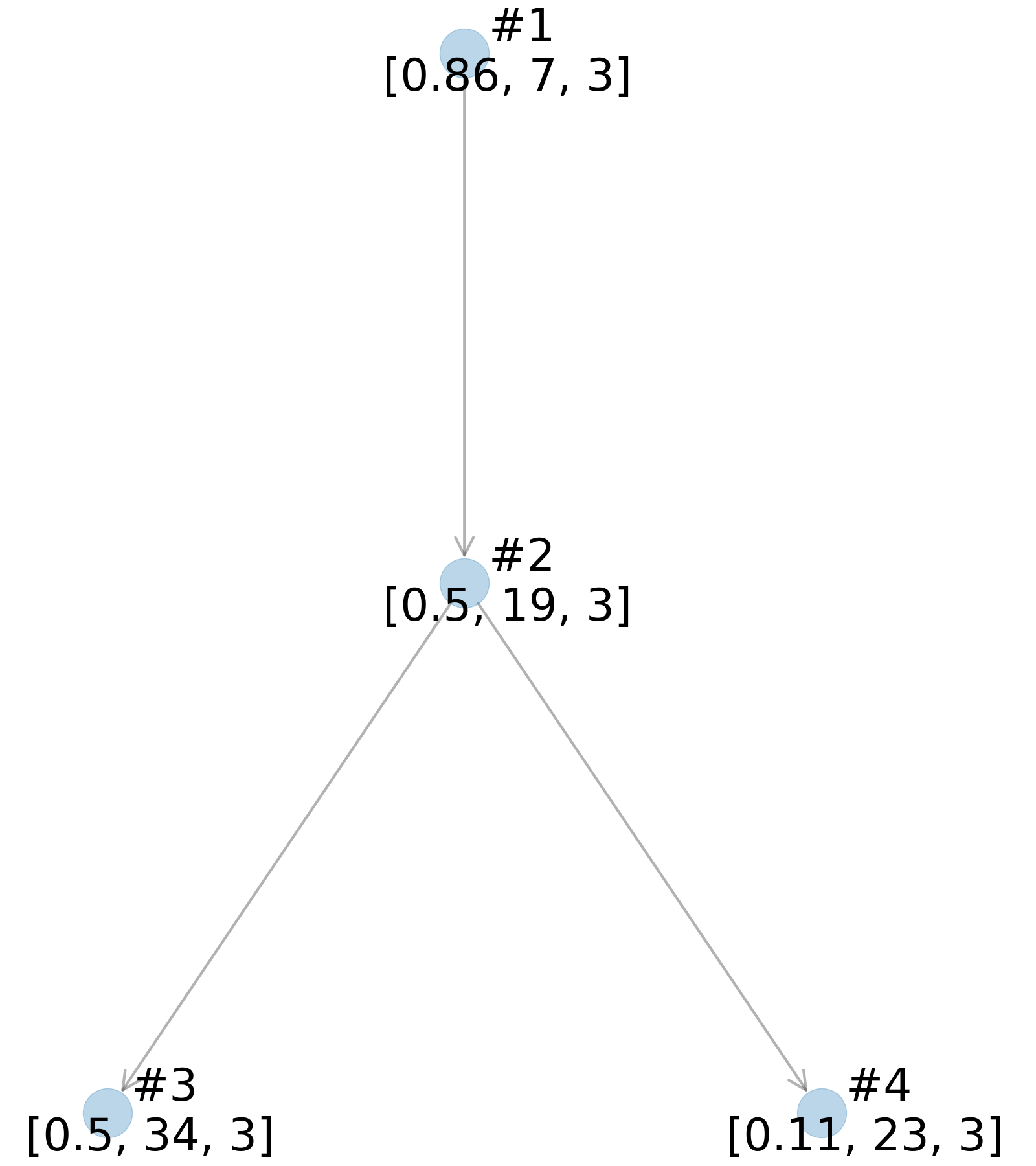}
         \caption{Student 3}
         \label{fig:comp:funcionesI:stu3}
     \end{subfigure}
               \begin{subfigure}[b]{0.19\textwidth}
         \centering
         \includegraphics[width=\textwidth, 
         height=4cm
         ,trim={0cm 0cm 0cm 0cm},clip
         ]
        {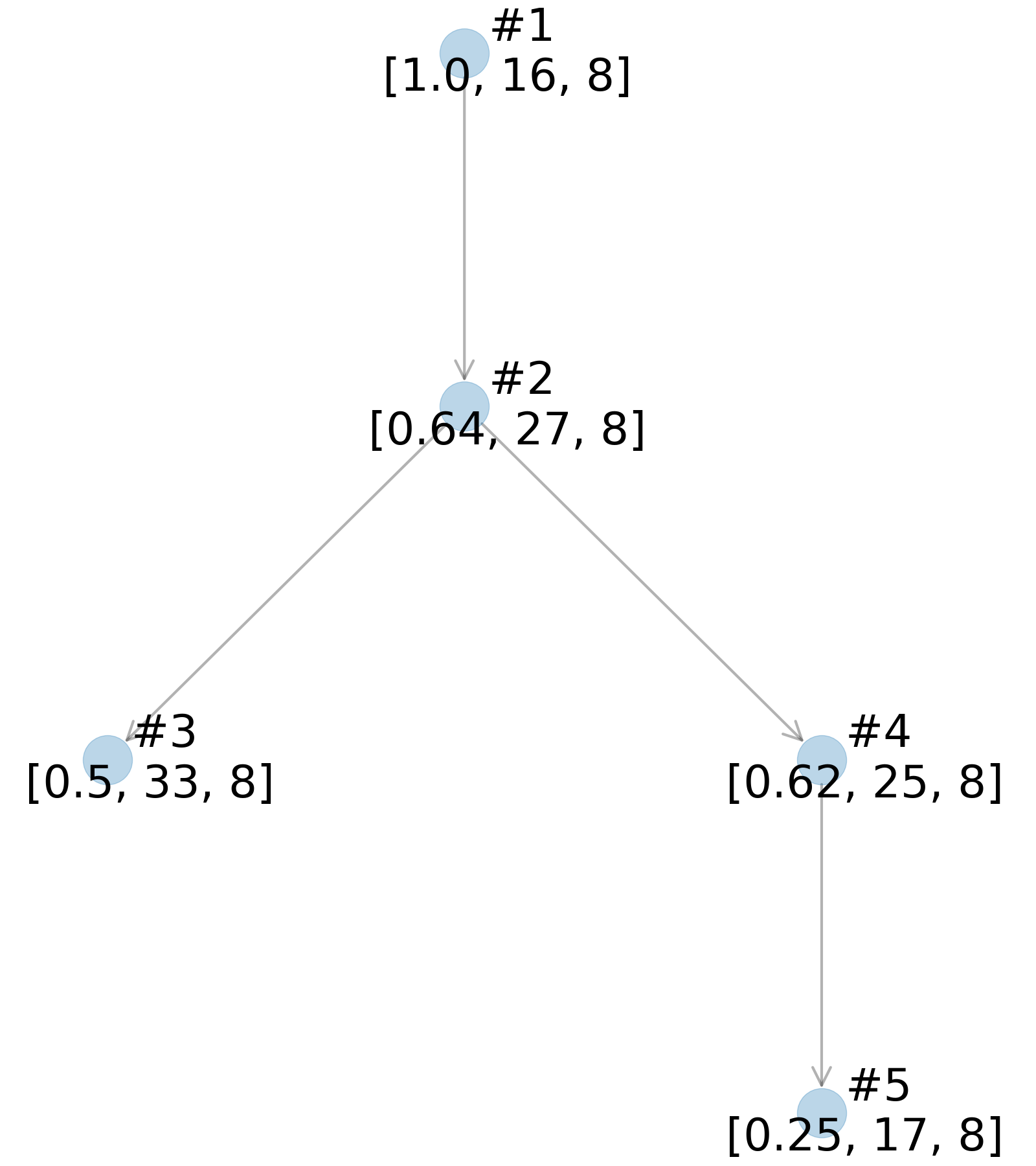}
         \caption{Student 4}
         \label{fig:comp:funcionesI:stu4}
     \end{subfigure}
               \begin{subfigure}[b]{0.19\textwidth}
         \centering
         \includegraphics[width=\textwidth, 
         height=4cm
         ,trim={0cm 0cm 0cm 0cm},clip
         ]
        {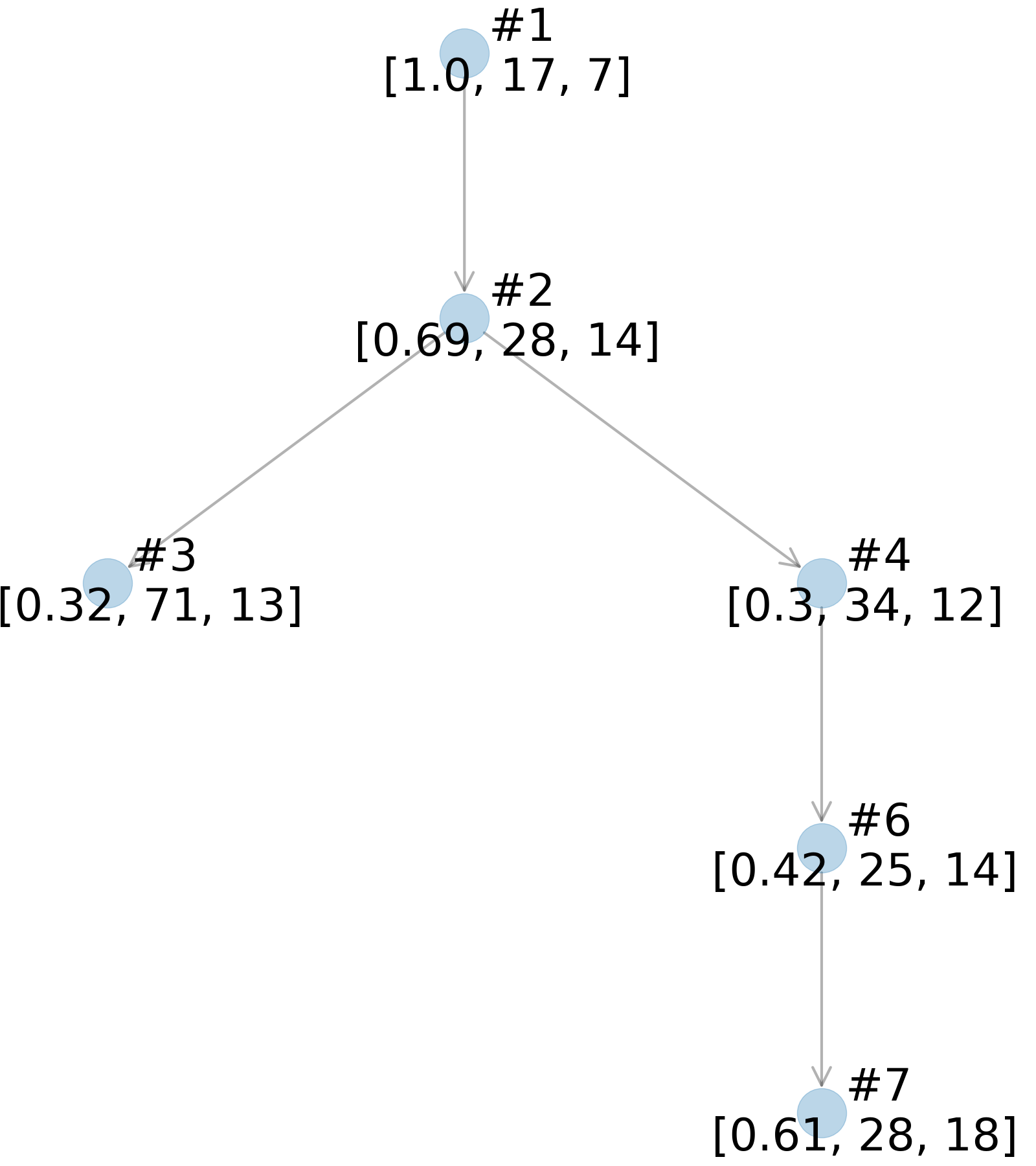}
         \caption{Student 5}
         \label{fig:comp:funcionesI:stu5}
     \end{subfigure}
         \caption{ Comparative Analysis for the Topic \textit{Funtions I} - Student Cohort and Their Curriculum-Based Learning Graphs 
        }
        \label{fig:comp:funcionesI}
\end{figure*}

\begin{figure*}[t!]
     \centering
     \begin{subfigure}[b]{0.18\textwidth}
         \centering
         \includegraphics[width=\textwidth, 
         height=4cm
         ,trim={0cm 0cm 0cm 0cm},clip
         ]
        {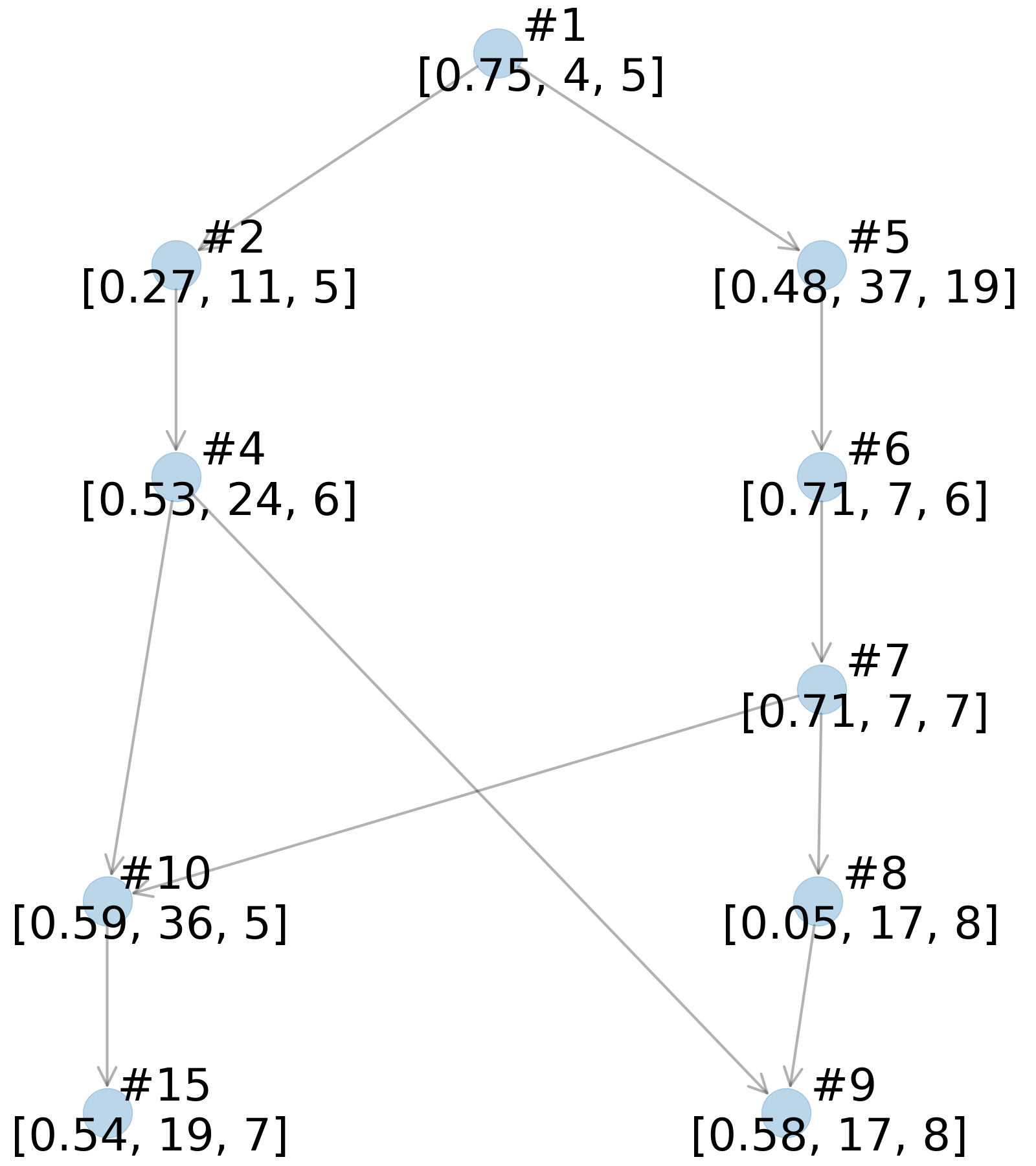}
         \caption{Student 1}
         \label{fig:comp:algebraII:stu1}
     \end{subfigure}
          \begin{subfigure}[b]{0.18\textwidth}
         \centering
         \includegraphics[width=\textwidth, 
         height=4cm
         ,trim={0cm 0cm 0cm 0cm},clip
         ]
        {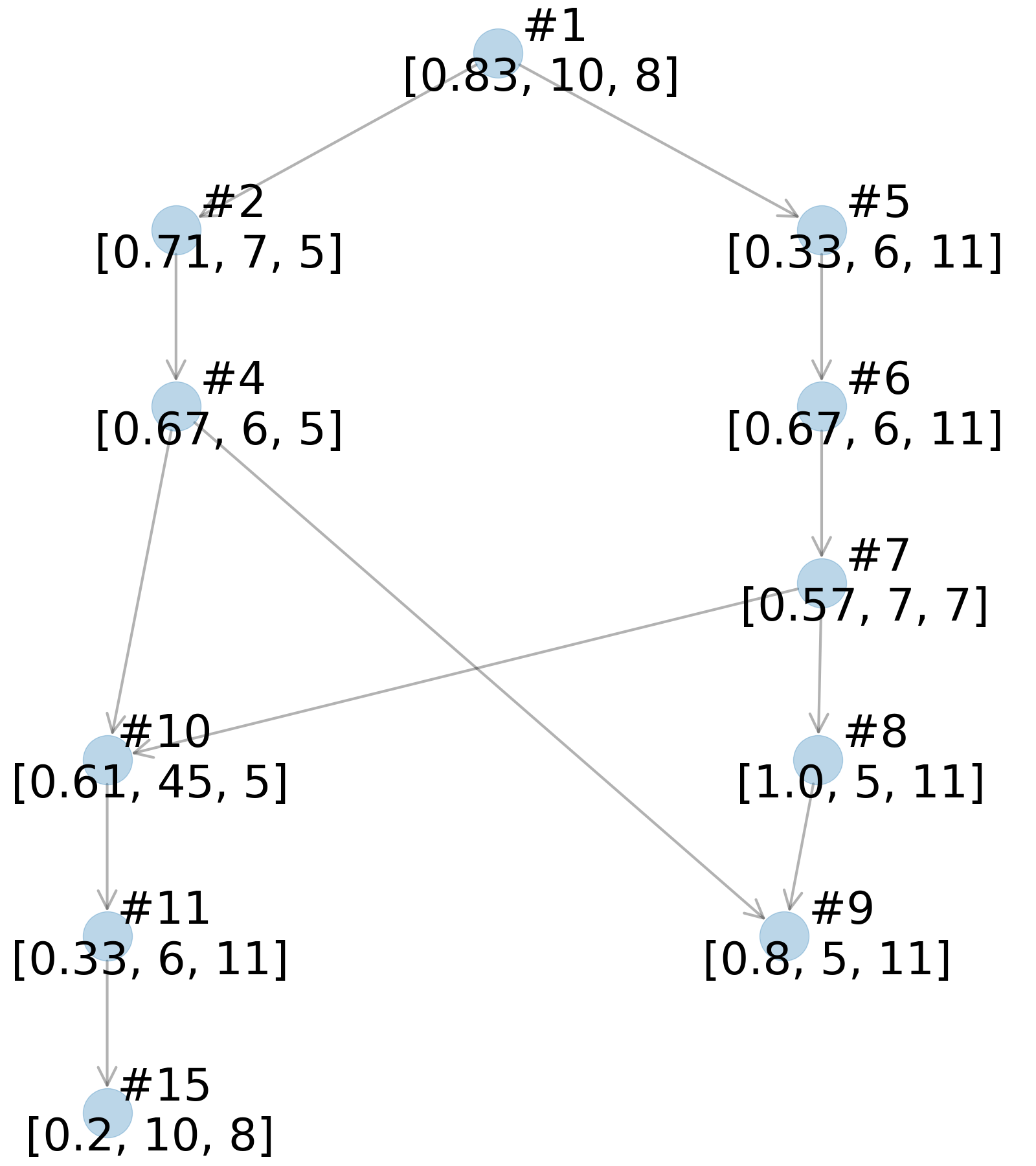}
         \caption{Student 2}
         \label{fig:comp:algebraII:stu2}
     \end{subfigure}
          \begin{subfigure}[b]{0.18\textwidth}
         \centering
         \includegraphics[width=\textwidth, 
         height=4cm
         ,trim={0cm 0cm 0cm 0cm},clip
         ]
        {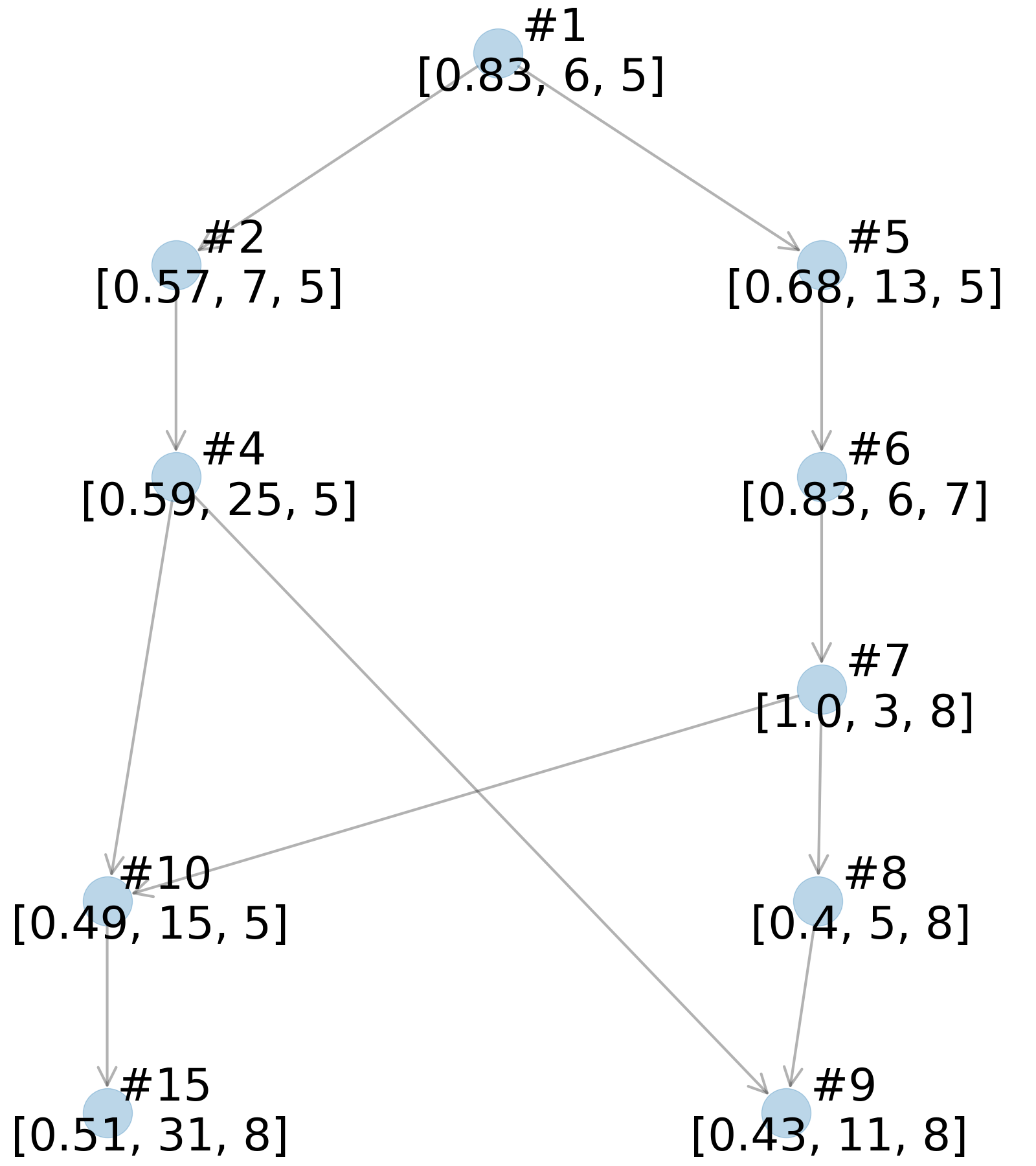}
         \caption{Student 3}
         \label{fig:comp:algebraII:stu3}
     \end{subfigure}
               \begin{subfigure}[b]{0.18\textwidth}
         \centering
         \includegraphics[width=\textwidth, 
         height=4cm
         ,trim={0cm 0cm 0cm 0cm},clip
         ]
        {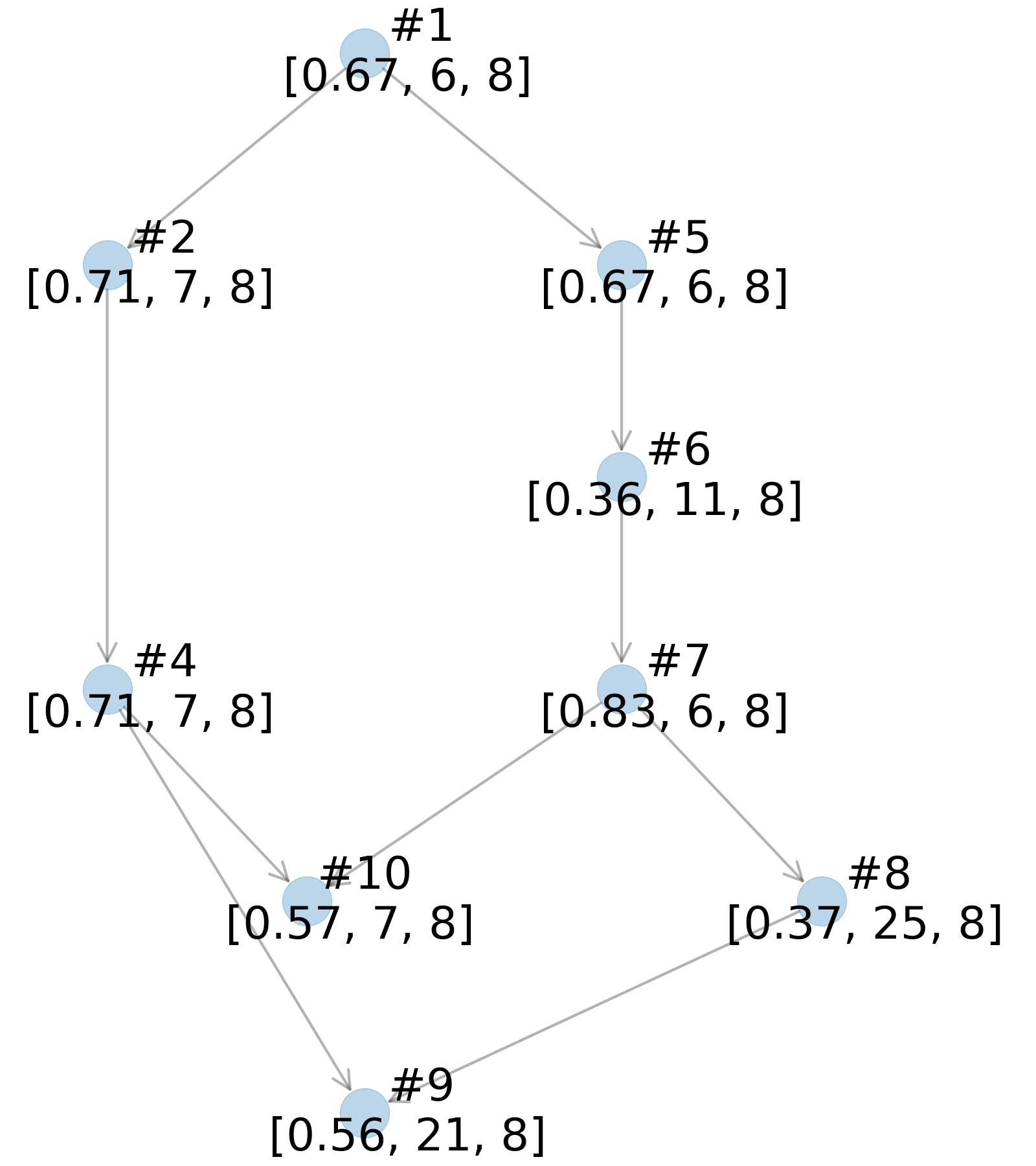}
         \caption{Student 4}
         \label{fig:comp:algebraII:stu4}
     \end{subfigure}
               \begin{subfigure}[b]{0.18\textwidth}
         \centering
         \includegraphics[width=\textwidth, 
         height=4cm
         ,trim={0cm 0cm 0cm 0cm},clip
         ]
        {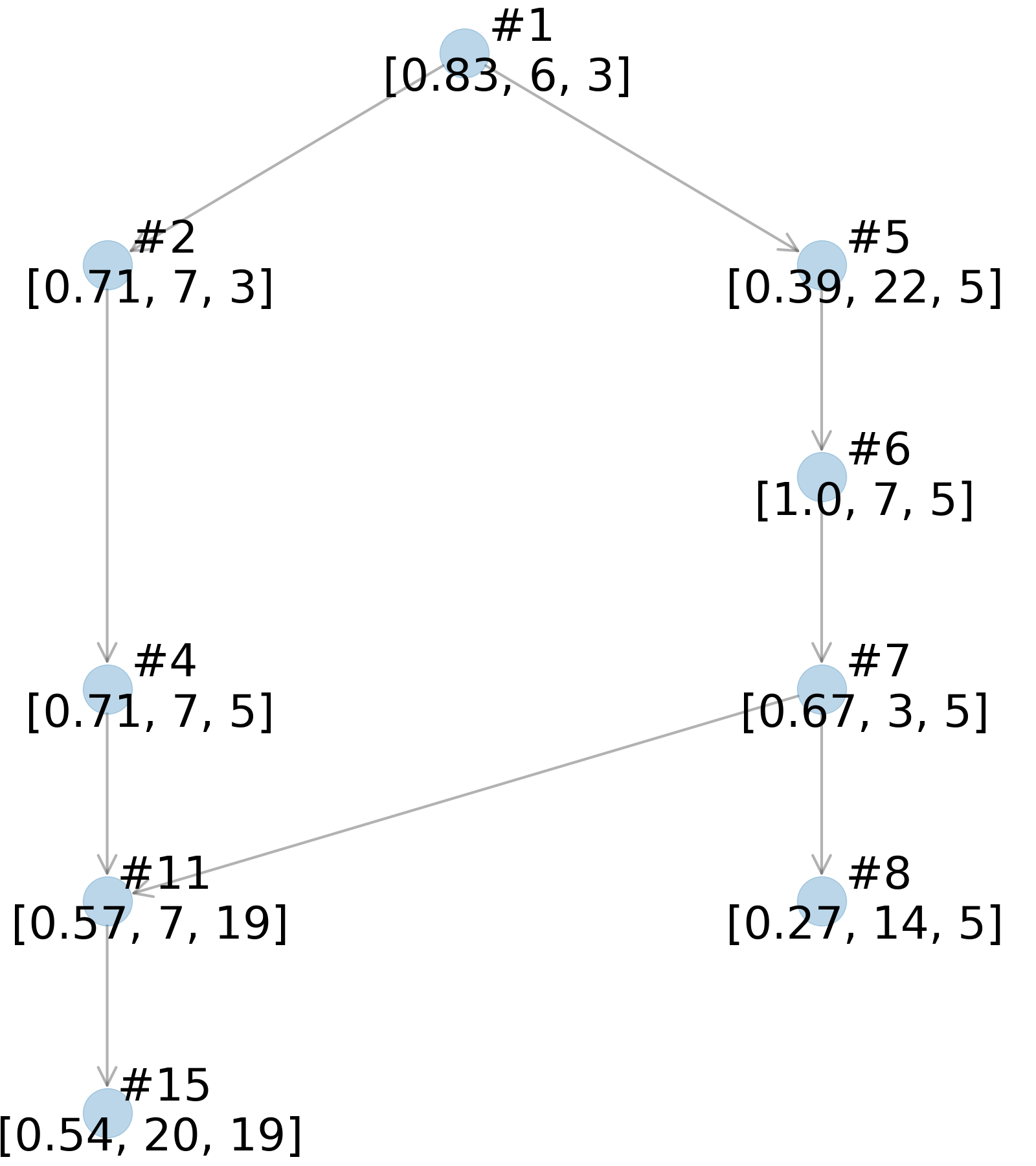}
         \caption{Student 5}
         \label{fig:comp:algebraII:stu5}
     \end{subfigure}
               \begin{subfigure}[b]{0.18\textwidth}
         \centering
         \includegraphics[width=\textwidth, 
         height=4cm
         ,trim={0cm 0cm 0cm 0cm},clip
         ]
        {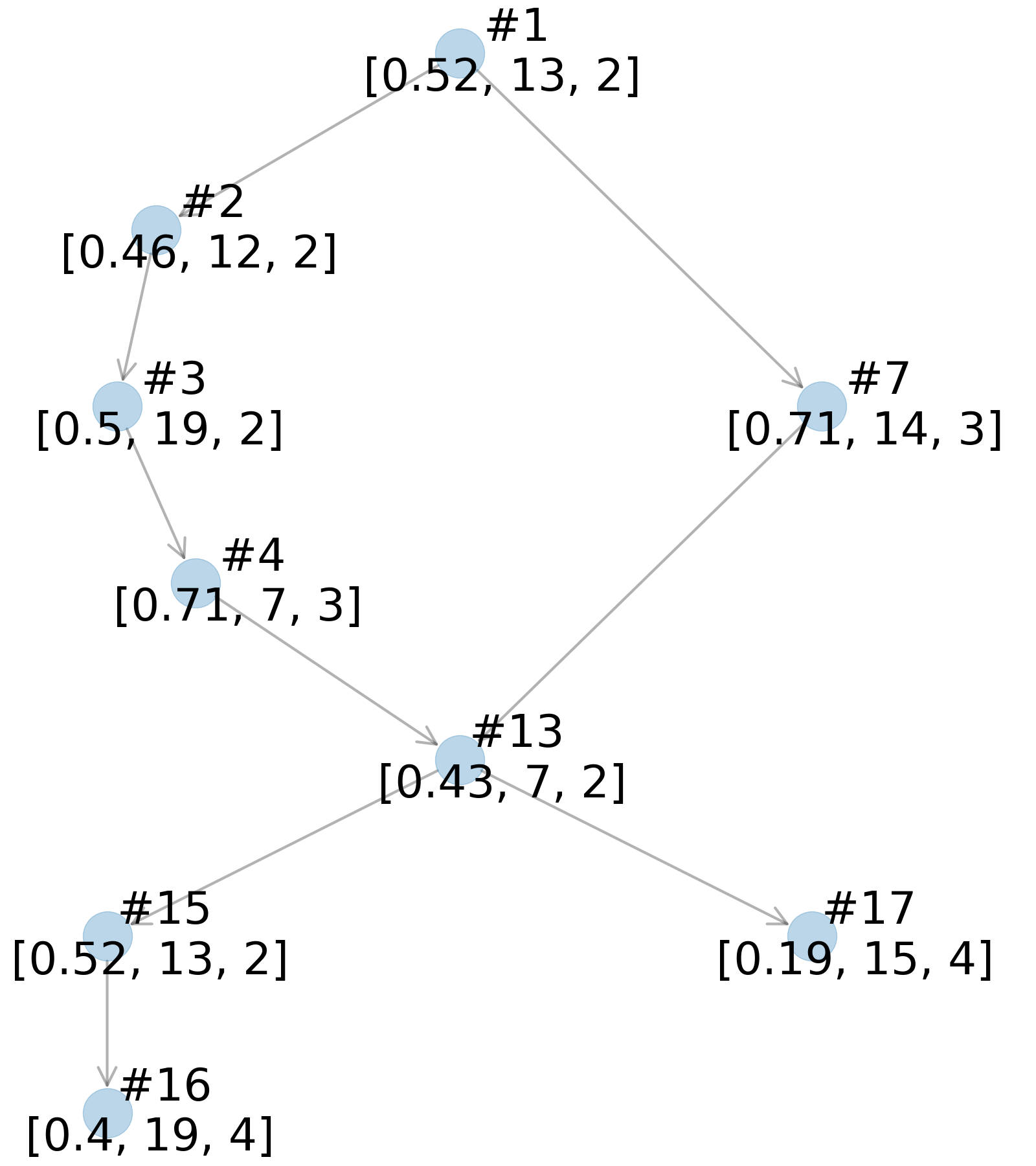}
         \caption{Student 6}
         \label{fig:comp:algebraII:stu6}
     \end{subfigure}
                    \begin{subfigure}[b]{0.18\textwidth}
         \centering
         \includegraphics[width=\textwidth, 
         height=4cm
         ,trim={0cm 0cm 0cm 0cm},clip
         ]
        {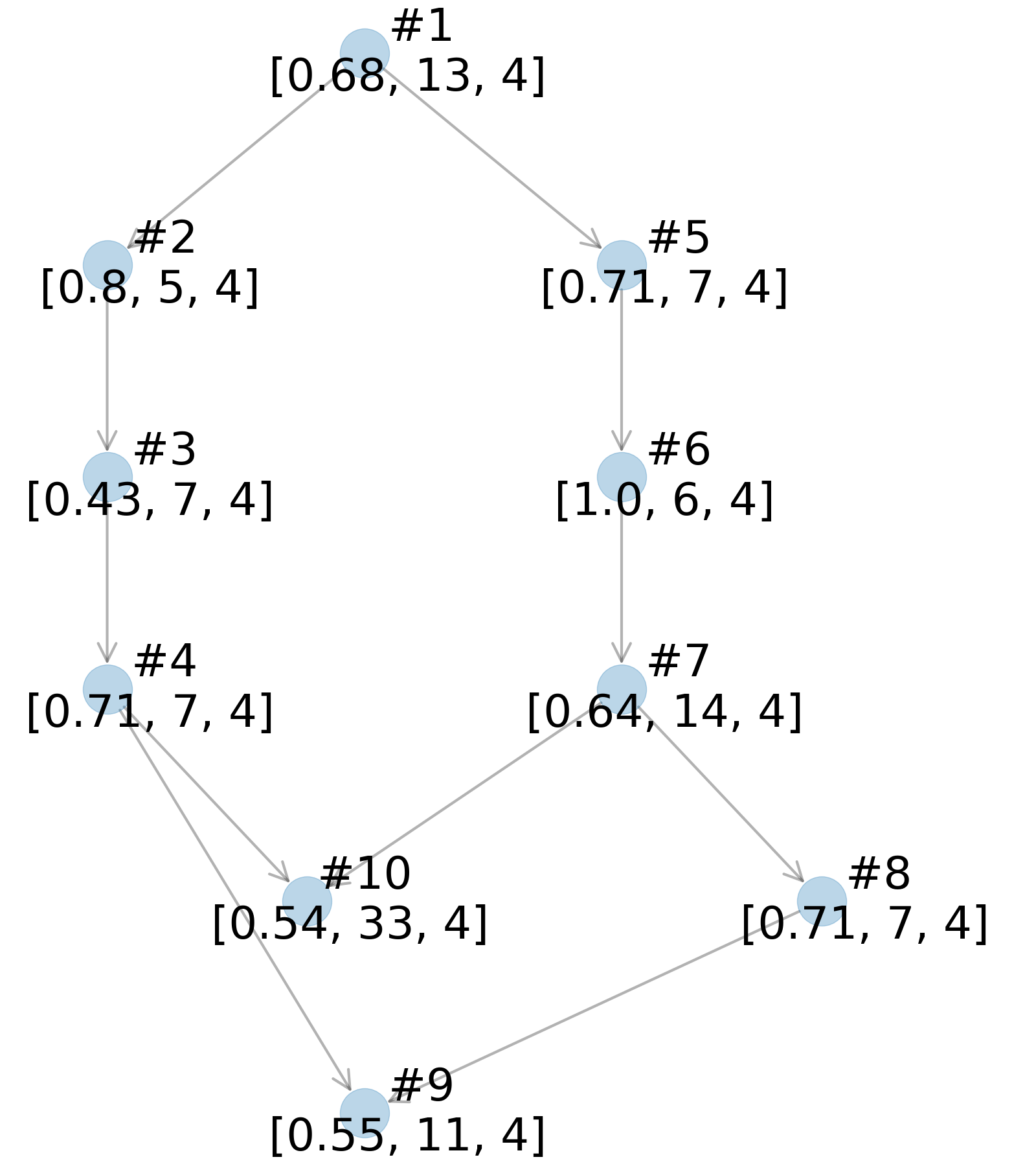}
         \caption{Student 7}
         \label{fig:comp:algebraII:stu7}
     \end{subfigure}
               \begin{subfigure}[b]{0.18\textwidth}
         \centering
         \includegraphics[width=\textwidth, 
         height=4cm
         ,trim={0cm 0cm 0cm 0cm},clip
         ]
        {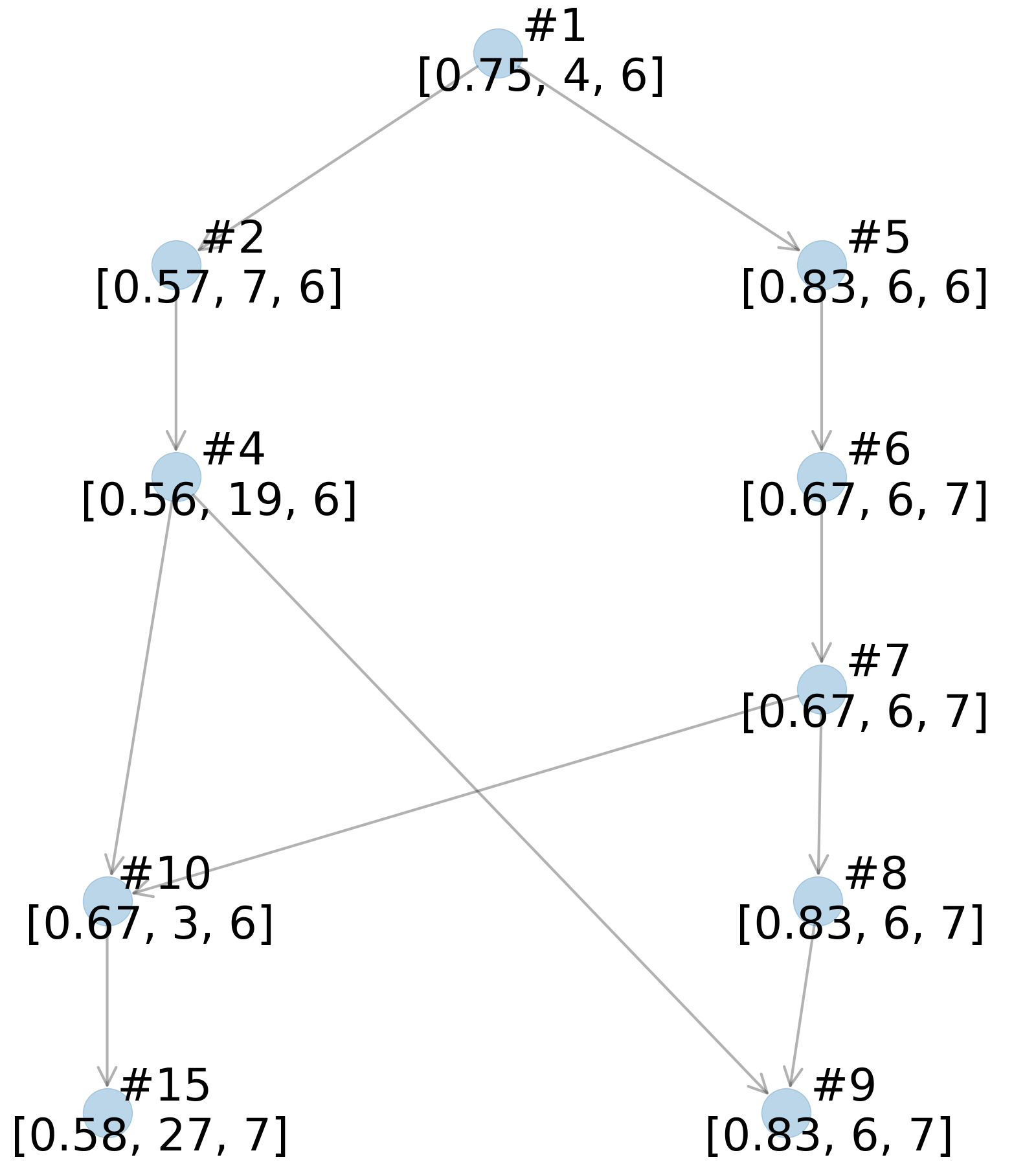}
         \caption{Student 8}
         \label{fig:comp:algebraII:stu8}
     \end{subfigure}
               \begin{subfigure}[b]{0.18\textwidth}
         \centering
         \includegraphics[width=\textwidth, 
         height=4cm
         ,trim={0cm 0cm 0cm 0cm},clip
         ]
        {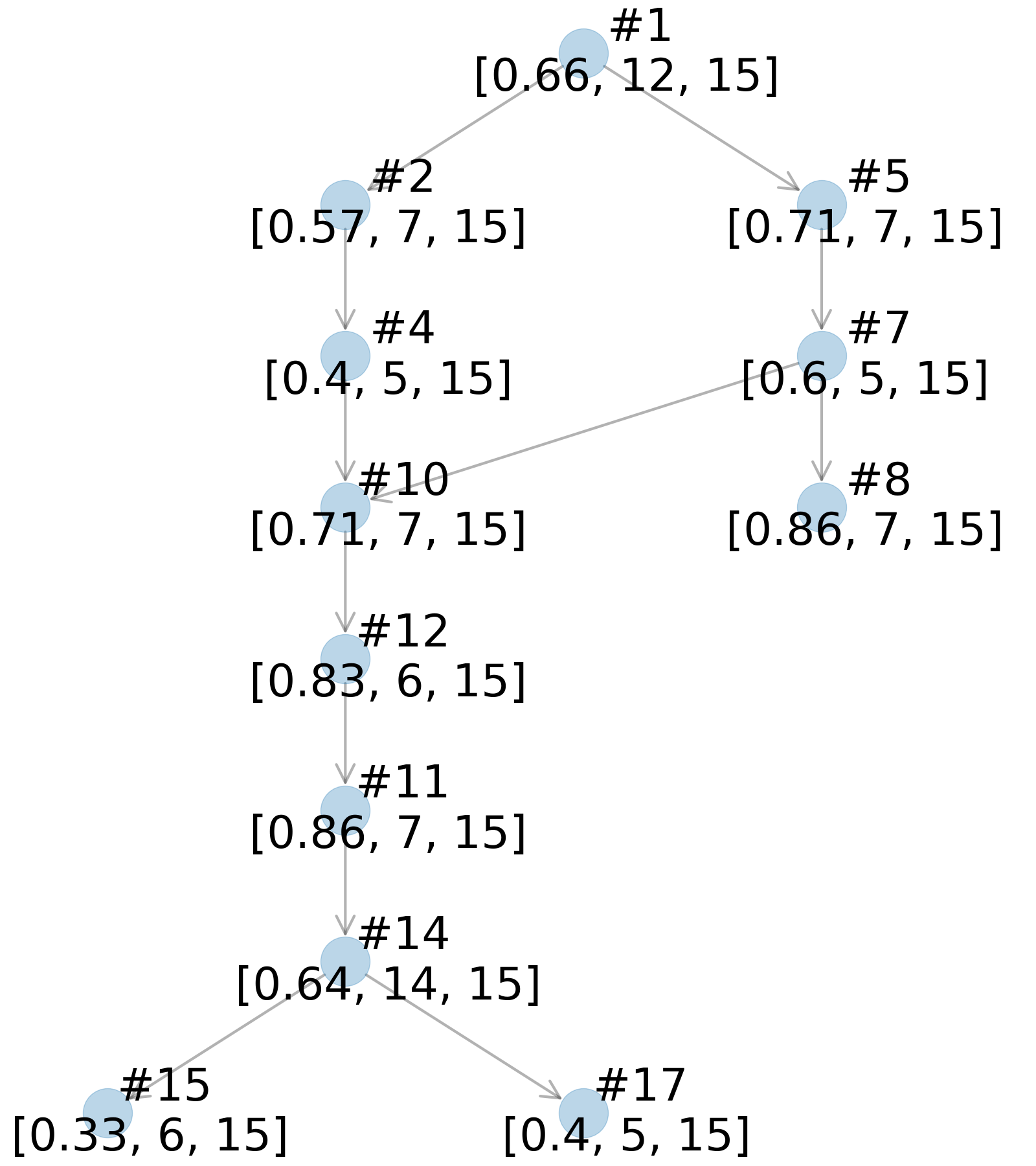}
         \caption{Student 9}
         \label{fig:comp:algebraII:stu9}
     \end{subfigure}
               \begin{subfigure}[b]{0.18\textwidth}
         \centering
         \includegraphics[width=\textwidth, 
         height=4cm
         ,trim={0cm 0cm 0cm 0cm},clip
         ]
        {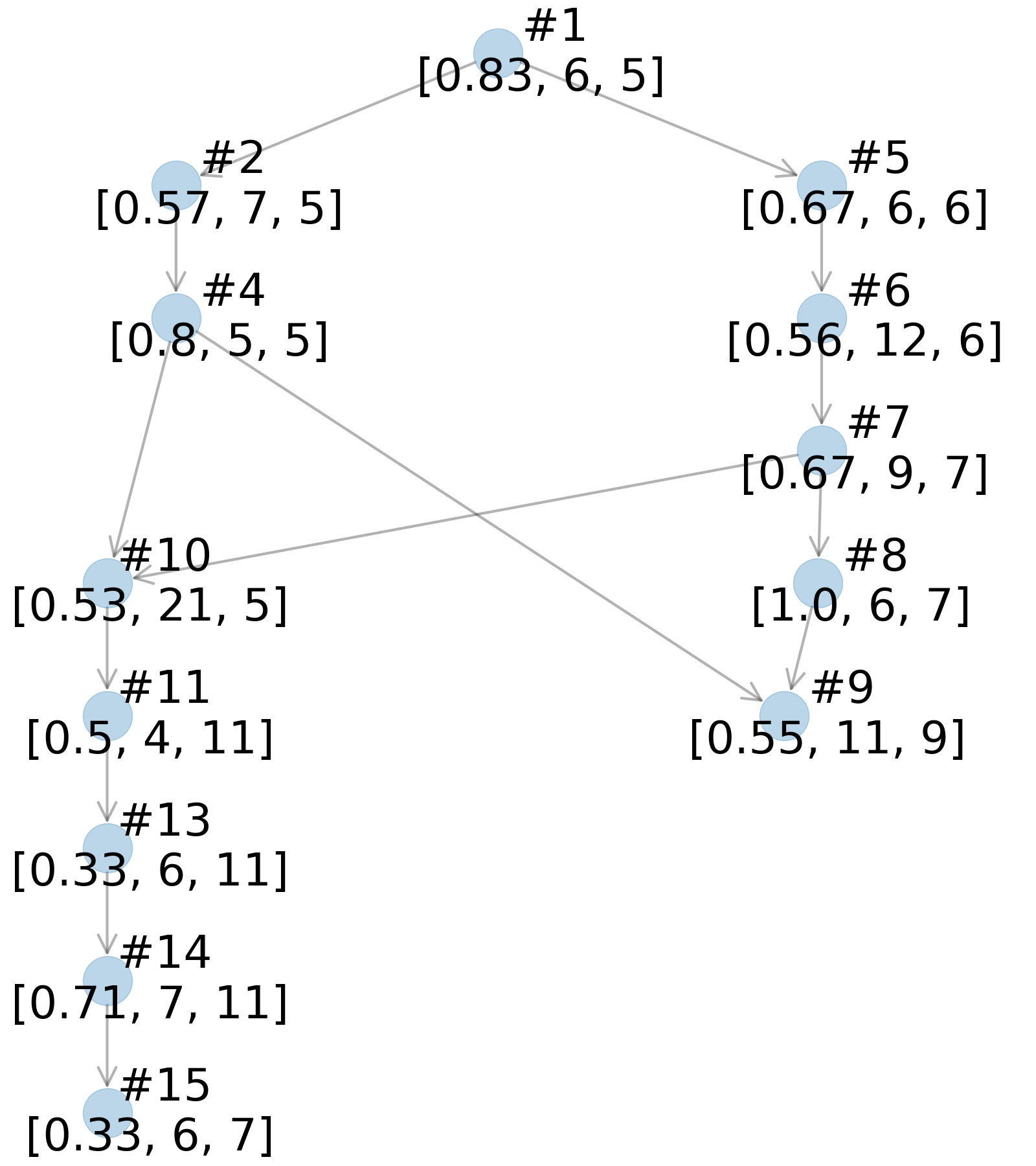}
         \caption{Student 10}
         \label{fig:comp:algebraII:stu10}
     \end{subfigure}

         \caption{ Comparative Analysis for the Topic \textit{Algebra II} - Student Cohort and Their Curriculum-Based Learning Graphs
        }
        \label{fig:comp:algebraII}
\end{figure*}

\begin{table}[htb]
\centering
\captionsetup{justification=centering}
\begin{filecontents*}{sections/tables/algebra-II.csv}
row_id,stu_id,avg_acc,num_of_concepts,sum_exercises,median_week
1,90b839f3-82f1-4f33-ab14-088c352a93f6,0.52,10,179,6
2,736c56a7-7031-4e1f-ab7c-a31831aa4e99,0.61,11,113,8
3,4c5e512c-ea4e-4dc2-8487-9fe375c2ceea,0.63,10,122,6
4,c039bda6-4324-4c52-82b1-6c2255226a1b,0.61,9,96,8
5,2c8980b9-51f3-4b1c-ae7f-4cfa86b0ca9a,0.63,9,93,5
6,6006926a-fc7a-4bfc-afc7-ccf661d0e4e9,0.49,9,119,2
7,7d9720d5-4028-4eba-af7e-2ff2c10ee0f6,0.68,10,110,4
8,2b2c8f36-e8d5-46d4-acc2-46dd5a9fe25c,0.70,10,90,6
9,3bfe48b6-10cc-4a6e-af2a-a486bb8ea798,0.63,12,88,15
10,e7955747-62b4-46d0-9dc8-9c9a4dc0c043,0.62,13,106,7
\end{filecontents*}
\csvreader[
    before reading=\footnotesize
        \caption{Comparative Analysis for \textit{Algebra II}: Aggregated Values of Tracing Attributes}\label{tab:algebraI}
          \setlength{\tabcolsep}{2.5pt},
        tabular={|>{\centering\arraybackslash}m{0.15\linewidth} |>{\centering\arraybackslash}m{0.15\linewidth} |>{\centering\arraybackslash}m{0.15\linewidth}|>{\centering\arraybackslash}m{0.15\linewidth} |>{\centering\arraybackslash}m{0.15\linewidth}|},
    table head =\hline Student ID & Average Accuracy & \# Concepts & {\# Attempts} & Median Week No.\\\hline\hline,
    late after line= \\,
    late after last line=\\\hline
    ]{sections/tables/algebra-II.csv}{}
{\csvcoli & \csvcoliii & \csvcoliv & \csvcolv & \csvcolvi}
\label{tab:algebraII}
\end{table}

To derive a cohort group, we first select a student as the starting point in the latent space and another student as the ending point. Let $s$ and $e$ denote the representation vectors of these two students. Secondly, we compute vector $v = s-e$. We then search the original latent space for students whose positions are closest to the vector $v$, as measured by the cosine distance. The top $k$ closest students found in this way, along with the two endpoint students, form a cohort group.


First, we show one example for the topic \textit{Functions I}, in which we demonstrate a cohort group where the students follow similar learning orders on the concepts. 
Table \ref{tab:funcionesI} shows the aggregated values of tracing attribute for each student in such group, and Figure \ref{fig:comp:funcionesI} shows their curriculum-based learning graphs. It is evident that the behaviors of all five students are largely similar in terms of sequence of learning concepts, average accuracy, the total number of attempts, and timing of learning. However, there are variances in the coverage of concepts as well as subtle difference in accuracy on certain concepts. For instance, Student 4 and Student 5 achieved higher accuracy for Concept 1 and Concept 2 but lower accuracy for the additional concepts they tackled later. This suggests that even though all these students have comparable average performance (in terms of average accuracy) as compared to Student 1-3, Students 4 and 5 actually made significant more effort on advanced concepts. Thus, despite their lower scores on these advanced concepts (i.e., where they started to lag behind), their stronger performance on basic concepts (Concepts 1-2) and efforts to broaden their knowledge base should be highlighted in the assessment of their achievement.

Secondly, we provide another example for the topic \textit{Algebra II}, in which the curriculum structure is more intricate. Table \ref{tab:algebraII} presents the aggregated values of tracing attributes for each student in a specific cohort group, and Figure \ref{fig:comp:algebraII} depicts their curriculum-based learning graphs. To compare with the majority of students, we check details in the 3D visualization view for the topic \textit{Algebra II} in \cref{fig:latent:algebraII:acc,fig:latent:algebraII:sum_exercises,fig:latent:algebraII:median_week}. As shown in Figure \ref{fig:latent:algebraII:acc}, compared to the majority of students, students in this cohort group generally achieved much lower average accuracy in their attempts. However, their total numbers of attempts are comparable to the majority. 

Furthermore, when compared to the other students in this group, there is also a noticeable expansion in the coverage of concepts for the last two students (i.e., Student 9 and 10). Analyzing their curriculum-based learning graphs in Figure \ref{fig:comp:algebraII:stu9} and \ref{fig:comp:algebraII:stu10}, we observe that 1) they performed better for the first 12 concepts in the topic's curriculum structure, and 2) their accuracy for the remaining concepts (i.e., Concepts 13-17) they tackled—compared to the rest of the group—is considerably lower. This suggests that these two students exerted greater effort in exploring advanced concepts, even if their performance on these advanced concepts is lower and their average performance remains consistent with the others in the cohort group.


\section{Conclusion}
In this study, we present \texttt{CTGraph}, a graph-level representation learning approach to profile learner behaviors and performance in a self-supervised manner. Our analysis shows the importance of considering the nuanced differences in student learning paths and multivariate metrics related to their behaviors and performance. The intricate details of students' curriculum-based learning graphs reveal the depth and breadth of their learning journeys. Notably, students may achieve similar overall performance but diverge in their engagement with advanced concepts or their mastery of foundational knowledge. This nuanced perspective emphasizes the value of holistic evaluation, recognizing students' efforts and progress in the context of the curriculum structure. As such, our approach can provide educators, pedagogy researchers and curriculum designers with a rich understanding of individual students' learning journeys and the learning effectiveness of various learning strategies and paths. Thereby, it opens new opportunities to develop and design more tailored and effective pedagogical strategies for ITS applications.


\section*{Acknowledgment}
This project has received funding from the European Union’s Horizon 2020 Research and Innovation Programme under the HUMAN+ COFUND Marie Skłodowska-Curie grant agreement No. 945447.

\bibliographystyle{IEEEtran}
\bibliography{sections/ref}
\end{document}